\documentclass[journal]{IEEEtran}

\usepackage{cite}
\usepackage{hyperref}
\usepackage[pdftex]{graphicx}
\usepackage{amsmath}
\usepackage[linesnumbered, ruled, vlined]{algorithm2e}
\usepackage{array}
\usepackage{multirow}
\usepackage{xcolor}
\usepackage{soul}
\usepackage{subfigure}
\usepackage{amssymb}
\usepackage{ulem}
\usepackage{caption}
\newcommand{\nop}[1]{}

\begin{document}

\normalem

\title{Evaluating Adversarial Attacks on Driving Safety in Vision-Based Autonomous Vehicles}

\author{Jindi~Zhang,
        Yang~Lou,
        Jianping~Wang,~\IEEEmembership{Senior~Member,~IEEE,}
        Kui~Wu,~\IEEEmembership{Senior~Member,~IEEE,}
        Kejie~Lu,~\IEEEmembership{Senior~Member,~IEEE,}
        and~Xiaohua~Jia,~\IEEEmembership{Fellow,~IEEE}%
\thanks{This work was partially supported by Hong Kong Research Grant Council under NSFC/RGC N$\_$CityU 140/20, Science and Technology Innovation Committee Foundation of Shenzhen under Grant No. JCYJ20200109143223052.}%
\thanks{Jindi Zhang and Yang Lou contributed equally to this work.}%
\thanks{Jindi Zhang, Yang Lou, Jianping Wang and Xiaohua Jia are with the Department of Computer Science, City University of Hong Kong, Hong Kong (e-mail: \href{mailto:jd.zhang@my.cityu.edu.hk}{jd.zhang@my.cityu.edu.hk}; \href{mailto:yanglou3-c@my.cityu.edu.hk}{yanglou3-c@my.cityu.edu.hk}; \href{mailto:jianwang@cityu.edu.hk}{jianwang@cityu.edu.hk}; \href{mailto:csjia@cityu.edu.hk}{csjia@cityu.edu.hk}).}%
\thanks{Kui Wu is with the Department of Computer Science, University of Victoria, Victoria BC V8P 5C2, Canada (e-mail: \href{mailto:wkui@uvic.ca}{wkui@uvic.ca}).}%
\thanks{Kejie Lu is with the Department of Computer Science and Engineering, University of Puerto Rico at Mayag{\"u}ez, Mayag{\"u}ez, Puerto Rico 00682 (e-mail: \href{mailto:kejie.lu@upr.edu}{kejie.lu@upr.edu}).}%
\thanks{This article has been accepted for publication in a future issue of this journal, but has not been fully edited. Content may change prior to final publication. Citation information: DOI 10.1109/JIOT.2021.3099164, IEEE Internet of
Things Journal. 2327-4662 (c) 2021 IEEE. Personal use is permitted, but republication/redistribution requires IEEE permission. See \url{http://www.ieee.org/publications_standards/publications/rights/index.html} for more information.}
}

\maketitle

\begin{abstract}
In recent years, many deep learning models have been adopted in autonomous driving. At the same time, these models introduce new vulnerabilities that may compromise the safety of autonomous vehicles. Specifically, recent studies have demonstrated that adversarial attacks can cause a significant decline in detection precision of deep learning-based 3D object detection models. Although driving safety is the ultimate concern for autonomous driving, there is no comprehensive study on the linkage between the performance of deep learning models and the driving safety of autonomous vehicles under adversarial attacks. In this paper, we investigate the impact of two primary types of adversarial attacks, perturbation attacks and patch attacks, on the driving safety of vision-based autonomous vehicles rather than the detection precision of deep learning models. In particular, we consider two state-of-the-art models in vision-based 3D object detection, Stereo R-CNN and DSGN. To evaluate driving safety, we propose an end-to-end evaluation framework with a set of driving safety performance metrics. By analyzing the results of our extensive evaluation experiments, we find that (1) the attack's impact on the driving safety of autonomous vehicles and the attack's impact on the precision of 3D object detectors are decoupled, and (2) the DSGN model demonstrates stronger robustness to adversarial attacks than the Stereo R-CNN model. In addition, we further investigate the causes behind the two findings with an ablation study. The findings of this paper provide a new perspective to evaluate adversarial attacks and guide the selection of deep learning models in autonomous driving.
\end{abstract}

\begin{IEEEkeywords}
Adversarial attacks, driving safety, 3D object detection, autonomous driving.
\end{IEEEkeywords}

\vspace{-0.1in}
\section{Introduction}
\label{sec.introduction}

\IEEEPARstart{O}{V}{E}{R} the past decade, autonomous driving has gained significant developments and demonstrated its great commercial potentials~\cite{davies18waymo, hawkins19tesla}. The commercial potentials have attracted enormous investment as well as various malicious attacks~\cite{ren19security, wyglinski13security}, for example,  close-proximity sensor attacks, remote cyberattacks, perturbation attacks, and patch attacks.

Environment perception and other tasks of autonomous driving systems heavily rely on deep learning models. Researchers have demonstrated that adversarial examples, which are originally designed to affect general-purpose deep learning models, can also be used to cause malfunctions in autonomous driving tasks~\cite{zhou18deep, ranjan19attacking, eykholt18robust, lu17adversarial, metzen17universal, song18physical, chen18shapeshifter, zhang20adversarial, mathew20monocular, cao19adversarial}. In these studies, researchers usually use the decline of accuracy, or the erroneous rate increase of the deep learning models, to measure the effectiveness of attacks. Amplified by media reports, these attacks are casting cloud and posing psychological barriers to the broader adoption of autonomous driving~\cite{slovick2017security}. 

From the perspective of autonomous driving, however, the ultimate concern is driving safety. Without a doubt, the inaccurate detection results of a deep learning model in the presence of attacks may impact driving safety, and in some situations, misdetection of traffic signs~\cite{eykholt18robust} might have disastrous consequence. Nevertheless, \textit{driving safety is a combined effort of many factors in a dynamic environment}, and the deteriorated model performance does not necessarily lead to safety hazards. The linkage between the performance of a deep learning model under adversarial attacks and driving safety is not studied in the literature. In particular, there are no clear answers to the following questions: 1) Does the precision decline or the erroneous rate increase of the deep learning models under attacks represent their robustness in regard to driving safety of autonomous vehicles? In other words, does a larger decline in accuracy of an attacked deep learning model indicate a higher risk of driving safety? Similarly, does a slight decrease in accuracy of a deep learning model under attacks indicate mild risk of driving safety? 2) If the answers to the previous questions are all no, what are the reasons behind?

In this paper, we aim to answer the aforementioned questions by evaluating the impact of two types of representative adversarial attacks, \emph{perturbation attacks} and \emph{patch attacks}, on driving safety of vision-based autonomous driving systems, rather than  the accuracy of deep learning models. We also investigate the reasons causing the decoupling between the detection precision of adversarial attacks and driving safety.

This study considers vision-based autonomous driving which mainly relies on stereo cameras for the task of environmental sensing. The vision-based object detectors that we consider in this paper are Stereo R-CNN~\cite{li19stereo} and DSGN~\cite{chen20dsgn}, two state-of-the-art methods in this area.

To facilitate this study, we propose an \emph{end-to-end} driving safety evaluation framework with a set of designed driving safety performance metrics, where the evaluation framework can directly take the results of the 3D object detector as input and outputs the scores of the driving safety performance metrics as the final assessment.

To implement such an evaluation framework, we are faced with two nontrivial technical challenges. First, the results of the 3D object detector only contain static information, such as position and dimension. Thus, we cannot determine which objects are moving and which are static. Second, to realistically generate a planned trajectory for the self-driving ego-vehicle, real driving constraints, such as speed limits for different road types and dynamics models for different vehicles, must be provided to the motion planning module. Considering that the driving scenarios change dynamically, we need to select appropriate real driving constraints accordingly for driving safety assessment.

To tackle the first challenge, we train a CNN-based classifier with manually labeled ground truth to categorize whether an object is moving or static. For the second challenge, we train another classifier with road type labels to classify the road type of each scenario, so as to select appropriate driving constraints.

To obtain comprehensive experiment results, we apply the aforementioned two types of adversarial attacks with different attack intensities in our evaluation framework and measure the rate that the motion planner successfully finds a trajectory, the rate of collision occurrence, and the rate that the ego-vehicle drives safely from the initial position to the goal region. In the meantime, we also measure the precision changes of the vision-based 3D object detectors when they are under attacks. By linking the impact of adversarial attacks on driving safety and on 3D object detection together, we manage to find the answers to our motivation questions. The main contributions of this paper can be summarized as follows.

\begin{itemize}
\item
We propose an \textit{end-to-end} driving safety evaluation framework that directly takes the produced results of the 3D object detector as input and outputs driving safety performance scores as the evaluation outcome. With modular design, each individual module can be easily replaced so that the framework can be adapted to evaluate the driving safety of different self-driving systems threatened by various attacks.
\item
We conduct extensive experiments and observe that the changes in object detection precision and the changes in driving safety performance metrics caused by adversarial attacks are decoupled. Therefore, the answers to our motivation questions are all no. And we also observe that DSGN is more robust than Stereo R-CNN in terms of driving safety.
\item
We investigate the reasons behind our answers to those questions. We find that the reason for the decoupling is that it is easier for perturbation attacks to mislead object detectors to detect ghost objects at roadside which cause little influence on driving safety but huge impact on detection precision. We also find out that the reason why DSGN is more robust than Stereo R-CNN is that the latter is purely based on deep neural network, while DSGN adopts the Spatial Pyramid Pooling (SPP) in its architecture which can alleviate the attack effects.
\end{itemize}

The rest of this paper is organized as follows. In Section~\ref{sec.relatedwork}, we first introduce the studies related to our work. In Section~\ref{sec.attackmodels}, we briefly introduce the attack model of the two adversarial attacks studied in this paper. In Section~\ref{sec.system}, we elaborate on our proposed end-to-end evaluation framework and the driving safety performance metrics. In Section~\ref{sec.experiments}, we present the experiment design and results. In Section~\ref{sec.ablationstudy}, we investigate the causes of our observations with an ablation study. Finally, we conclude the paper in Section~\ref{sec.conclusion}.

\vspace{-0.1in}
\section{Related Work}
\label{sec.relatedwork}

In this section, we review the related work from the perspectives of attacks on autonomous driving, vision-based 3D object detection, and motion planning.

\textbf{Attacks on Autonomous Driving.} A survey of general vulnerabilities in autonomous driving can be found in~\cite{wyglinski13security, ren19security}. Among all the vulnerabilities, the perturbation attack and the patch attack are the most dangerous threats of vision-based autonomous driving systems, since they can directly cause damages to input images.

Both the perturbation attack and the patch attack fall into the domain of adversarial attacks. The main idea of adversarial attacks is to leverage small changes in the input to trigger significant errors in the output of deep learning models. According to~\cite{yuan19adversarial}, adversarial attacks can be either universal (effective to all valid inputs) or input-specific (only effective to a specific input). There are mainly two categories of methods to achieve adversarial attacks, namely, optimization-based methods and fast gradient step method (FGSM)-based approach. Optimization-based methods can be used for either universal attacks or input-specific attacks. An example is L-BFGS proposed by Szegedy \textit{et al.}~\cite{szegedy14intriguing}. FGSM-based methods include FGSM~\cite{goodfellow15explaining} and its extensions, such as I-FGSM~\cite{kurakin17adversarial}, MI-FGSM~\cite{dong18boosting}, and PGD~\cite{madry18towards}. These methods are usually used only for input-specific attacks.

The perturbation attack and the patch attack work in different ways. The perturbation attack usually affects all pixels in an input image but the changes in pixel values are very small, while the patch attack only affects a small number of pixels but the changes in pixel value are larger. Both the attacks were studied concerning different functional modules needed in vision-based autonomous driving. For example, the perturbation attack was studied regarding sign classification in~\cite{eykholt18robust}, 2D object detection in~\cite{lu17adversarial}, semantic segmentation in~\cite{metzen17universal, xiao18characterizing}, and monocular depth estimation in~\cite{zhang20adversarial, mathew20monocular}, while the patch attack was studied regarding lane keeping in~\cite{zhou18deep}, optical flow estimation in~\cite{ranjan19attacking}, 2D object detection in~\cite{song18physical, chen18shapeshifter}, and monocular depth estimation in~\cite{mathew20monocular}. None of these studies, however, focus directly on the attacks' impact on driving behavior and driving safety of autonomous vehicles.

\textbf{Vision-Based 3D Object Detection.} Vision-based 3D object detection provides a more budget-friendly approach to perform object detection in 3D space by mainly leveraging stereo cameras instead of expensive LiDARs. It is the core of vision-based autonomous driving. Traditional approaches, e.g., 3DOP~\cite{chen183d} and Pseudo-LiDAR~\cite{wang19pseudo}, first generate a pseudo point cloud with depth estimation and then perform 3D object detection with similar methods used in LiDAR-based 3D object detection. As a result, they are usually not comparable to LiDAR-based methods in terms of accuracy and efficiency. Different from traditional approaches, Stereo R-CNN~\cite{li19stereo} and DSGN~\cite{chen20dsgn} are the two leading methods in this area. The network of Stereo R-CNN consists of a Region Proposal Network (RPN) and a regression part. The 2D bounding box candidates generated by the RPN are fed to the regression part where keypoints of 3D bounding boxes are predicted. The network of DSGN includes a single-stage pipeline which exacts pixel-level features for stereo matching and high-level features for object recognition. Both methods can achieve comparable performance to LiDAR-based methods.

\textbf{Motion Planning.} Motion planning is a key task for autonomous vehicles. Given an initial vehicle state, a goal state region, a cost function, and vehicle dynamics, motion planning finds collision-free trajectories. Currently, sampling-based motion planning algorithms are the mainstream methods. They can be viewed as a discrete planner, such as RRT~\cite{lavalle98rapidly}, greedy BFS, and A*~\cite{hart68a}, in combination with a C-space sampling scheme.

\vspace{-0.05in}
\section{Attack Models}
\label{sec.attackmodels}

\begin{figure*}[!t]
    \centering
    \includegraphics[width=1.0\textwidth]{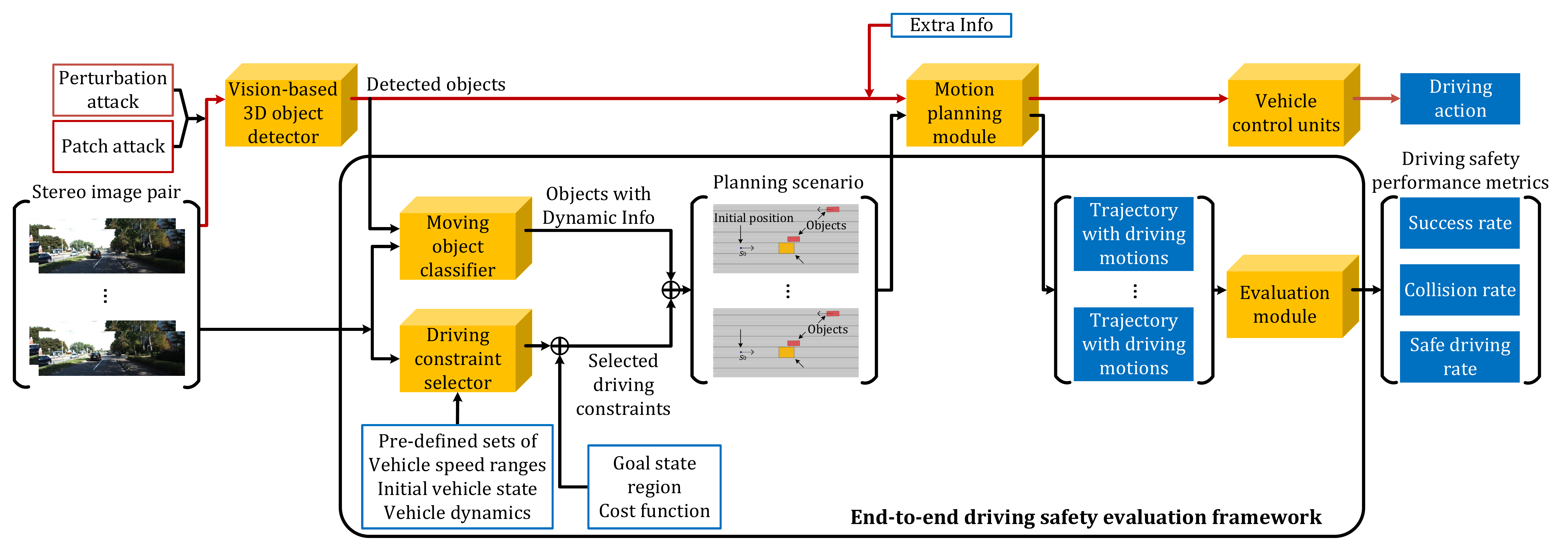}
    \caption{The end-to-end driving safety evaluation framework.}
    \label{fig:system_model}
\end{figure*}

We assess the impact of two types of adversarial attacks, perturbation attacks and patch attacks, on driving safety. Here, we briefly introduce the attack models.

\textbf{Perturbation Attack.} The goal of this type of adversarial attacks is to make the deep learning model dysfunctional by adding small changes to each pixel in the image that are imperceptible to human eyes. With prior knowledge of the deep learning model, attackers can launch perturbation attacks by tapping into the self-driving system and perturbing camera images. We consider the method of PGD~\cite{madry18towards} to achieve input-specific attacks. Consider a perturbation $\delta^{\text{per}}$ and an image pair $(I_{l}, I_{r})$, where $\delta^{\text{per}}$ has the same dimension as $I_{l}$ and $I_{r}$. Let the initial perturbed image pair $(\tilde{I}^{\text{per}}_{l, 0}, \tilde{I}^{\text{per}}_{r, 0}) = (I_{l}, I_{r})$. The attack is carried out by updating the perturbation using the projected loss gradient of the 3D object detector through multiple iterations with
\begin{equation}\label{eq:eq_2}
\delta^{\text{per}}_{n} = \text{Clip}_{\epsilon}\{\alpha \times \text{sign}(\nabla_{(I_{l}, I_{r})}L(O_{\theta}(\tilde{I}^{\text{per}}_{l, n}, \tilde{I}^{\text{per}}_{r, n}), b^{\text{true}}))\}
\end{equation}
and
\begin{equation}\label{eq:eq_3}
(\tilde{I}^{\text{per}}_{l, n+1}, \tilde{I}^{\text{per}}_{r, n+1}) = (\tilde{I}^{\text{per}}_{l, n} + \delta^{\text{per}}_{n}, \tilde{I}^{\text{per}}_{r, n} + \delta^{\text{per}}_{n})
\end{equation}
where $\text{Clip}_{\epsilon}\{\cdot\}$ ensures that the value is within $[-\epsilon, \epsilon]$, $\alpha$ is the parameter that controls the attack intensity, $\text{sign}(\cdot)$ denotes the sign function, $O_{\theta}(\cdot, \cdot)$ represents the vision-based 3D object detector parametrized by $\theta$, $L(\cdot, \cdot)$ is the loss function of $O_{\theta}(\cdot, \cdot)$, $b^{\text{true}}$ is the ground truth label paired with $(I_{l}, I_{r})$, and $0 \leqslant n \leqslant N-1$. For convenience, we denote the perturbation attack as $(\tilde{I}^{\text{per}}_{l}, \tilde{I}^{\text{per}}_{r}) = A^{\text{per}}(I_{l}, I_{r}, b^{\text{true}}, \epsilon, \alpha, N)$.

\textbf{Patch Attack.} The patch attack is designed to model the real-world poster-printing attack  in~\cite{eykholt18robust}. In the context of vision-based 3D object detection, a patch attack is launched to mislead the detector so that it detects ghost objects by including the patch in the view of the image. With prior knowledge of the deep learning model, attackers can train a patch offline, print it out, and put the physical patch inside the view of cameras to launch the attack. For example, the attacker can place the patch at the roadside where the vision-based self-driving car passes by. Since a real-world 3D point appears at different positions in two stereo images, we consider a patch $\delta^{\text{pat}}$ that is pasted onto $I_{l}$ at $loc_{l}$ and onto $I_{r}$ at $loc_{r}$, where $(loc_{l}, loc_{r}) \in \mathcal{L}$, $\mathcal{L}$ represents a set of random position pairs. Let $\lambda_{loc_{l}, loc_{r}} \in \Lambda$ be the disparity between $loc_{l}$ and $loc_{r}$, where $\Lambda$ denotes a set of valid disparities in the physical world. Let $\tau \in \mathcal{T}$ be a transformation that can be applied to $\delta^{\text{pat}}$, where $\mathcal{T}$ includes rotations. Then, the patched image pair can be represented as $(\tilde{I}^{\text{pat}}_{l}, \tilde{I}^{\text{pat}}_{r}) = A^{\text{pat}}(I_{l}, I_{r}, \delta^{\text{pat}}, loc_{l}, loc_{r}, \tau)$. To implement this attack, the patch is optimized as
\begin{equation}\label{eq:eq_4}
\mathop{\text{argmin}}_{\delta^{\text{pat}}} \mathbb{E}_{(I_{l}, I_{r}) \sim \mathcal{I}, (loc_{l}, loc_{r}) \sim \mathcal{L}, \tau \sim \mathcal{T}} L(O_{\theta}(\tilde{I}^{\text{pat}}_{l}, \tilde{I}^{\text{pat}}_{r}), b^{*}),
\end{equation}
where $b^{*}$ denotes the predefined 3D bounding boxes used for misleading the object detector and serves as the optimization target here.

\vspace{-0.05in}
\section{End-to-end Driving Safety Evaluation Framework}
\label{sec.system}

As discussed in Section~\ref{sec.relatedwork}, many previous studies only showed that deep learning models of autonomous driving can be compromised by adversarial attacks, but they did not systematically assess the attack impact on driving safety. Our goal is to answer the questions raised in Section~\ref{sec.introduction} by investigating the impact of perturbation attacks and patch attacks on driving safety of vision-based autonomous vehicles. This investigation considers not only the performance of the attacked deep learning models but also their impact on the overall safety, which is a combined effect of different functional modules involved in autonomous driving.

To this end, we design an end-to-end driving safety evaluation framework. \emph{End-to-end} means that our system directly takes 3D object detection results as input and outputs the driving safety scores. Moreover, our evaluation framework adopts a modular design, so that each module can be easily replaced with other methods to assess the driving safety of different autonomous driving systems. Note that  the existing simulators, such as Baidu Apollo~\cite{apollo19perception} and CARLA~\cite{dosovitskiy17carla}, either have a low level of customization or are not compatible with real autonomous driving datasets. Therefore, we implement our own evaluation framework with real autonomous driving dataset to evaluate the impact of adversarial attacks on driving safety. In this section, we first introduce our evaluation framework model for vision-based autonomous driving and the driving safety metrics, then elaborate on the framework implementation details.

\vspace{-0.05in}
\subsection{Framework Model}

Our evaluation framework works along with the data flow of vision-based autonomous driving systems. In Figure~\ref{fig:system_model}, the black lines represent the data flow of our evaluation framework, while the red lines are for the data flow of the autonomous driving system. Usually inside the vision-based autonomous vehicle, a pair of stereo images $(I_{l}, I_{r} \in \mathbb{R}^{h \times w \times 3})$ is first fed as the input to the 3D object detection module $O_{\theta}(\cdot, \cdot)$, which is parameterized by $\theta$ and generates detected objects in 3D bounding boxes $b$ (denoted as $b=O_{\theta}(I_{l}, I_{r})$) as the output. Next, the bounding boxes $b$ along with some extra information are passed to the the motion planning module $M(\cdot)$. At last, the vehicle control units execute the driving motion orders from the motion planning module. As depicted in Figure~\ref{fig:system_model}, our proposed end-to-end driving safety evaluation framework directly takes the detected objects of $O_{\theta}(\cdot, \cdot)$ as input, uses the same motion planning module of the autonomous vehicle, and outputs scores for driving safety metrics. This modular design makes it possible for our evaluation framework to be adopted by other self-driving systems which have different implementation of the aforementioned modules.

As described in Section~\ref{sec.introduction},  two main technical challenges need to be addressed after the object detection results are fed into our evaluation framework. First, the 3D bounding boxes $b$ as the object detection results only contain static information, i.e., object category, box dimensions, box center position in 3D space, and the confidence score. Base on the static information of one frame of data, we cannot distinguish between moving objects and static objects. However, the subsequent motion planning module requires dynamic information of objects as part of its input. Second, to realistically produce planned trajectory for the self-driving vehicle, driving constraints, including speed limits for different road types and dynamics constraints for different moving vehicles (acceleration, jerk, energy, etc.), must be considered to comply with the real driving scenarios. In addition, as the real driving scenarios can change dynamically, we must choose appropriate real driving constraints accordingly for driving safety evaluation.

\begin{figure*}[!t]
\centering

\subfigure[Clean image input.]{
    \includegraphics[width=0.35\textwidth]{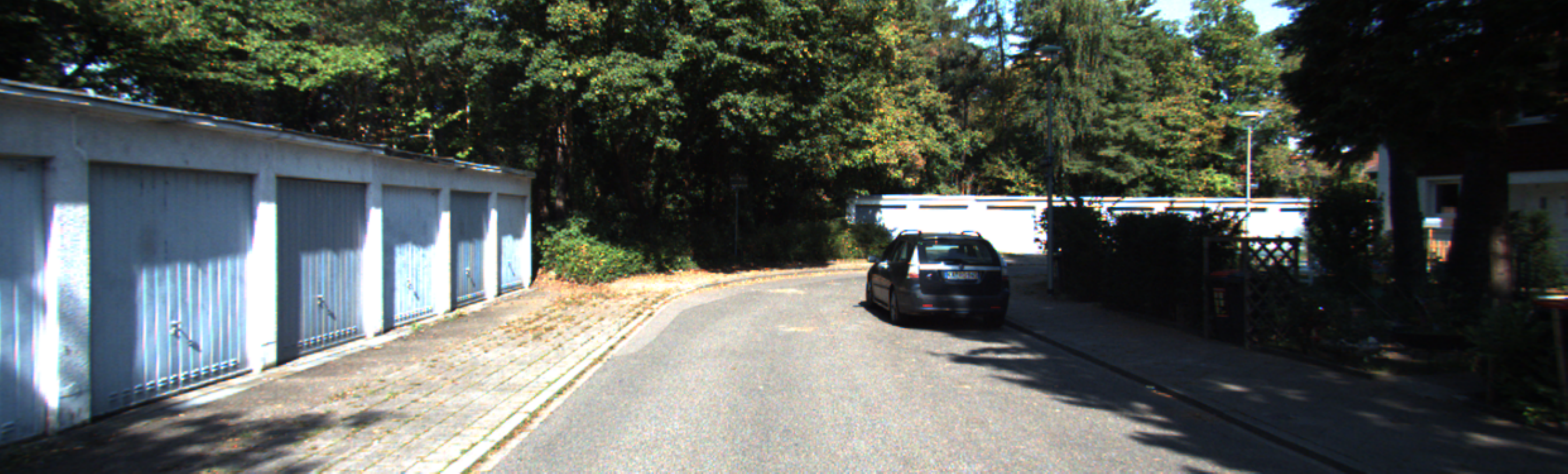}
}
\subfigure[Ground truth of object detection.]{
    \includegraphics[width=0.35\textwidth]{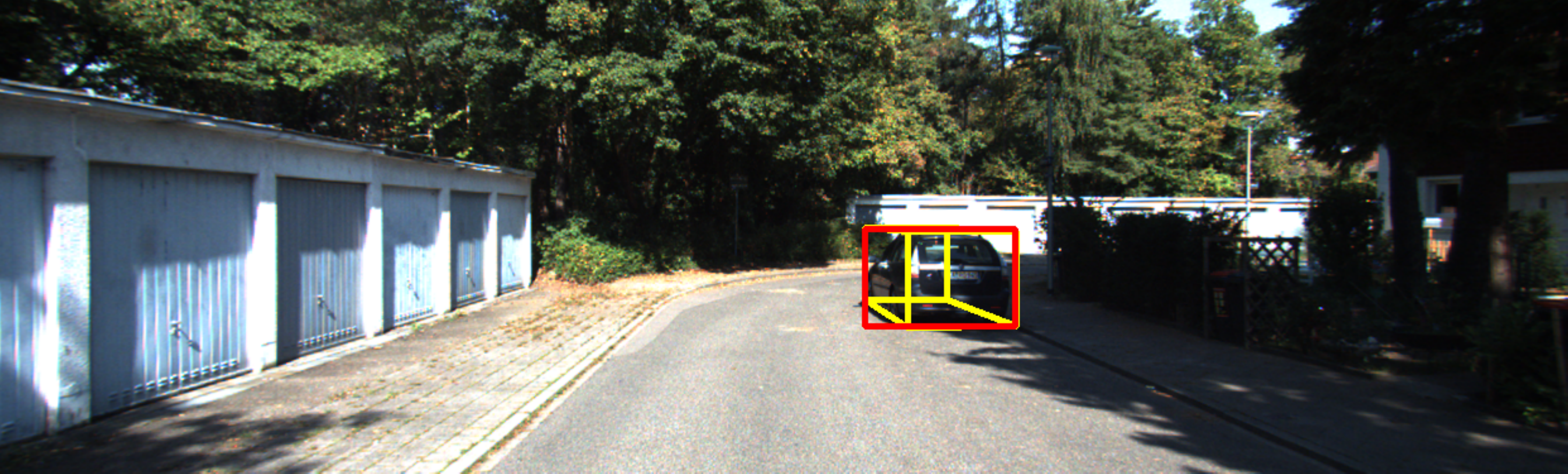}
} \\
\subfigure[Detection results of DSGN without attack.]{
    \includegraphics[width=0.35\textwidth]{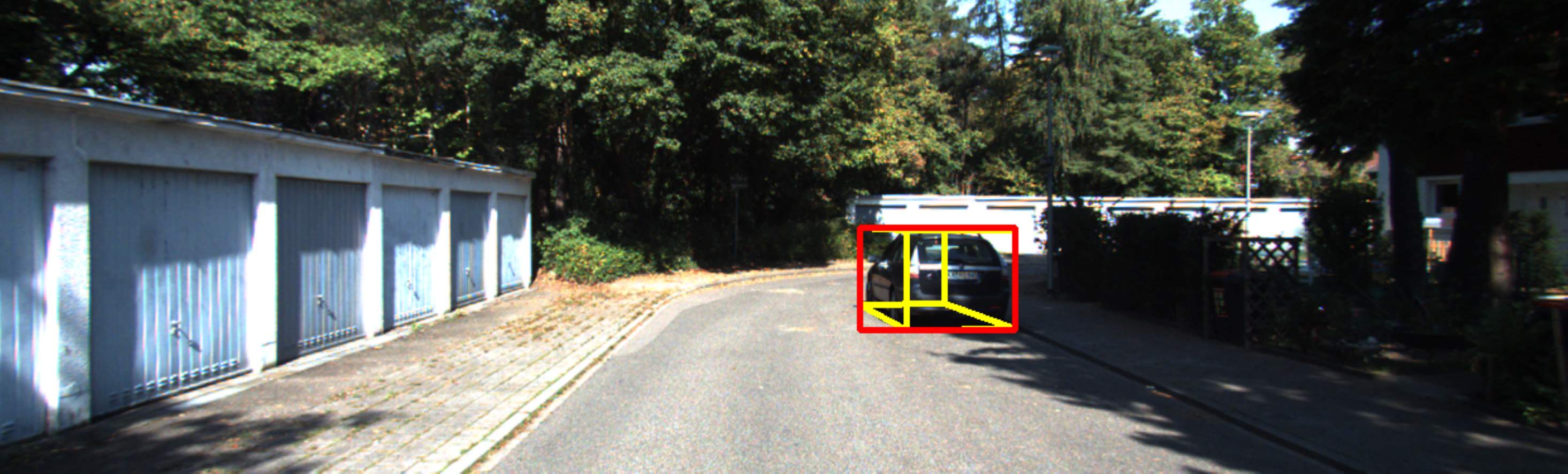}
}
\subfigure[Detection results of DSGN under attack.]{
    \includegraphics[width=0.35\textwidth]{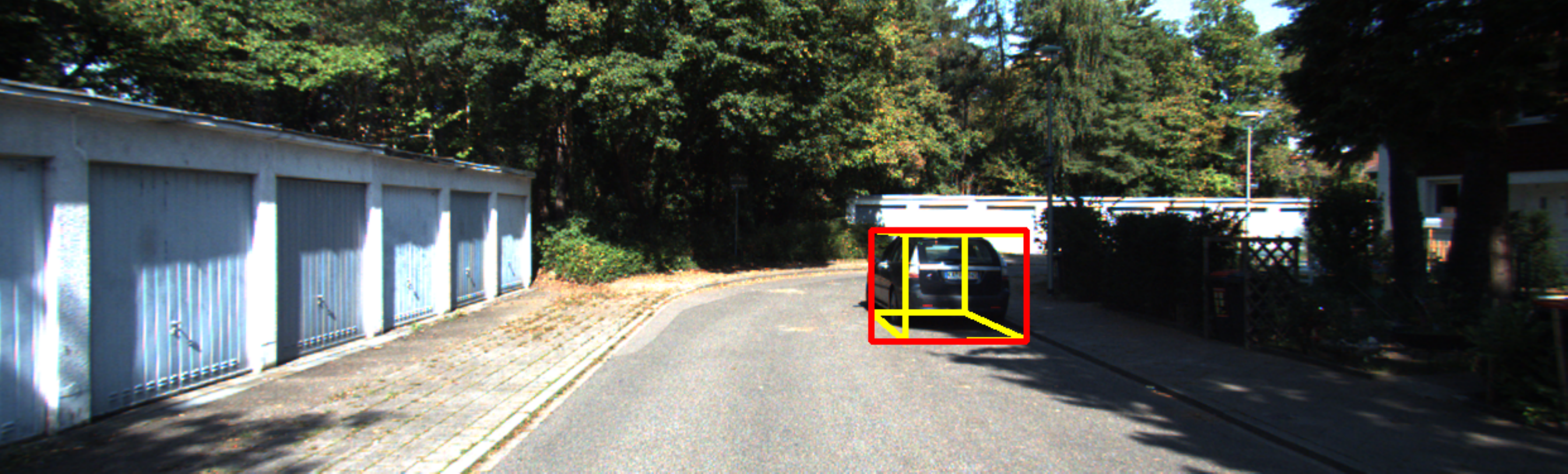}
} \\
\subfigure[Detection results of Stereo R-CNN without attack.]{
    \includegraphics[width=0.35\textwidth]{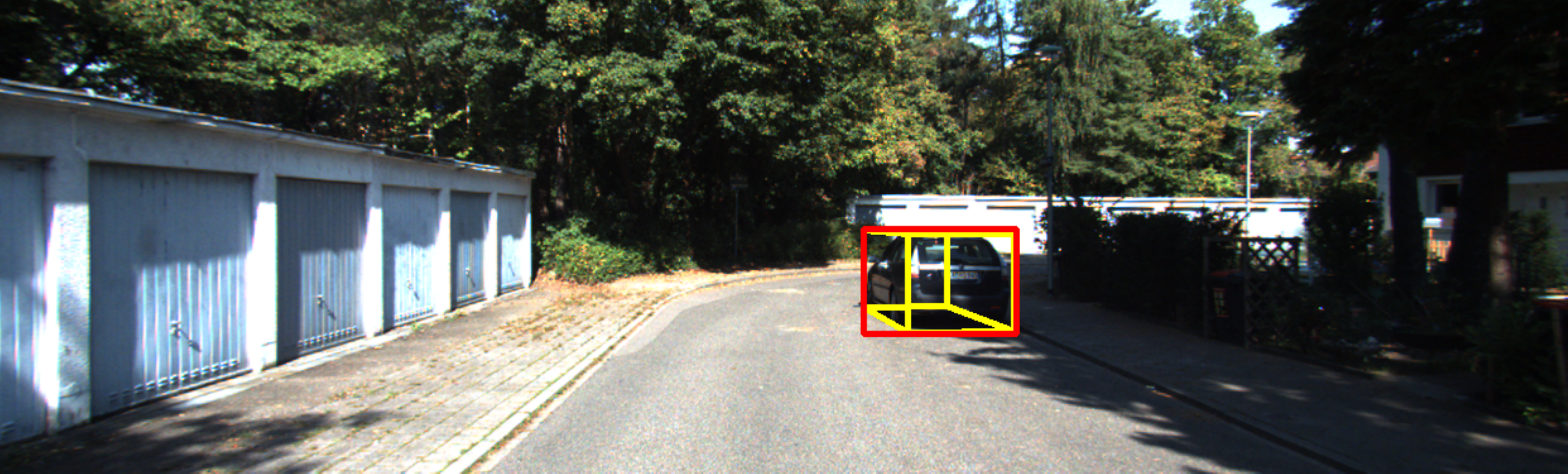}
}
\subfigure[Detection results of Stereo R-CNN under attack.]{
    \includegraphics[width=0.35\textwidth]{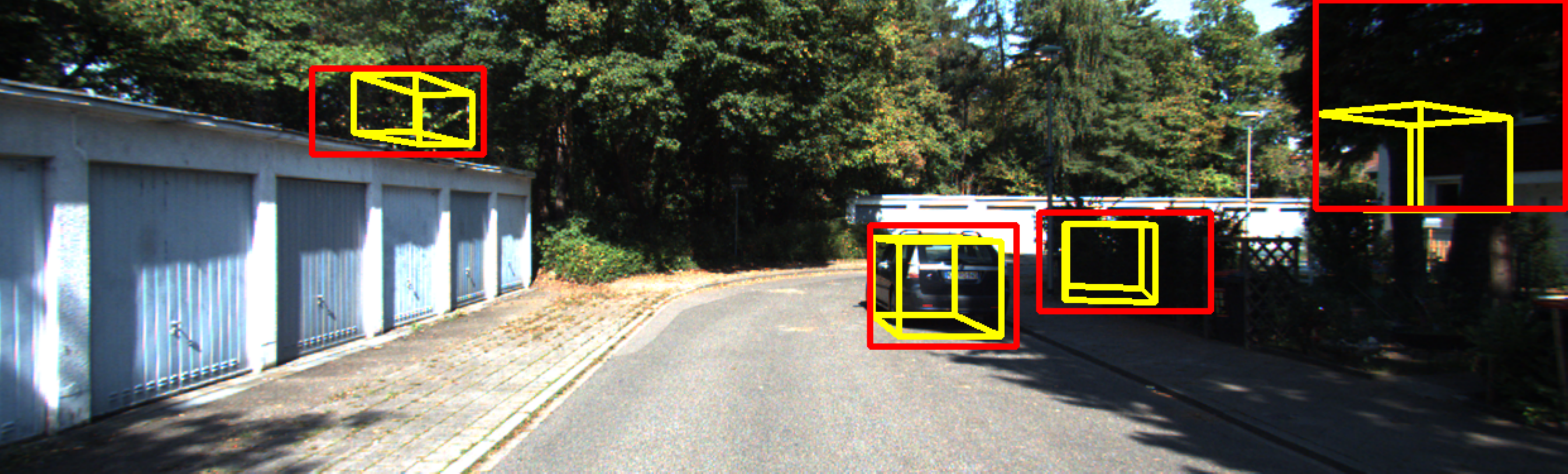}
} \\
\subfigure[Clean image input.]{
    \includegraphics[width=0.35\textwidth]{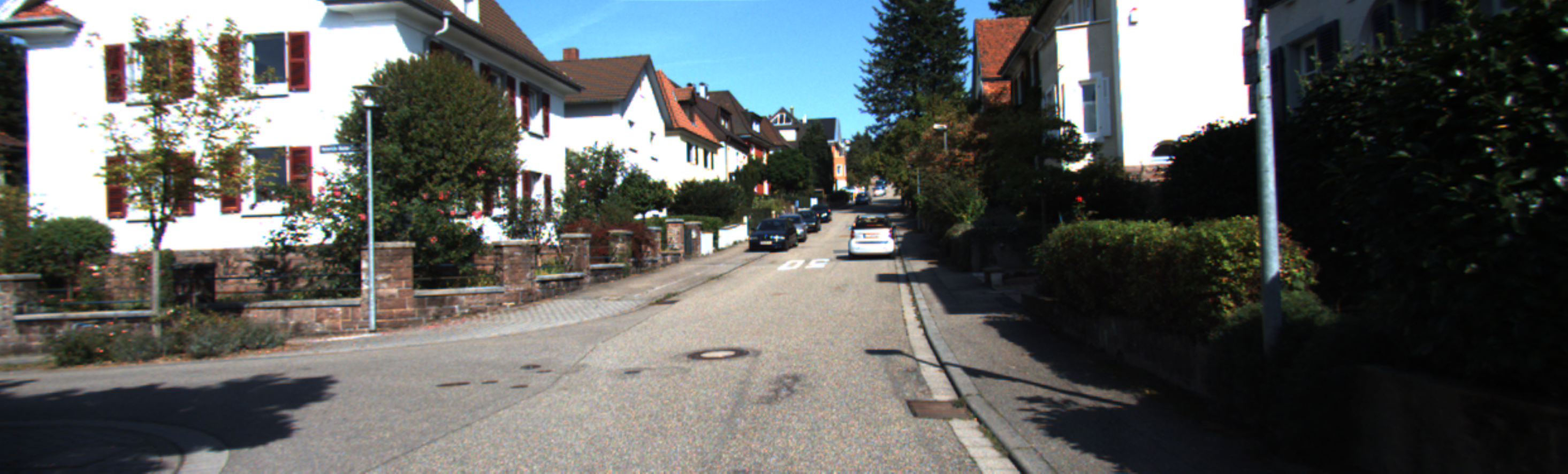}
}
\subfigure[Ground truth of object detection.]{
    \includegraphics[width=0.35\textwidth]{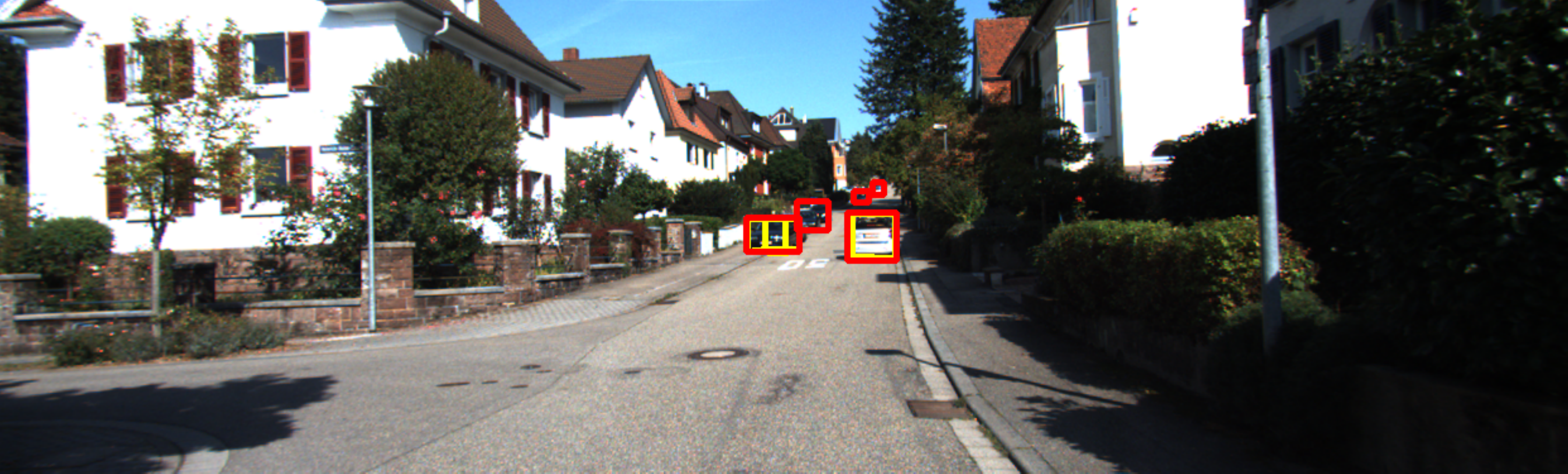}
} \\
\subfigure[Detection results of DSGN without attack.]{
    \includegraphics[width=0.35\textwidth]{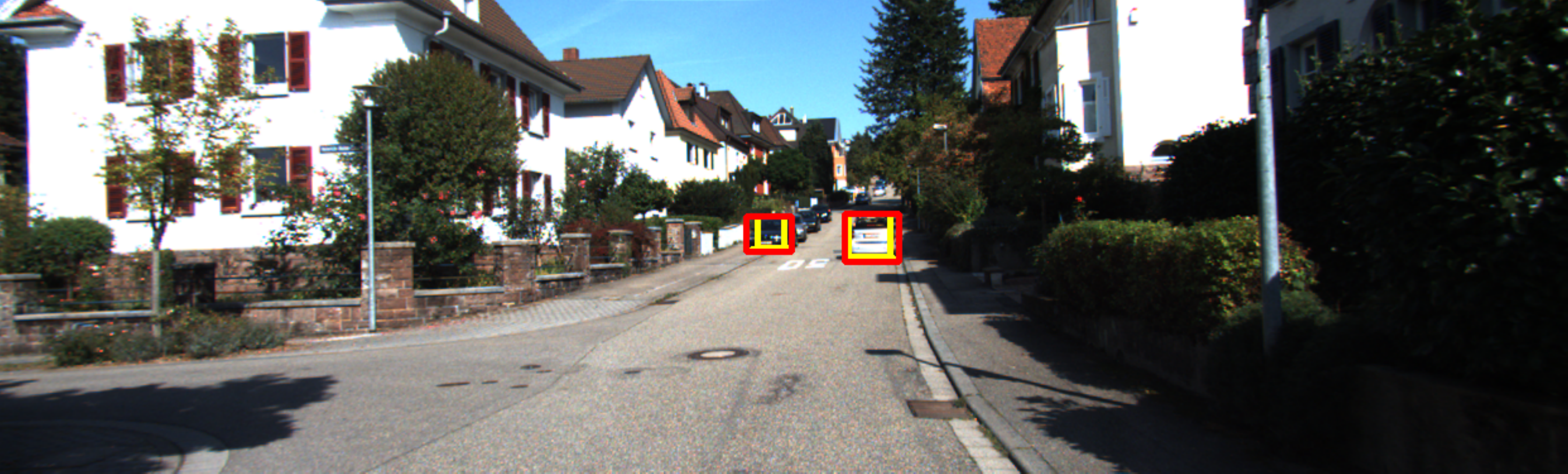}
}
\subfigure[Detection results of DSGN under attack.]{
    \includegraphics[width=0.35\textwidth]{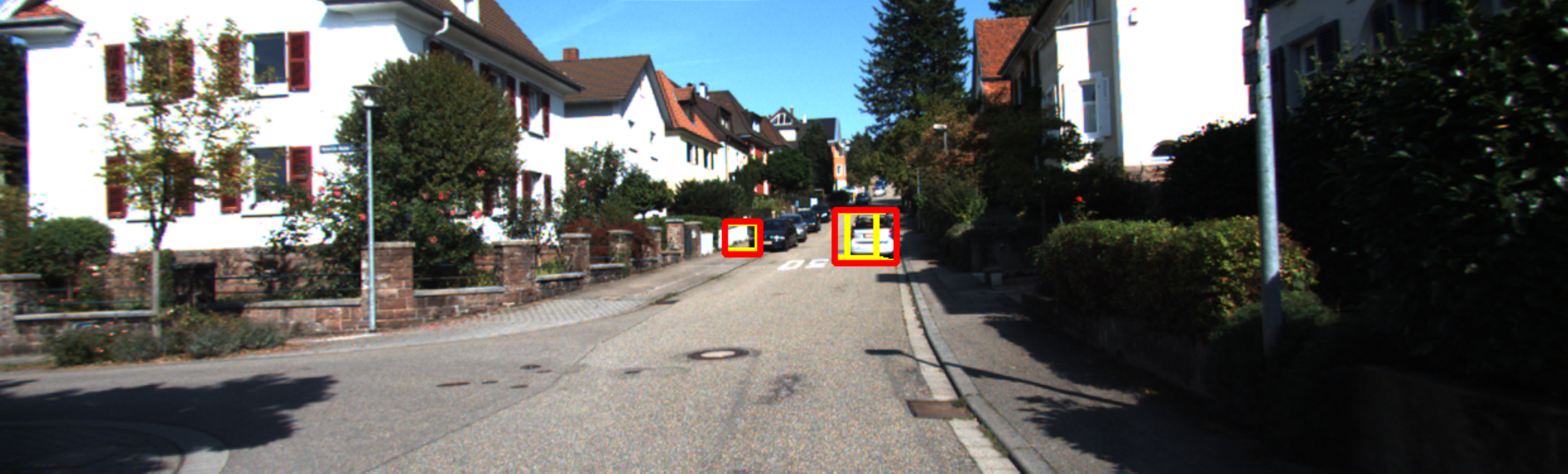}
} \\
\subfigure[Detection results of Stereo R-CNN without attack.]{
    \includegraphics[width=0.35\textwidth]{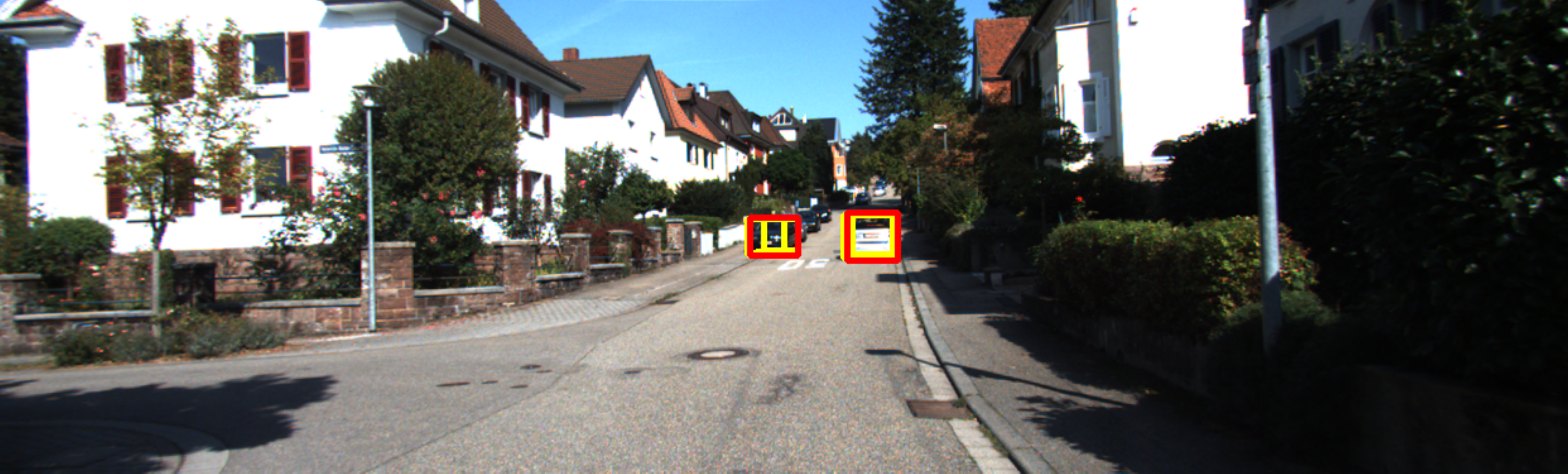}
}
\subfigure[Detection results of Stereo R-CNN under attack.]{
    \includegraphics[width=0.35\textwidth]{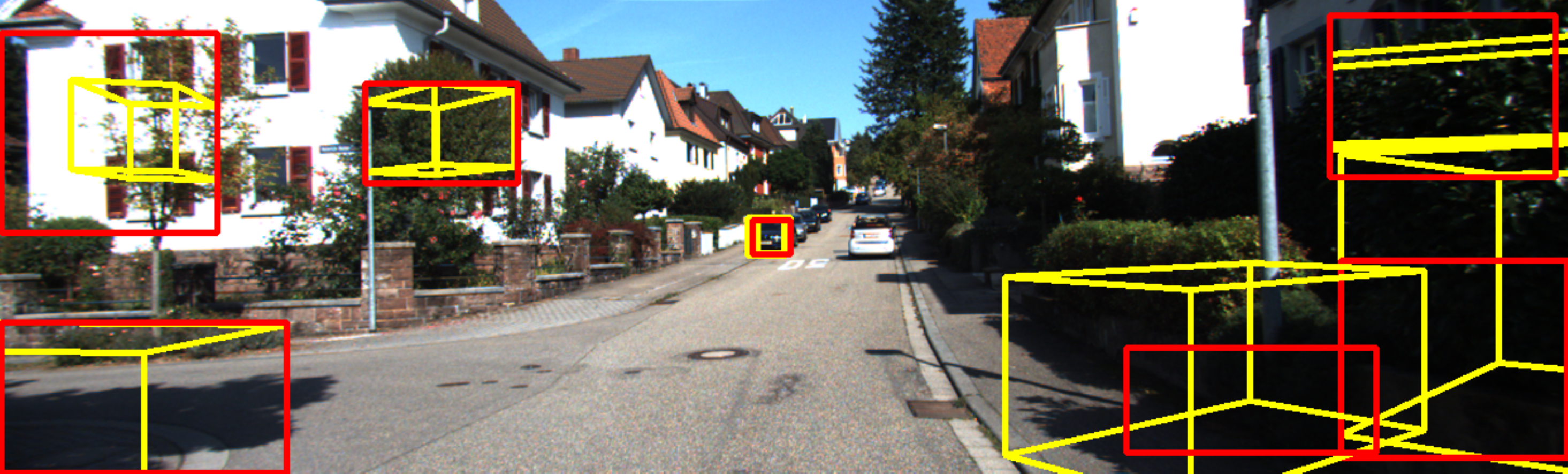}
} \\
\caption{When there is no attack, both Stereo R-CNN and DSGN can detect objects accurately as shown in (c), (e), (i), and (k). When the perturbation attack is launched, the two models produce erroneous object detection results including inaccurate detection of real objects in (d), (j), and false detection of ghost objects in (f), (l).}
\label{fig:plot_perturbation}
\end{figure*}

To tackle the first challenge, we train a CNN-based moving object classifier $C(\cdot, \cdot, \cdot)$ to distinguish between dynamic and static objects by leveraging continuous frames of image data. We manually label each object with the ground truth indicating whether this object is moving or not. By doing this, we associate the object detection results with dynamic information. We denote this process as $\vec{b}=C(b, I_{l}, I_{r})$.

To address the second challenge, we train another CNN-based model $S(\cdot, \cdot)$ with road type labels as the driving constraint selector, so that it can classify the road type of driving scenarios and select proper real driving constraints for the evaluation. We denote this part as $(s_{0}, r, d)=S(I_{l}, I_{r})$, where $s_{0}$ is the initial vehicle state, $r$ is the allowed speed range, and $d$ represents the vehicle dynamics. In this paper, we define a vehicle state $s := (p, v, \varphi, \omega)$ as a combination of position $p$, velocity $v$, orientation $\varphi$, and steering angle $\omega$ at a specific moment. Note that both the two aforementioned models, $C(\cdot, \cdot, \cdot)$ and $S(\cdot, \cdot)$, are trained on KITTI raw dataset~\cite{geiger13vision}.

Then, together with goal region $g$ and cost function $c$, we combine the processed results of both the moving object classifier and the driving constraint selector to form a planning scenario. After that, the scenario is fed to the motion planning module $M(\cdot)$ that outputs a temporal sequence of planned vehicle states $\{s_{t}\}$ (a trajectory with planned driving motions), which is denoted as $\{s_{t}\} = M(\vec{b}, s_{0}, r, g, c, d)$, where $1 \leqslant t \leqslant T$.

The final assessment of driving safety is conducted by the evaluation module based on processing a large number of driving scenarios. Specifically, the evaluation module incorporates the planned trajectory into the planning scenario and detects collision for each driving scenario in the dataset. Then, it generates driving safety performance scores based on all detected collisions. Note that we refer to a collision as the physical contact of objects. In this paper, we evaluate driving safety on the KITTI object detection dataset~\cite{geiger12are}.

Next, we introduce the driving safety performance metrics and present the details of the framework implementation.

\subsection{Driving Safety Performance Metrics}

To evaluate the driving safety of the vision-based autonomous driving system in a quantitative manner, we define a set of driving safety performance metrics as follows.

\begin{itemize}

\item
\emph{Successful planning rate}. In some scenarios, the motion planning module may not be able to generate a trajectory solution, which imposes a risk in driving safety. Thus, we define the successful planning rate as $m_{suc}=\frac{k_{trj}}{k_{dts}}$, where $k_{dts}$ is the total number of scenarios in a dataset, and $k_{trj}$ is the number of scenarios in that dataset where a trajectory can be successfully generated, no matter whether it is collision-free or not. For the sake of simplicity, this metric is referred to as \textit{the success rate}.

\item
\emph{Collision rate}. We define the collision rate, $m_{cls}$, as the percentage of scenarios in all successfully planned trajectories where a collision occurs. Let $m_{cls}=\frac{k_{cls}}{k_{trj}}$, where $k_{cls}$ is the number of scenarios with collision occurrence. Collision rate approximately reflects the collision probability under different levels of threats.

\item
\emph{Safe driving rate}. The safe driving rate, $m_{saf}$, is defined as the percentage of scenarios in a dataset where a collision-free trajectory can be produced by the motion planning module. We denote it as $m_{saf}=\frac{k_{trj}-k_{cls}}{k_{dts}}=m_{suc}-\frac{k_{cls}}{k_{trj}}\frac{k_{trj}}{k_{dts}}=m_{suc}-m_{cls}m_{suc}=(1-m_{cls})m_{suc}$. 
\end{itemize}

In this paper, we only focus on fatal driving risks when referring to the driving safety. By measuring successful planning rate and collision rate, we capture the two most risky driving scenarios in autonomous driving, i.e., the failure of path planning and collision.

Note that successful planning rate and collision rate are also common performance metrics measuring the quality of motion planning.
Furthermore, safe driving rate is jointly determined by both successful planning rate and collision rate, which is a more direct measure of driving safety.

\begin{figure*}[!t]
    \centering
    
    \subfigure[DSGN, $\alpha=0.4$]{
        \includegraphics[width=0.19\textwidth]{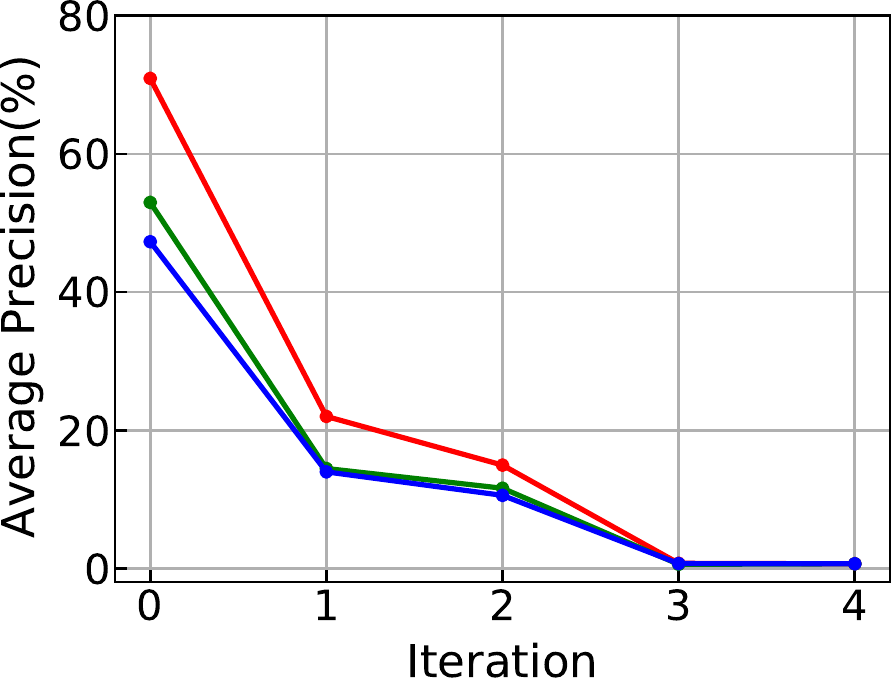}
    }\hspace{18pt}
    \subfigure[Stereo R-CNN, $\alpha=0.4$]{
        \includegraphics[width=0.19\textwidth]{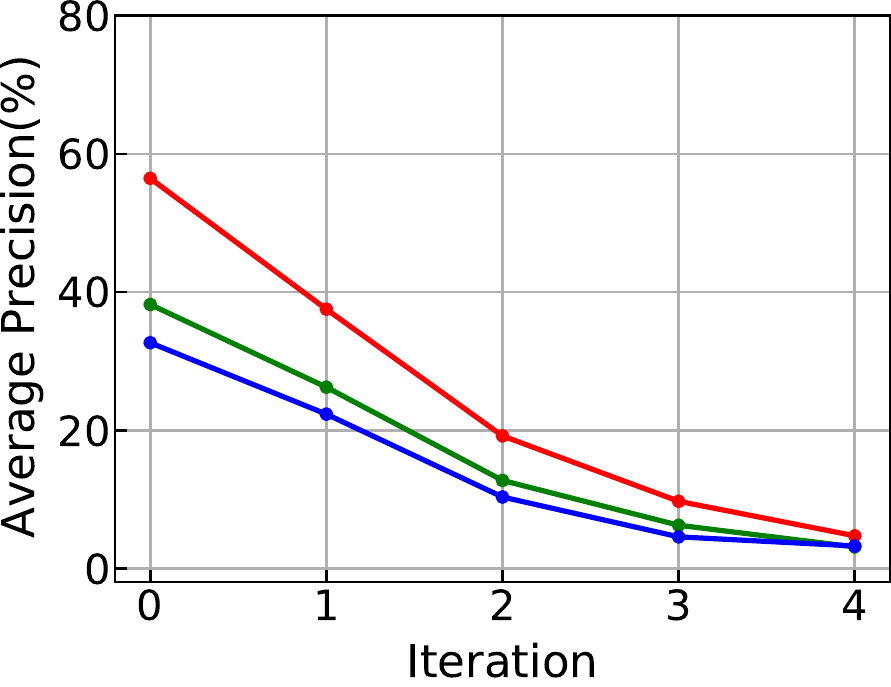}
    }\hspace{18pt}
    \subfigure[DSGN, $\alpha=1$]{
        \includegraphics[width=0.19\textwidth]{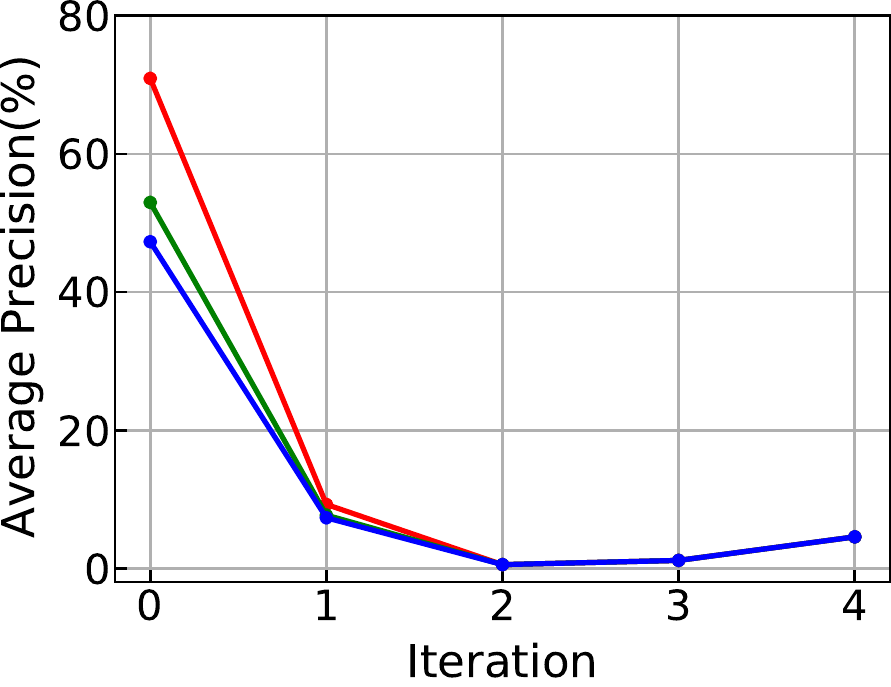}
    }\hspace{18pt}
    \subfigure[Stereo R-CNN, $\alpha=1$]{
        \includegraphics[width=0.19\textwidth]{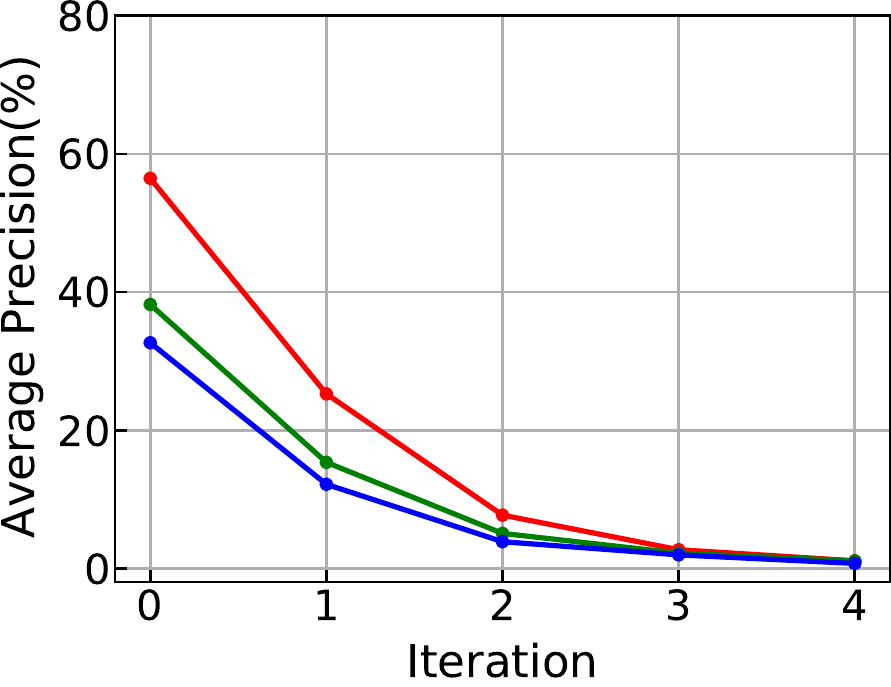}
    } \\
    \includegraphics[width=0.3\textwidth]{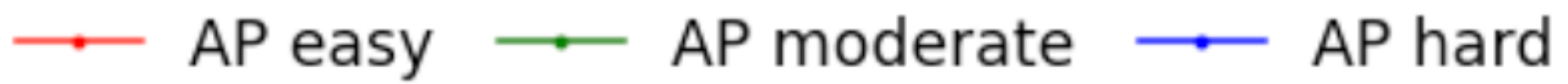}
    
    \caption{Average precision for 3D object detection under the perturbation attack.}
    \label{fig:ap_perturbation}
\end{figure*}

\begin{figure*}[!t]
    \centering
    
    \text{Changing to left lane}\par\medskip
    \begin{minipage}[t]{\textwidth}
    \centering 
    \includegraphics[width=0.22\textwidth]{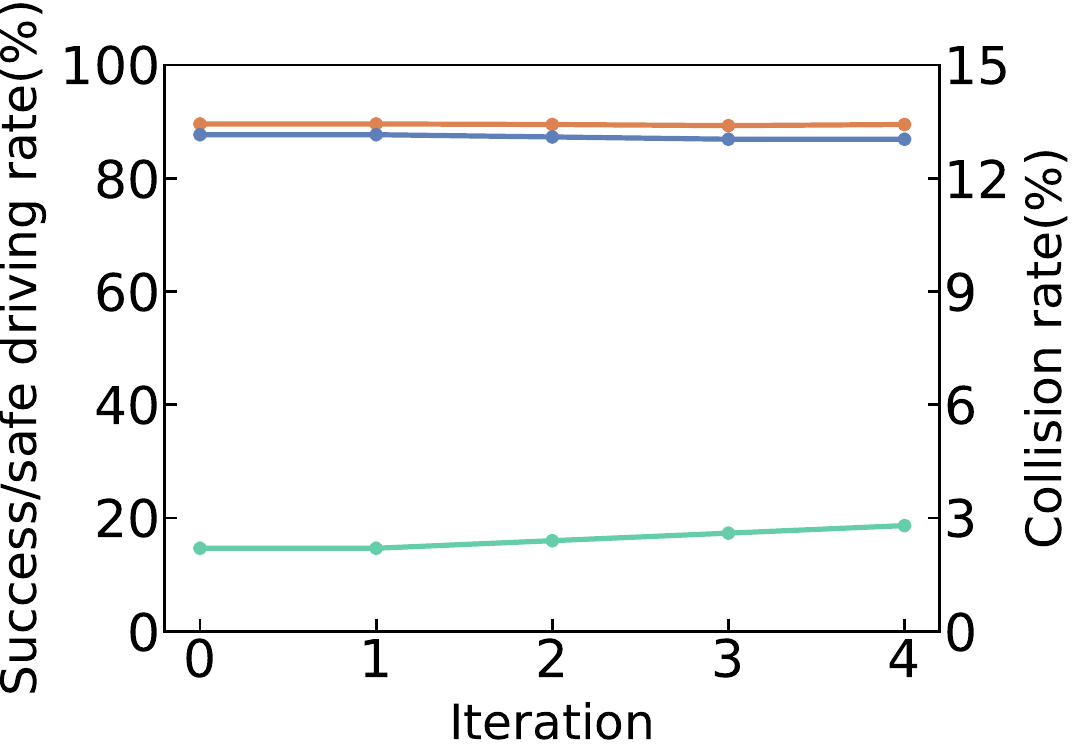}\hspace{6pt}
    \includegraphics[width=0.22\textwidth]{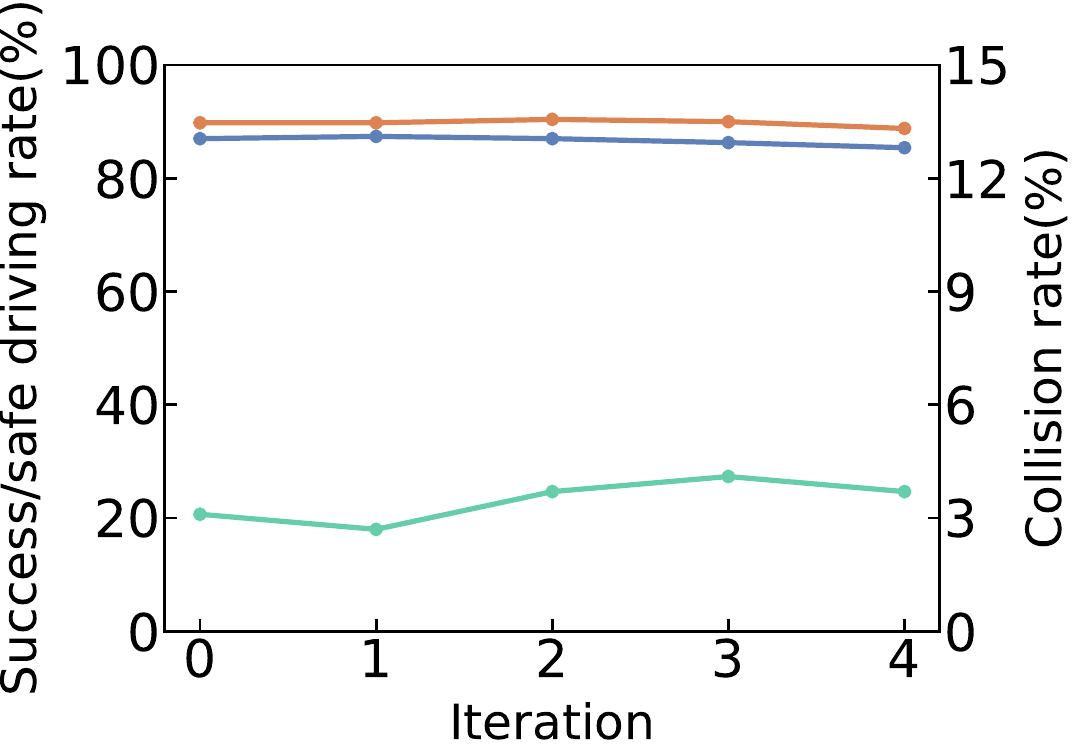}\hspace{6pt}
    \includegraphics[width=0.22\textwidth]{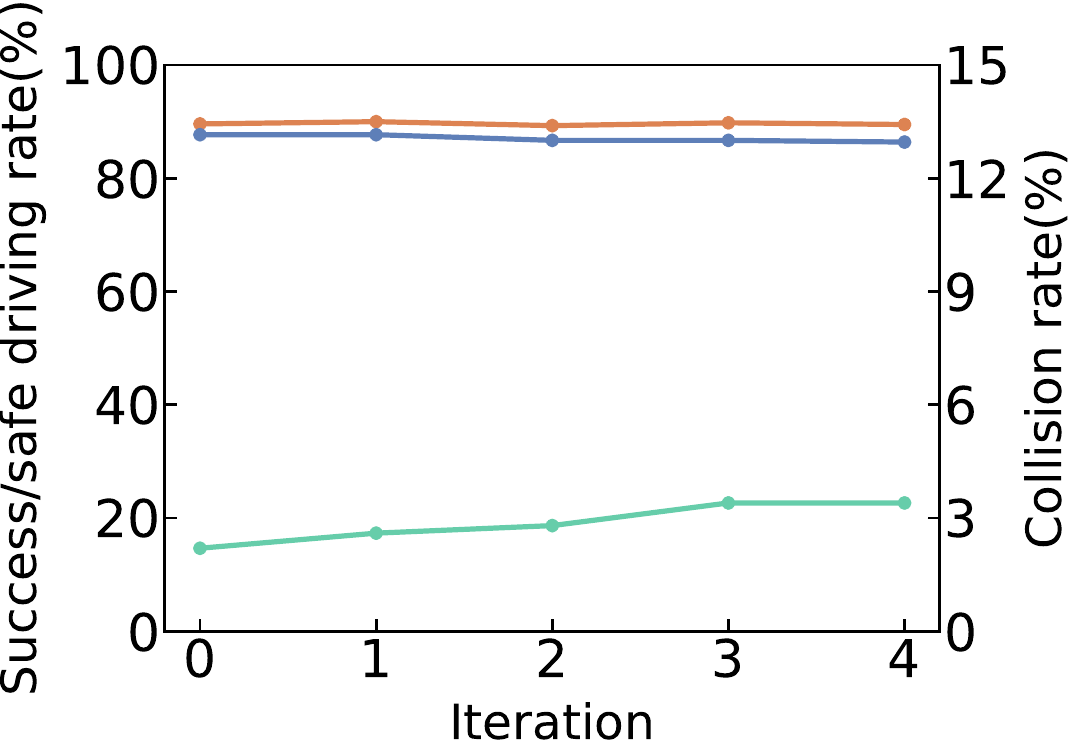}\hspace{6pt}
    \includegraphics[width=0.22\textwidth]{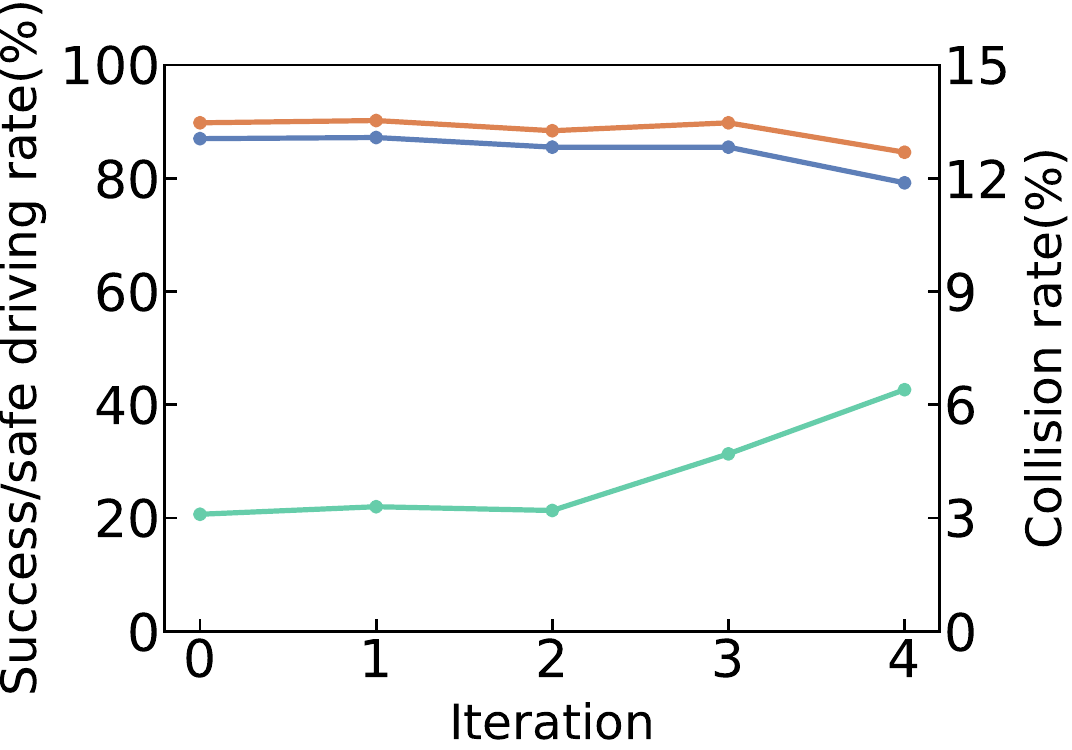}
    \end{minipage}
    
    \text{Keeping lane}\par\medskip
    \begin{minipage}[t]{\textwidth}
    \centering 
    \includegraphics[width=0.22\textwidth]{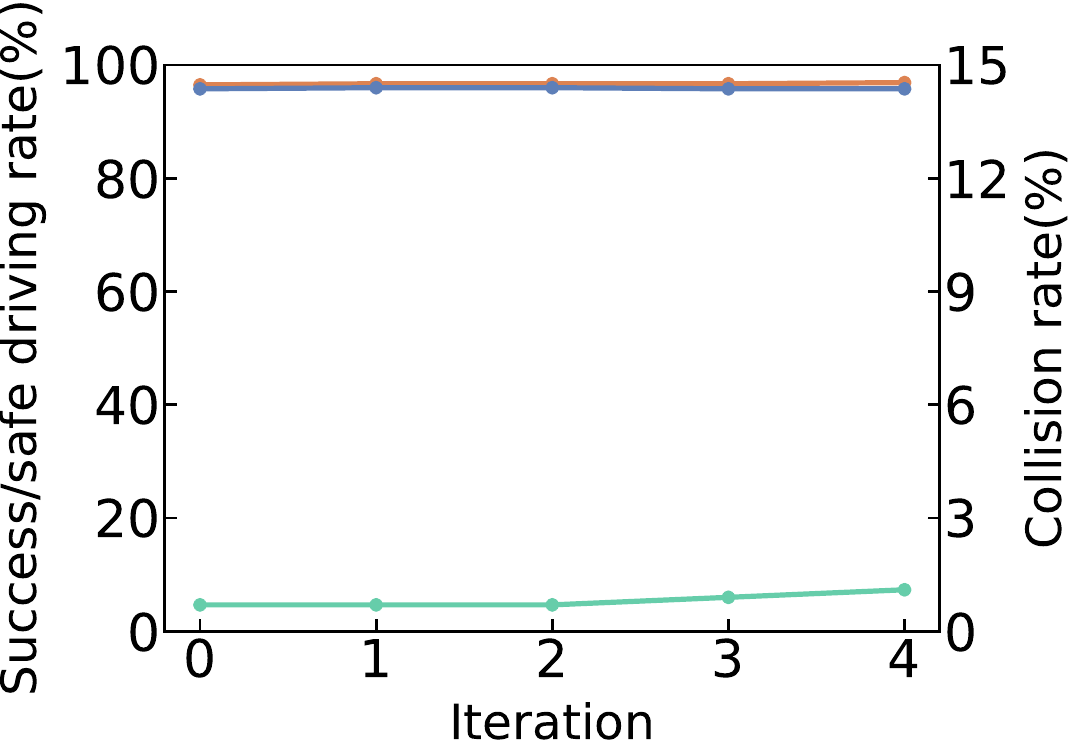}\hspace{6pt}
    \includegraphics[width=0.22\textwidth]{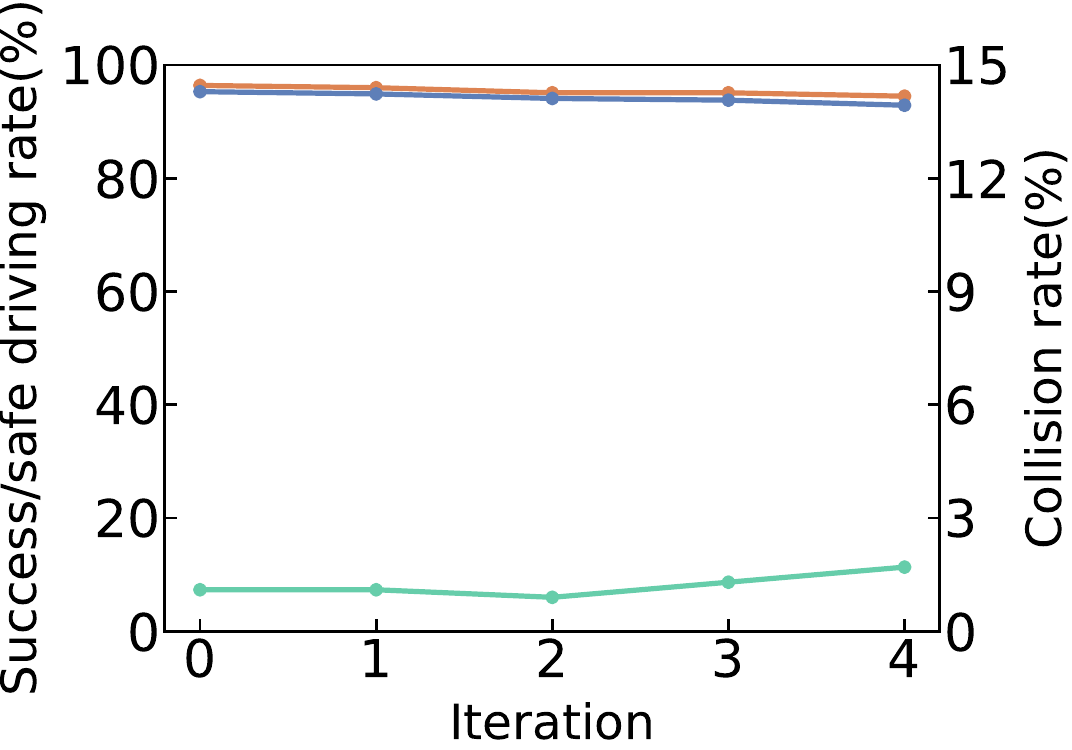}\hspace{6pt}
    \includegraphics[width=0.22\textwidth]{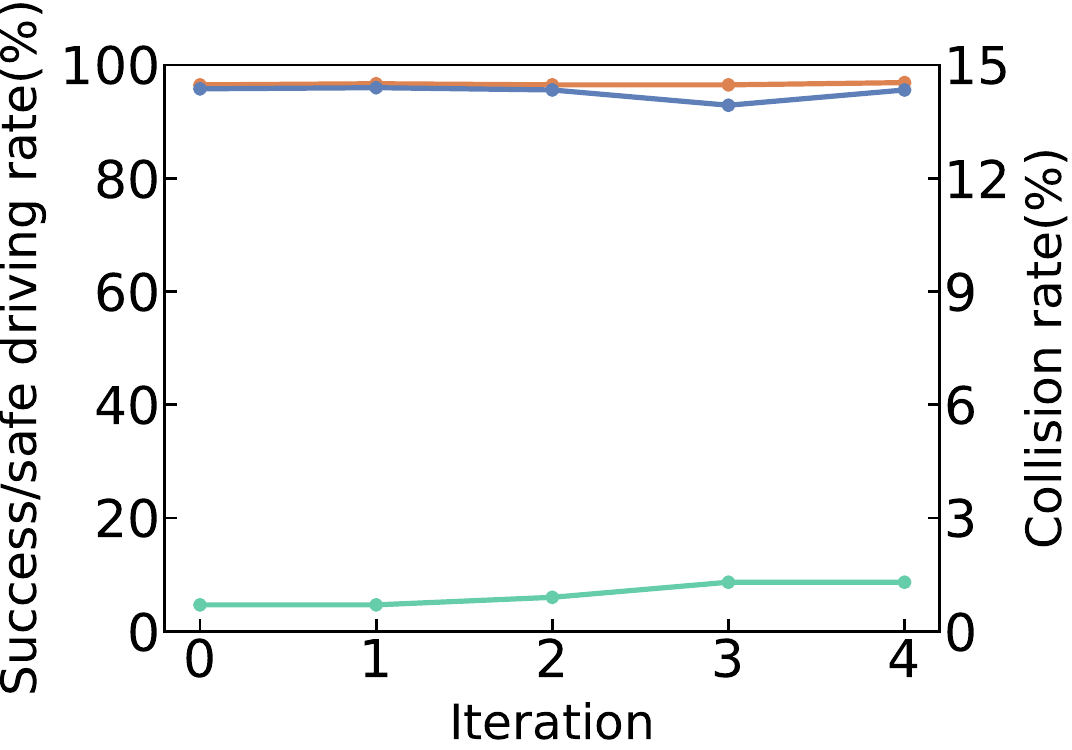}\hspace{6pt}
    \includegraphics[width=0.22\textwidth]{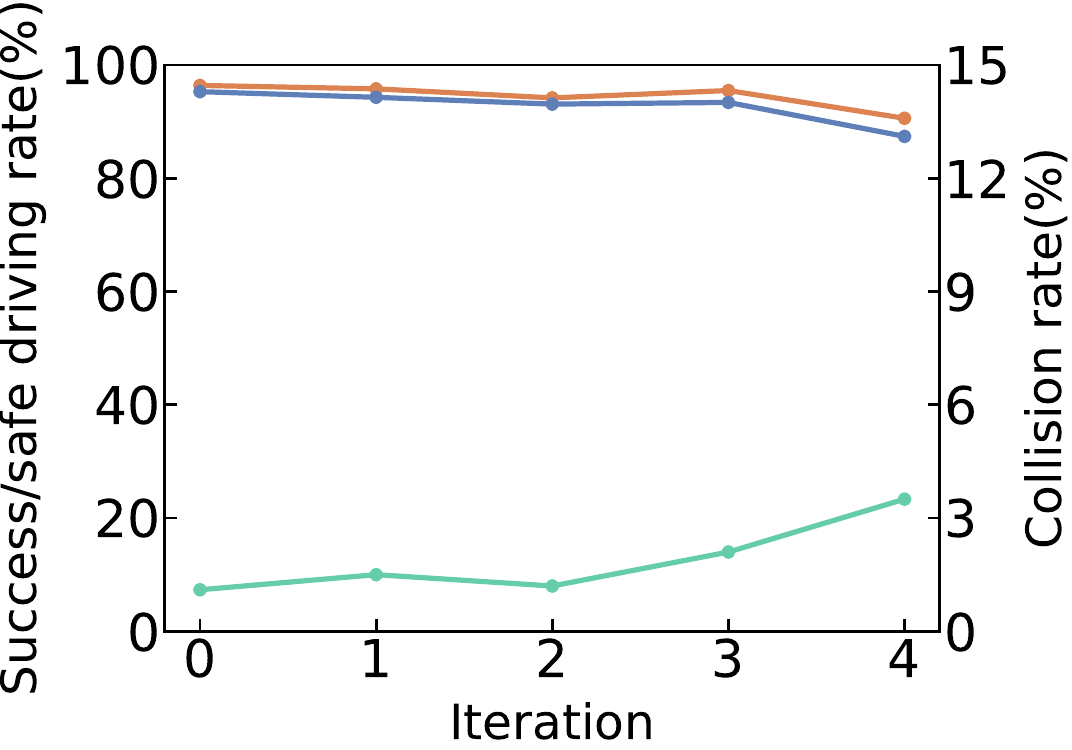}
    \end{minipage}
    
    \text{Changing to right lane}\par\medskip
    \subfigure[DSGN, $\alpha=0.4$]{
        \includegraphics[width=0.22\textwidth]{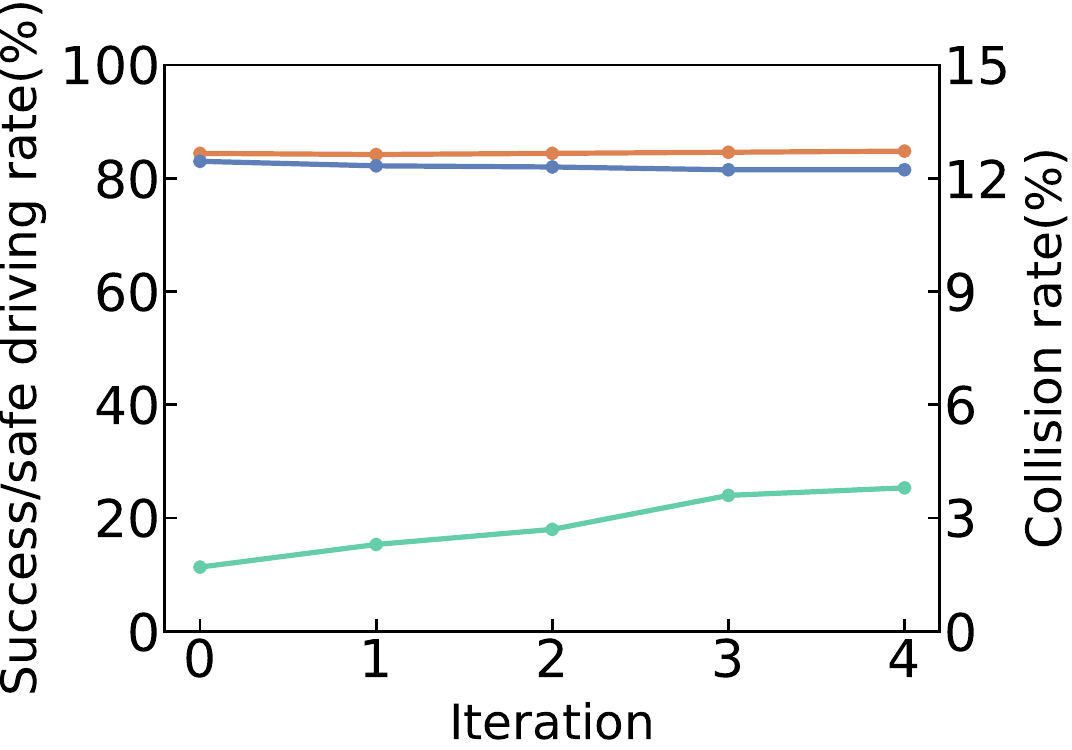}
    }\hspace{1.5pt}
    \subfigure[Stereo R-CNN, $\alpha=0.4$]{
        \includegraphics[width=0.22\textwidth]{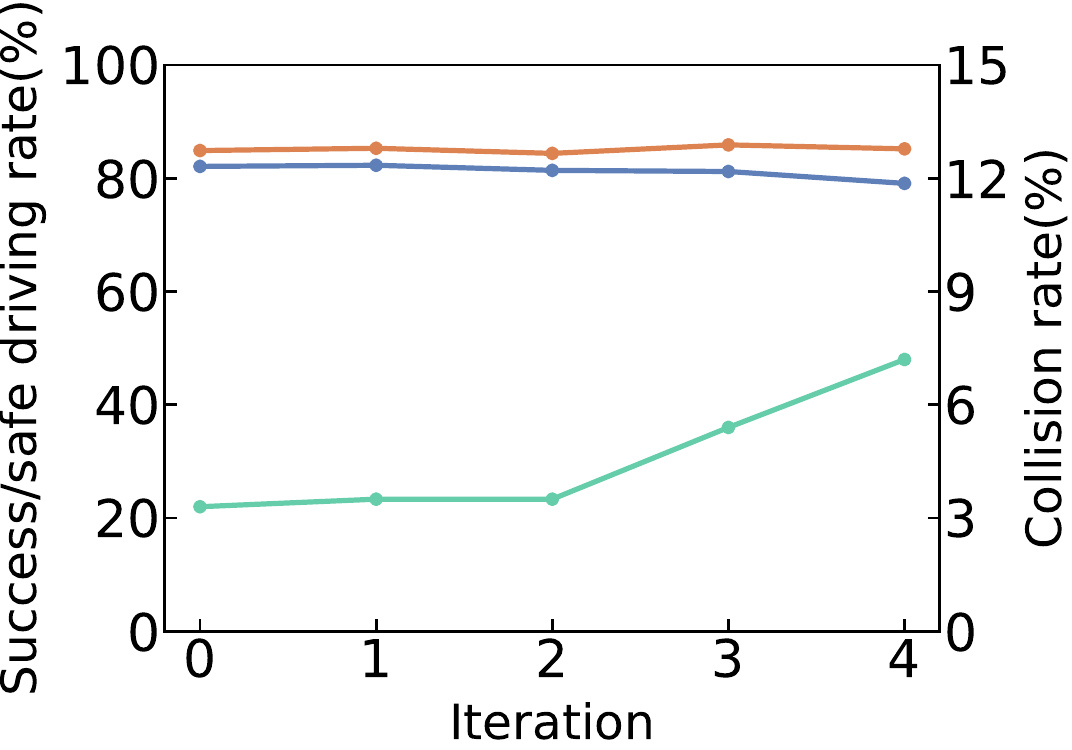}
    }\hspace{1.5pt}
    \subfigure[DSGN, $\alpha=1$]{
        \includegraphics[width=0.22\textwidth]{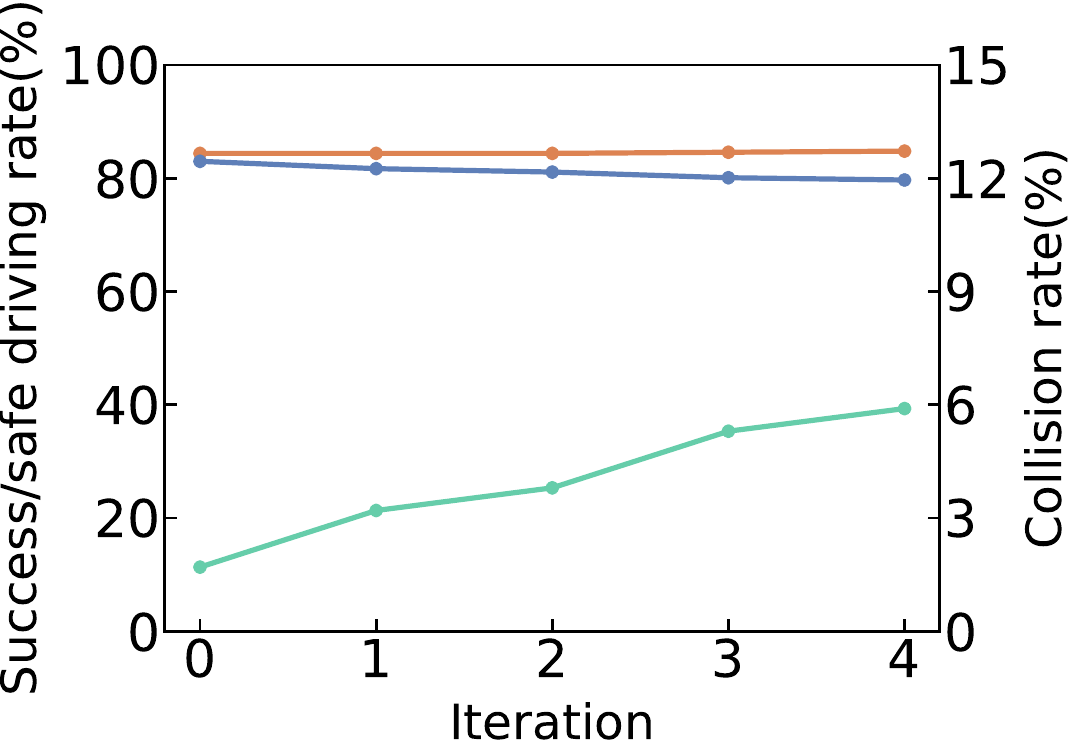}
    }\hspace{1.5pt}
    \subfigure[Stereo R-CNN, $\alpha=1$]{
        \includegraphics[width=0.22\textwidth]{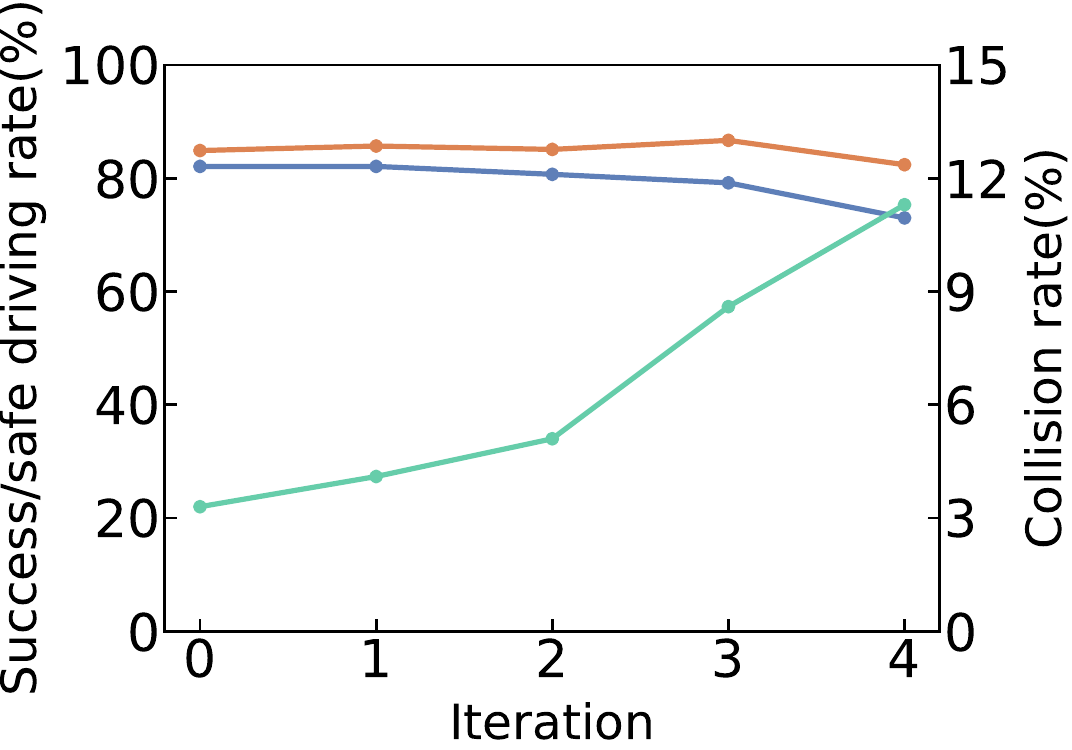}
    } \\
    \includegraphics[width=0.4\textwidth]{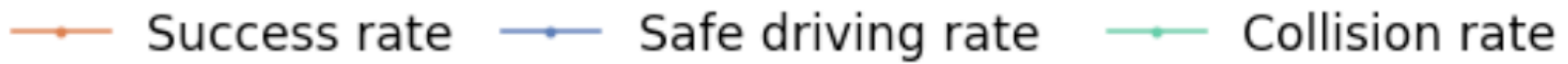}
    
    \caption{Driving safety performance metrics under the perturbation attack.}
    \label{fig:ds_perturbation}
\end{figure*}

\subsection{Implementation}

To implement this end-to-end driving safety evaluation framework for vision-based autonomous driving, we adopt two pre-trained models for the object detection module, namely, Stereo R-CNN~\cite{li19stereo} and DSGN~\cite{chen20dsgn}, which are currently two state-of-the-art methods in the area of vision-based 3D object detection. As for the motion planning module and the evaluation module, we use CommonRoad~\cite{althoff17commonroad} as the framework and leverage the built-in A*~\cite{hart68a} with sampled motion primitives as the motion planning method.

\begin{table*}[hbt!]
    \centering
    \caption{\\\textsc{Driving Safety Performance Metrics under the Perturbation Attack ($\alpha=0.4$)}}
    \resizebox{0.8\textwidth}{!}{
        \begin{tabular}{c|c|ccccc|ccccc}
        \hline
        \multirow{2}{*}{}          & Model & \multicolumn{5}{c|}{DSGN}             & \multicolumn{5}{c}{Stereo R-CNN}      \\ \cline{2-12} 
         & Iteration & Unattacked & 1     & 2     & 3     & 4     & Unattacked & 1     & 2     & 3     & 4     \\ \hline
        \multirow{3}{*}{Success rate ($\%$)}      & Left  & 89.6 & 89.6 & 89.5 & 89.3 & 89.5 & 89.8 & 89.8 & 90.4 & 90.0 & 88.8 \\
         & Straight  & 96.5      & 96.7 & 96.7 & 96.7 & 96.9 & 96.4      & 96.0 & 95.1 & 95.1 & 94.5 \\
         & Right     & 84.4      & 84.2 & 84.4 & 84.6 & 84.8 & 84.9      & 85.3 & 84.4 & 85.9 & 85.2 \\ \hline
        \multirow{3}{*}{Collision rate ($\%$)}    & Left  & 2.2 & 2.2 & 2.4 & 2.6 & 2.8 & 3.1 & 2.7 & 3.7 & 4.1 & 3.7 \\
         & Straight  & 0.7      & 0.7 & 0.7 & 0.9 & 1.1 & 1.1      & 1.1 & 0.9 & 1.3 & 1.7 \\
         & Right     & 1.7      & 2.3 & 2.7 & 3.6 & 3.8 & 3.3      & 3.5 & 3.5 & 5.4 & 7.2 \\ \hline
        \multirow{3}{*}{Safe driving rate ($\%$)} & Left  & 87.7 & 87.7 & 87.3 & 86.9 & 86.9 & 87.0 & 87.4 & 87.0 & 86.3 & 85.4 \\
         & Straight  & 95.8      & 96.0 & 96.0 & 95.8 & 95.8 & 95.3      & 94.9 & 94.1 & 93.8 & 92.9 \\
         & Right     & 83.0      & 82.2 & 82.0 & 81.5 & 81.5 & 82.1      & 82.3 & 81.4 & 81.2 & 79.1 \\ \hline
        \end{tabular}
    }
    \label{tab:ds_perturbation_0.4}
\end{table*}

\begin{table*}[hbt!]
    \centering
    \caption{\\\textsc{Driving Safety Performance Metrics under the Perturbation Attack ($\alpha=1$)}}
    \resizebox{0.8\textwidth}{!}{
        \begin{tabular}{c|c|ccccc|ccccc}
        \hline
        \multirow{2}{*}{}          & Model & \multicolumn{5}{c|}{DSGN}             & \multicolumn{5}{c}{Stereo R-CNN}      \\ \cline{2-12} 
         & Iteration & Unattacked & 1     & 2     & 3     & 4     & Unattacked & 1     & 2     & 3     & 4     \\ \hline
        \multirow{3}{*}{Success rate ($\%$)}      & Left  & 89.6 & 90.0 & 89.3 & 89.8 & 89.5 & 89.8 & 90.2 & 88.4 & 89.8 & 84.6 \\
         & Straight  & 96.5      & 96.7 & 96.5 & 96.5 & 96.9 & 96.4      & 95.8 & 94.2 & 95.5 & 90.6 \\
         & Right     & 84.4      & 84.4 & 84.4 & 84.6 & 84.8 & 84.9      & 85.7 & 85.1 & 86.7 & 82.4 \\ \hline
        \multirow{3}{*}{Collision rate ($\%$)}    & Left  & 2.2 & 2.6 & 2.8 & 3.4 & 3.4 & 3.1 & 3.3 & 3.2 & 4.7 & 6.4 \\
         & Straight  & 0.7      & 0.7 & 0.9 & 1.3 & 1.3 & 1.1      & 1.5 & 1.2 & 2.1 & 3.5 \\
         & Right     & 1.7    & 3.2 & 3.8 & 5.3 & 5.9 & 3.3      & 4.1 & 5.1 & 8.6 & 11.3 \\ \hline
        \multirow{3}{*}{Safe driving rate ($\%$)} & Left  & 87.7 & 87.7 & 86.7 & 86.7 & 86.4 & 87.0 & 87.2 & 85.5 & 85.5 & 79.2 \\
         & Straight  & 95.8      & 96.0 & 95.6 & 92.9 & 95.6 & 95.3      & 94.3 & 93.1 & 93.4 & 87.4 \\
         & Right     & 83.0      & 81.7 & 81.1 & 80.1 & 79.7 & 82.1      & 82.1 & 80.7 & 79.2 & 73.0 \\ \hline
        \end{tabular}
    }
    \label{tab:ds_perturbation_1}
\end{table*}

To implement the moving object classifier, we extract in total $600$ real driving scenarios from the KITTI raw dataset~\cite{geiger13vision} and manually label each object in each driving scenario with a moving/static property. We use 500 scenarios for training and 100 scenarios for validation. To determine whether an object is moving or not in a driving scenario, we refer to the previous and the subsequent image frames of that scenario. Though it is easy for human to judge the moving object from sequential frames, two people are assigned to manually label the scenarios independently in order to eliminate personal bias or errors in manual labelling. The independently produced labels are checked together for consistency, and no inconsistent labelling is found. We adopt the $16$-layer VGG net~\cite{simonyan14very} as the core network of the moving object classifier and replace its fully connected layers, i.e., $fc6$, $fc7$, and $fc8$, with a flatten layer, a new fully connected layer with a dropout layer and ReLU activation function, and another new fully connected layer with a dropout layer and a sigmoid activation function, respectively, to make sure that there is only one output score to indicate the probability of a moving object. The validation results suggest that the accuracy of the trained moving object classifier is 98.31\%.

To implement the driving constraint selector, we also leverage the KITTI raw dataset~\cite{geiger13vision} to train the model so that it can classify the road type of a scenario. Specifically, we divide the dataset into two subsets, i.e., \emph{street} and \emph{highway}. The street subset consists of city and residential scenarios where the traffic speed is relatively low, while the highway subset contains highway scenarios in which vehicles move much faster. Accordingly, we pre-define two sets of motion primitives for two road types so that the selector can pick the motion primitive with appropriate speed ranges and steering angle ranges after classifying the road type. The selector also chooses the dynamics constraints for moving vehicles predicted by the moving object classifier. The network architecture of the driving constraint selector consists of $5$ convolution layers connected by max-pooling layers and $1$ fully connected layer with dropout. Both convolution layers and the fully connected layer use ReLU as the activation function. After excluding the scenarios without cars, we select 444 scenarios as the training dataset and 112 scenarios as the validation dataset. The validation result indicates that the accuracy of the driving constraint selector achieves 94.64\%.

\begin{figure*}[!t]
\centering
\subfigure[Clean image input.]{
    \includegraphics[width=0.35\textwidth]{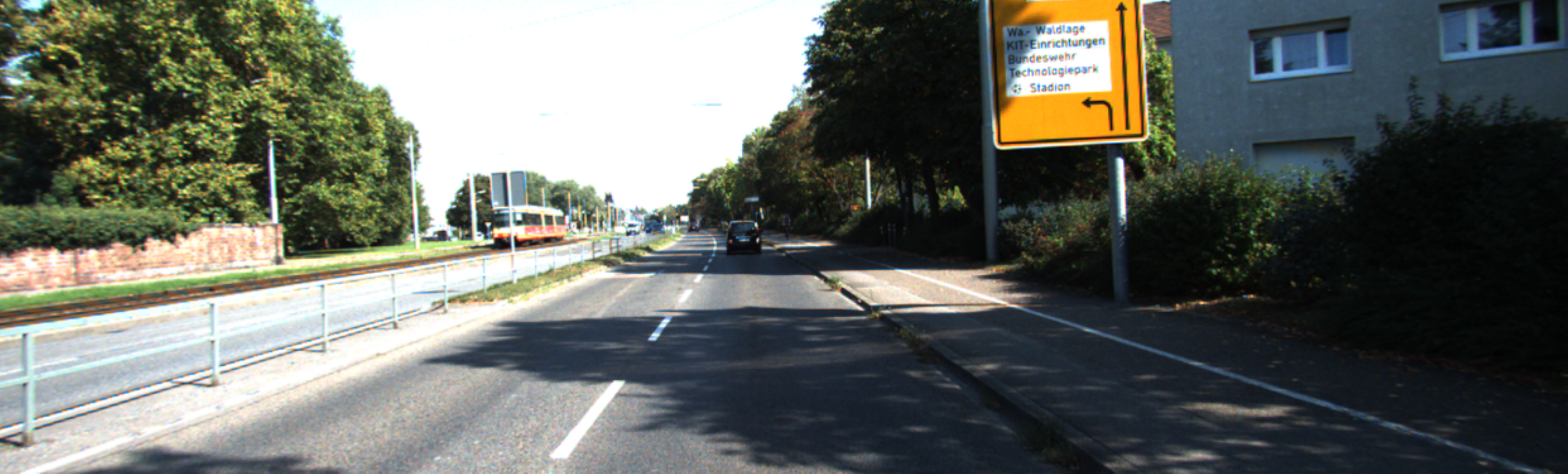}
}
\subfigure[Ground truth of object detection.]{
    \includegraphics[width=0.35\textwidth]{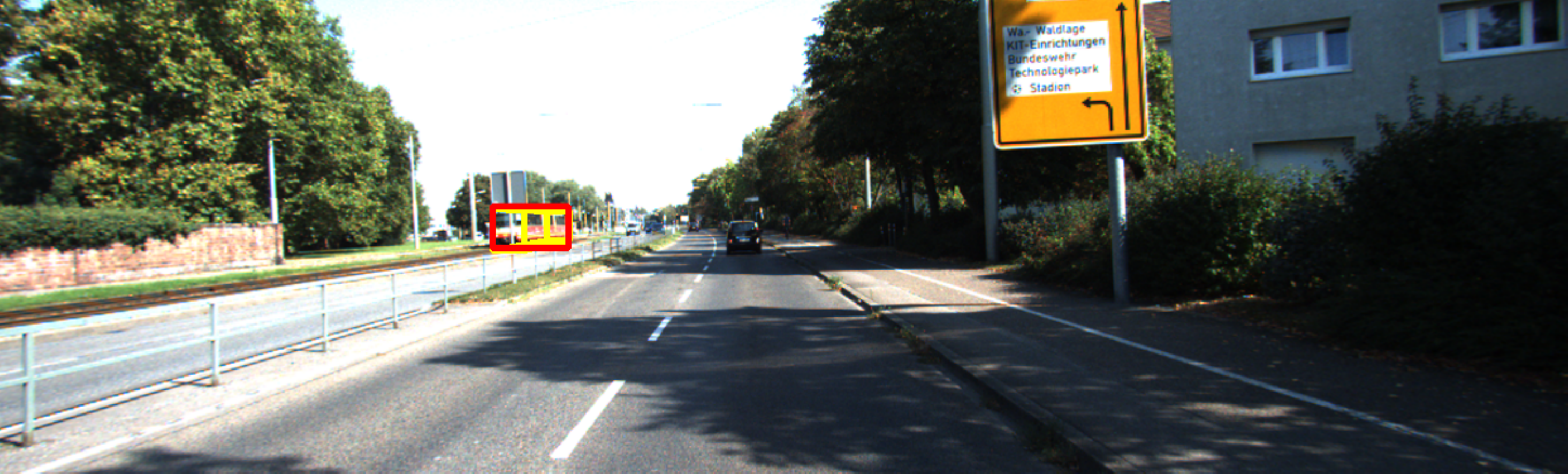}
} \\
\subfigure[Detection results of DSGN without attack.]{
    \includegraphics[width=0.35\textwidth]{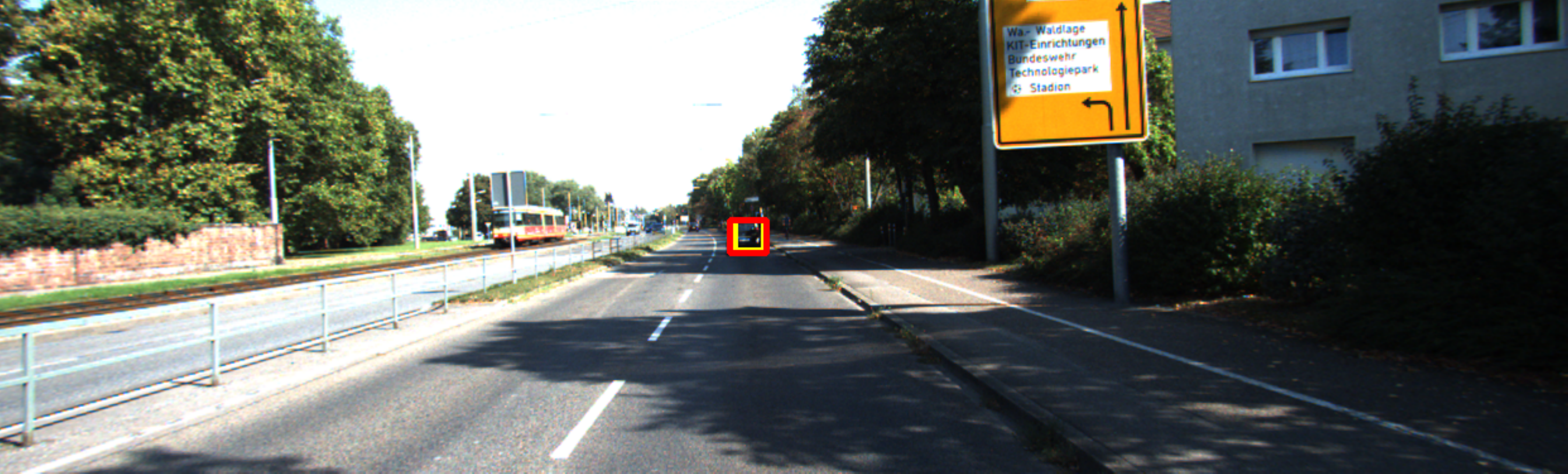}
}
\subfigure[Detection results of DSGN under attack.]{
    \includegraphics[width=0.35\textwidth]{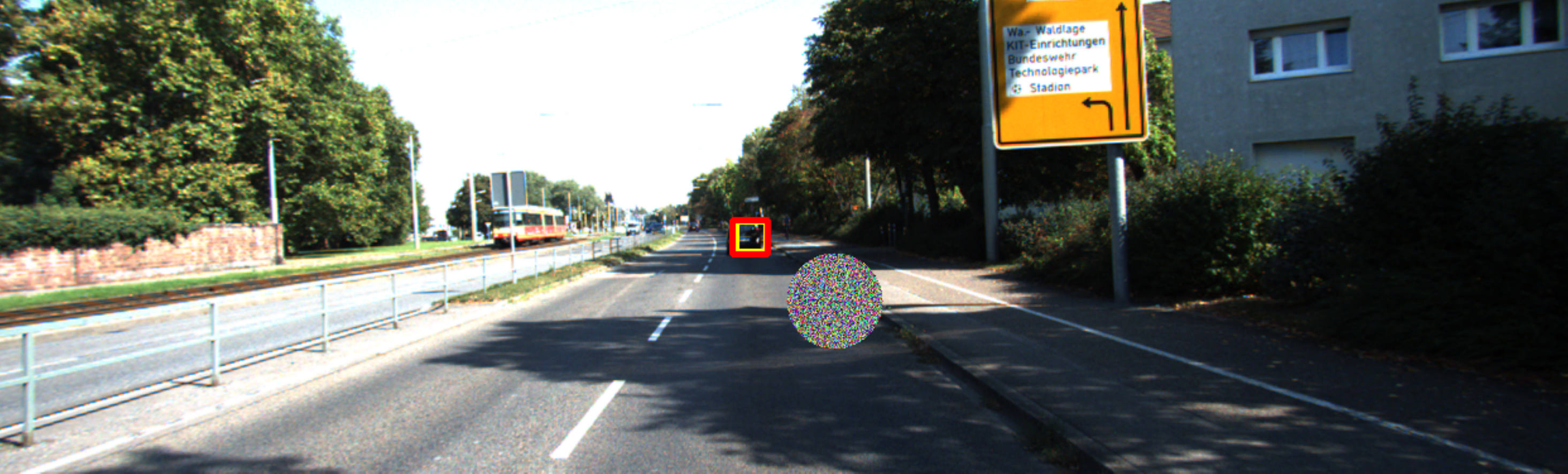}
} \\
\subfigure[Detection results of Stereo R-CNN without attack.]{
    \includegraphics[width=0.35\textwidth]{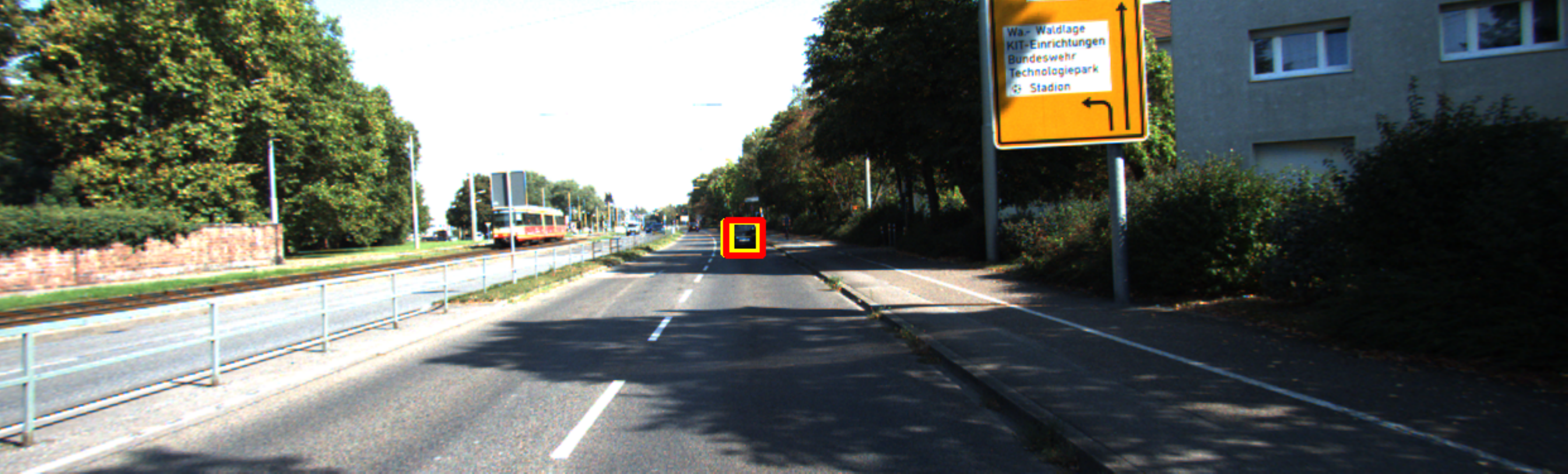}
}
\subfigure[Detection results of Stereo R-CNN under attack.]{
    \includegraphics[width=0.35\textwidth]{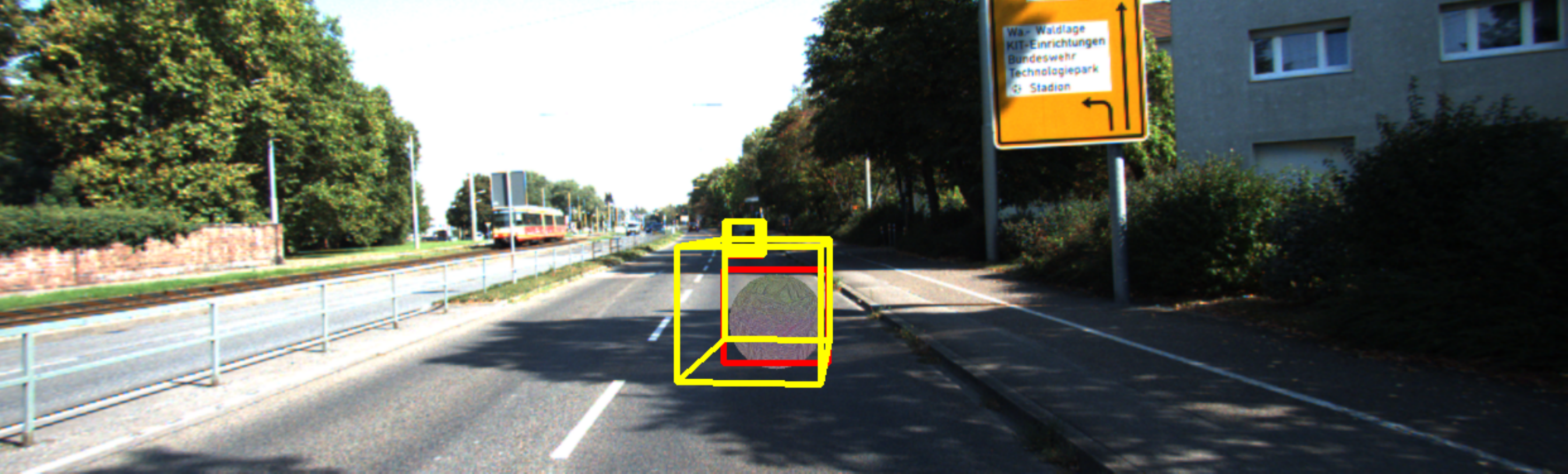}
} \\
\subfigure[Clean image input.]{
    \includegraphics[width=0.35\textwidth]{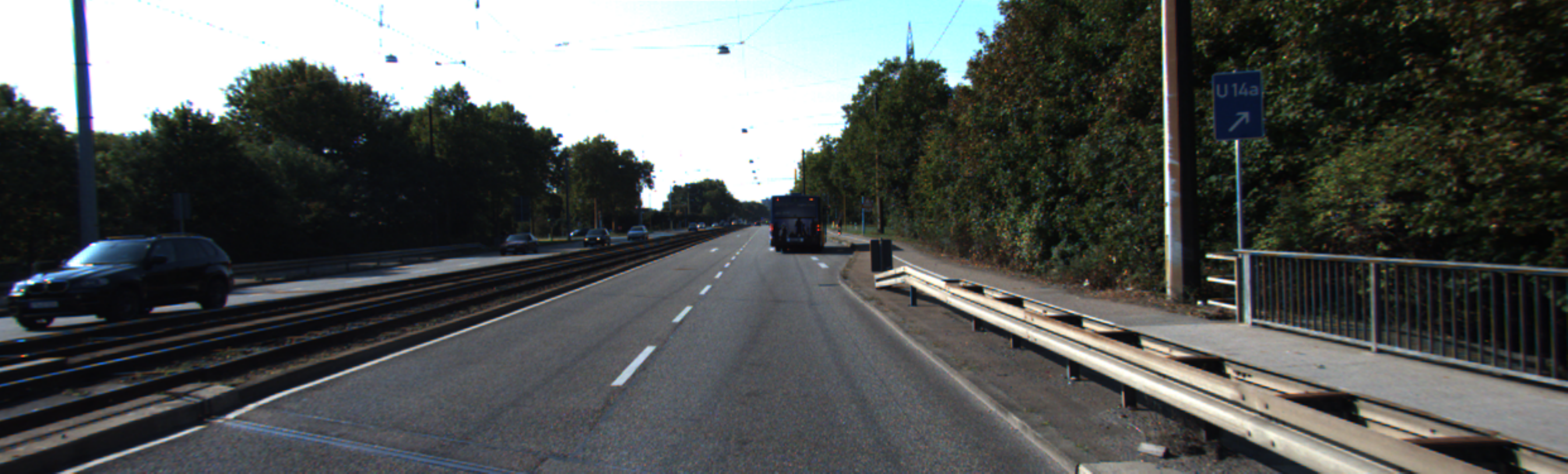}
}
\subfigure[Ground truth of object detection.]{
    \includegraphics[width=0.35\textwidth]{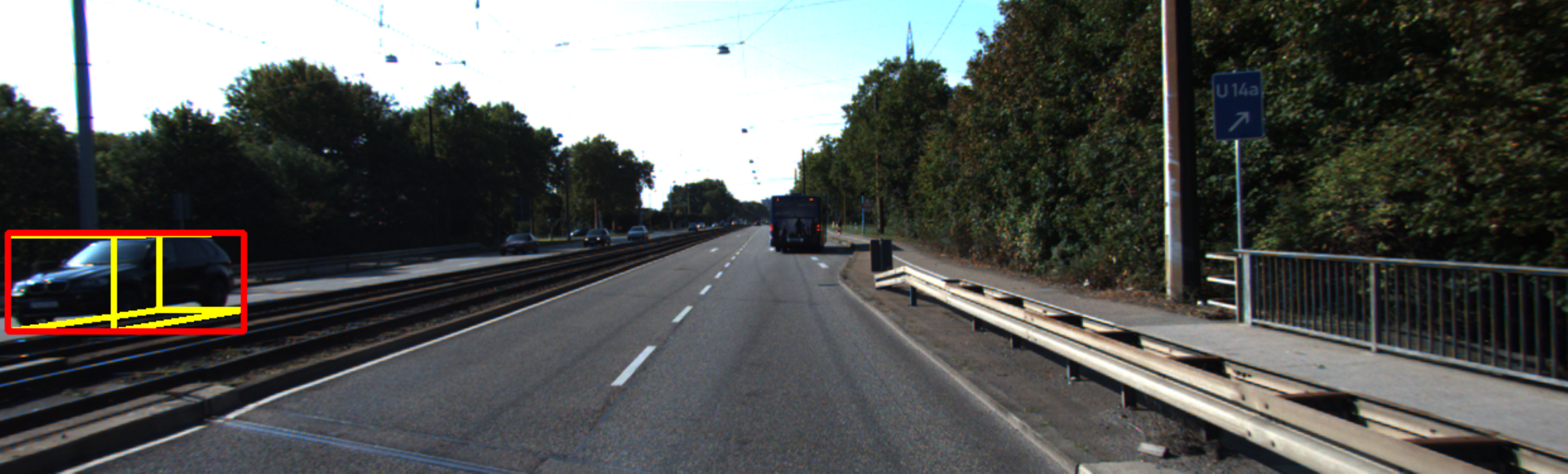}
} \\
\subfigure[Detection results of DSGN without attack.]{
    \includegraphics[width=0.35\textwidth]{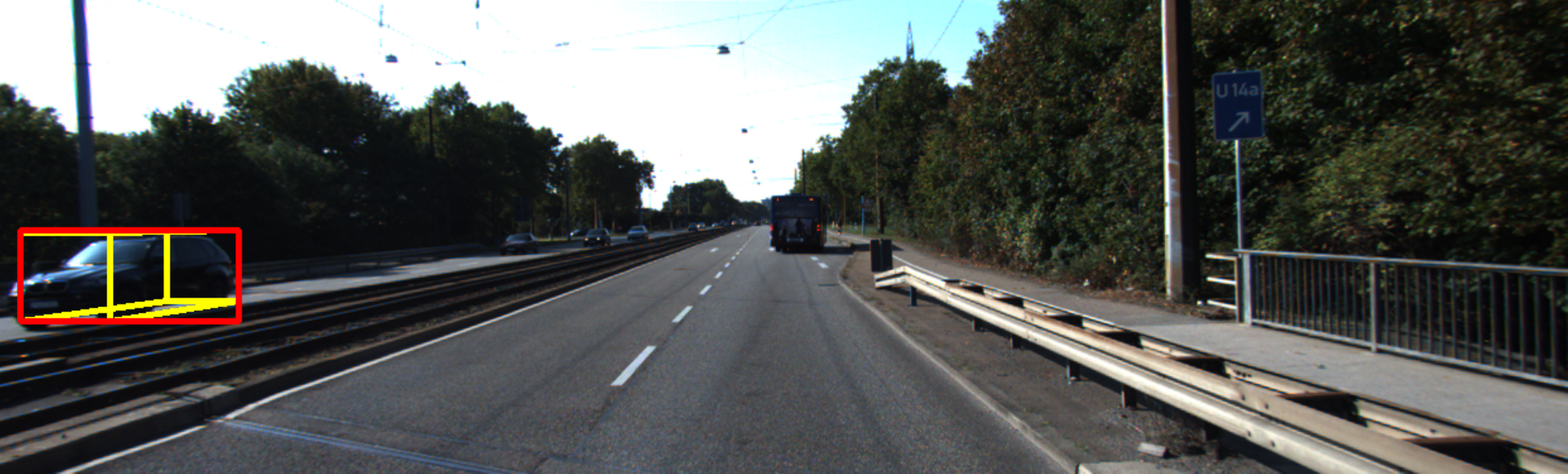}
}
\subfigure[Detection results of DSGN under attack.]{
    \includegraphics[width=0.35\textwidth]{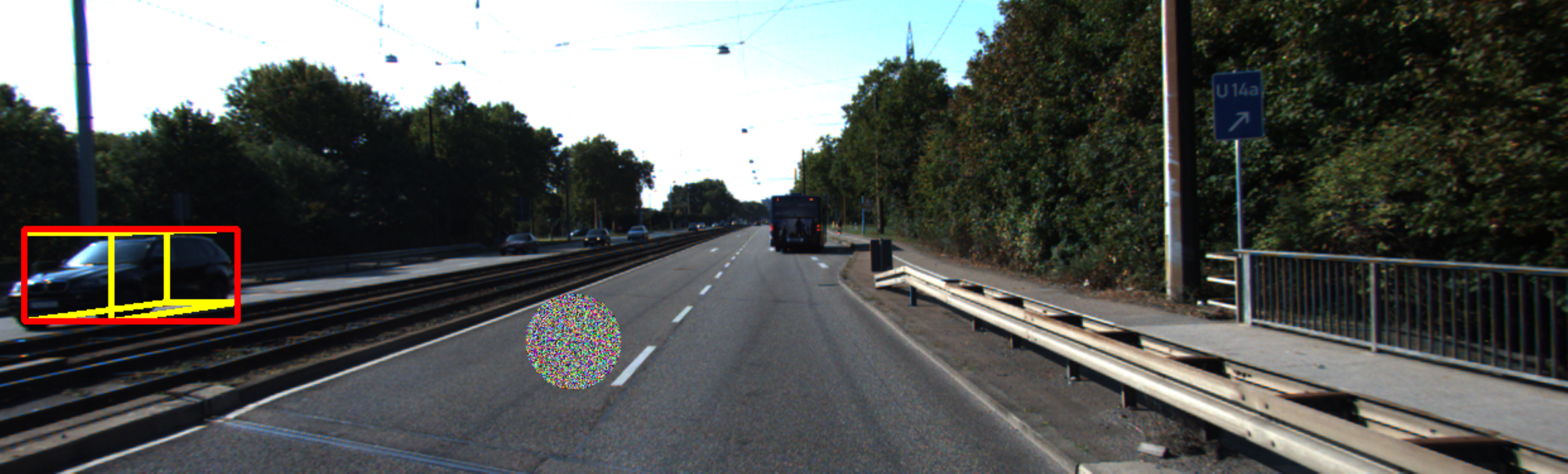}
} \\
\subfigure[Detection results of Stereo R-CNN without attack.]{
    \includegraphics[width=0.35\textwidth]{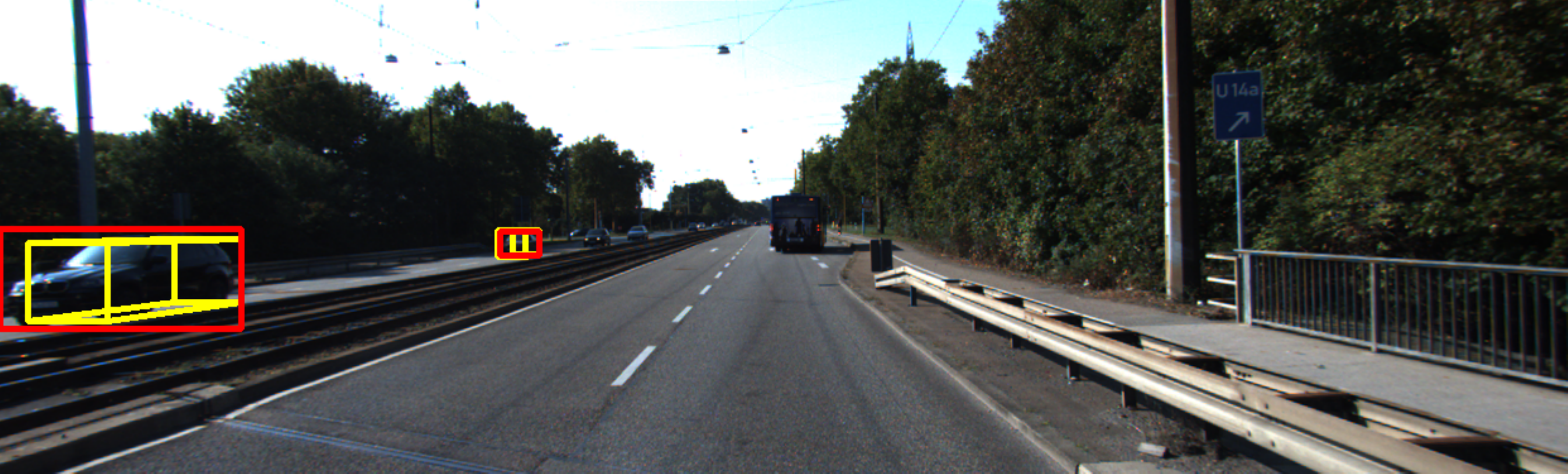}
}
\subfigure[Detection results of Stereo R-CNN under attack.]{
    \includegraphics[width=0.35\textwidth]{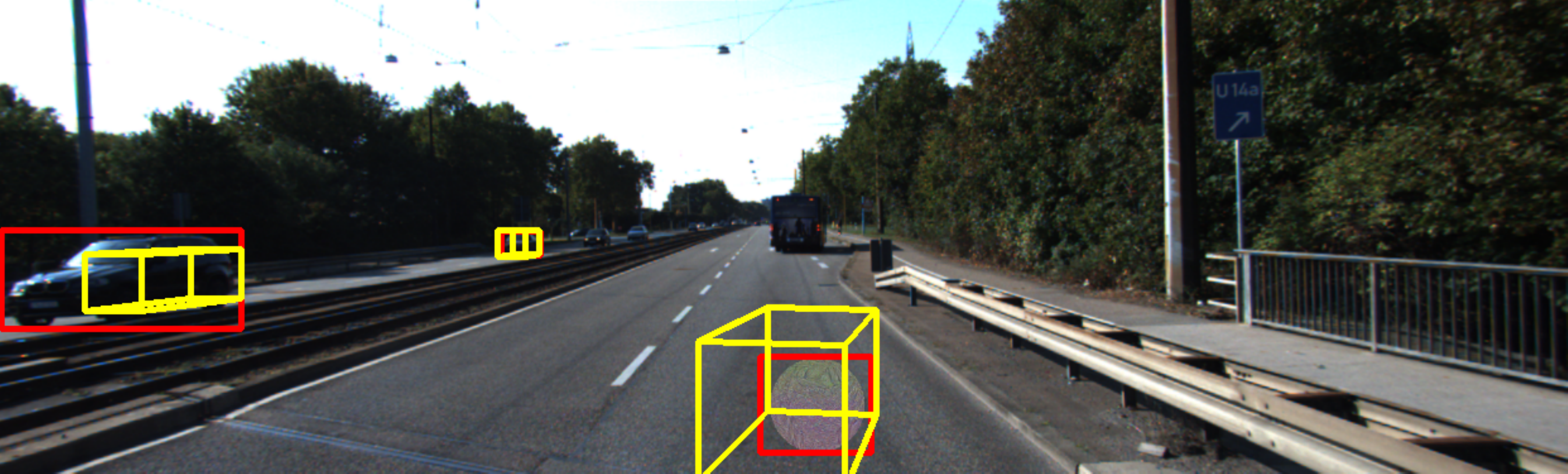}
} 
\caption{The patch attack triggers 3D object detectors to generate ghost bounding boxes at in the area of the patch as shown in (f) and (l). It has little influence on the detection of the objects away from the patch.}
\label{fig:plot_patch}
\end{figure*}

\section{Experiments}
\label{sec.experiments}

we conduct extensive experiments to investigate the impact of  perturbation attacks and patch attacks on driving safety of vision-based autonomous vehicles. We first introduce the common setup for all experiments, then elaborate on the specific settings for each attack experiment and present corresponding evaluation results. Finally, we summarize our findings at the end of this section.

\subsection{Common Setup}

In this paper, we conduct all experiments by applying the two types of adversarial attacks with different settings in our driving safety evaluation framework. Specifically, the evaluation framework includes two object detection modules, Stereo R-CNN~\cite{li19stereo} and DSGN~\cite{chen20dsgn}. In order to assess the impact comprehensively, we gradually increase the attack intensity by changing the attack settings in fine-grained steps. For the same purpose, we also consider three driving intentions of the ego-vehicle for each scenario when planning the trajectory, namely, \emph{changing to left lane}, \emph{changing to right lane}, and \emph{keeping lane}, which are abbreviated as \emph{left}, \emph{right}, and \emph{straight}, respectively. For these three cases, the initial position of the ego-vehicle is the same and the goal region is located 15 meters away from the initial position but within three different adjacent lanes. Moreover, we randomly assign  an initial speed within the selected speed range to each moving vehicle, including the ego-car. Specifically, the  initial speed for moving vehicles in street scenarios is randomly assigned within the range of $[22, 29]$ km/h, considering the $30$ km/h speed limit in German cities, campus and residential areas. The initial speed in highway scenarios is randomly assigned within the range of $[40, 47]$ km/h, concerning the $50$ km/h speed limit of built-up roads in Germany. For each attack, after the framework processes all the scenarios and generates the motion planning results, it assesses the attack impact on the performance metrics of driving safety as well as on the accuracy of 3D object detector. By linking these two attack impacts together, we manage to obtain evaluation results that help answer the questions raised in Section~\ref{sec.introduction}. The models of Stereo R-CNN~\cite{li19stereo} and DSGN~\cite{chen20dsgn} are pretrained with 3712 data points from the KITTI object detection dataset~\cite{geiger12are}. For each experiment setting, we test 600 real driving scenarios. The platform that we use is an Ubuntu 18.04 server equipped with an Nvidia Tesla V100 GPU.

In our experiments, the evaluation of driving safety is based on the trajectory produced by the motion planning module and measured by the driving safety performance metrics. In terms of evaluating the accuracy of the vision-based 3D object detector, we adopt the KITTI object detection benchmark that tests the detector with a three-level standard, namely, \emph{easy}, \emph{moderate}, and \emph{hard}~\cite{geiger12are}. We follow the standard to measure the \emph{average precision} (AP) of the detector with \emph{Intersection over Union} (IoU) larger than 70\%.

\begin{table*}[hbt!]
    \centering
    \caption{\\\textsc{Average Precision for 3D Object Detection under Patch Attack}}
    \resizebox{0.9\textwidth}{!}{
        \begin{tabular}{c|c|ccccc|ccccc}
        \hline
        \multirow{3}{*}{}        & Model    & \multicolumn{5}{c|}{DSGN}                & \multicolumn{5}{c}{Stereo R-CNN}          \\ \cline{2-12} 
         &
          \multirow{2}{*}{Scenario} &
          \multirow{2}{*}{Unattacked} &
          \multirow{1}{*}{Random} &
          \multicolumn{3}{c|}{Specific Attack} &
          \multirow{2}{*}{Unattacked} &
          \multirow{1}{*}{Random} &
          \multicolumn{3}{c}{Specific Attack} \\ \cline{5-7} \cline{10-12} 
                                         &          &       & Attack & Left  & Straight & Right &       & Attack & Left  & Straight & Right \\ \hline
        \multirow{1}{*}{3D} & easy     & 70.94 & 63.85 & 65.72 & 64.96    & 64.87 & 56.47 & 53.17 & 48.14 & 47.82    & 50.07 \\
                          Detection    & moderate & 52.98 & 48.20 & 51.47 & 50.77    & 51.63 & 38.20 & 37.07 & 36.27 & 35.23    & 38.02 \\
                                  AP ($\%$)   & hard     & 47.29 & 44.30 & 46.80 & 46.15    & 46.48 & 32.66 & 31.88 & 31.21 & 30.60    & 32.45 \\ \hline
        \end{tabular}
    }
    \label{tab:ap_patch}
\end{table*}

\begin{table*}[hbt!]
    \centering
    \caption{\\\textsc{Driving Safety Performance Metrics under the Patch Attack}}
    \resizebox{0.9\textwidth}{!}{
        \begin{tabular}{c|c|ccc|ccc}
        \hline
        \multirow{2}{*}{}              & Model    & \multicolumn{3}{c|}{DSGN}                    & \multicolumn{3}{c}{Stereo R-CNN}              \\ \cline{2-8} 
                                              & Scenario & Unattacked & Random Attack & Specific Attack & Unattacked & Random Attack & Specific Attack \\ \hline
        \multirow{3}{*}{Success rate(\%)}   & Left     & 89.6 & 90.0 & 90.0 & 89.8 & 70.3 & 83.7 \\
                                            & Straight & 96.5 & 96.5 & 96.7 & 96.4 & 81.2 & 57.5 \\
                                            & Right    & 84.4 & 84.6 & 84.8 & 84.9 & 68.7 & 74.4 \\ \hline
        \multirow{3}{*}{Collision rate(\%)} & Left     & 2.2  & 2.6  & 2.8  & 3.1  & 1.9  & 2.1  \\
                                            & Straight & 0.7  & 0.7  & 0.9  & 1.1  & 1.4  & 4.7  \\
                                            & Right    & 1.7  & 2.3  & 2.1  & 3.3  & 3.9  & 4.2  \\ \hline
        \multirow{3}{*}{Safe driving rate(\%)} & Left     & 87.7       & 87.7          & 87.5            & 87.0       & 68.9          & 81.9            \\
                                            & Straight & 95.8 & 95.8 & 95.8 & 95.3 & 80.1 & 54.7 \\
                                            & Right    & 83.0 & 82.6 & 83.0 & 82.1 & 65.9 & 71.2 \\ \hline
        \end{tabular}
    }
    \label{tab:ds_patch}
\end{table*}

\subsection{Perturbation Attack}

In order to perform the perturbation attack against autonomous driving systems at various intensities, we adjust two parameters $\alpha$ and $n$ in Eqn.~(\ref{eq:eq_2}). To ensure that the perturbation is imperceivable to human eyes, their values usually should be set as small as possible. Specifically, we set the value of $\alpha$ as $0.4$ and $1$, to represent medium to high attack intensities, respectively. The number of iterations $n$ changes from $1$ to $4$ accordingly, so that the modification on image pixel values is constrained within the range of $[0.4,4]$. We note that even the attack with the lowest attack intensity, i.e., $\alpha=0.4$ and $n=1$, can cause significant decline in the accuracy of 3D object detectors. Moreover, the produced perturbation and the input stereo images have the same dimension.

\begin{table*}[hbt!]
\centering
\caption{\\\textsc{Driving Safety Performance Metrics of Stereo R-CNN under the Patch Attack with Various Intentions}}
\begin{tabular}{c|c|ccc}
\hline
                       & Specific attack & \multicolumn{3}{c}{Random attack}                  \\ \cline{2-5} 
                       & —               & Same intentions$^1$ & Different intentions$^2$ & Unattacked  \\ \hline
Success rate (\%)      & 71.9            & 76.2            & 91.6                 & 90.4       \\ \hline
Collision rate (\%)    & 3.5             & 2.3             & 1.5                  & 2.4        \\ \hline
Safe driving rate (\%) & 69.3            & 74.4            & 90.1                 & 88.1       \\ \hline
\end{tabular}
\\
$^1$ "Same intentions" refers to cases where the attack intention and the driving intention are the same.
\\
$^2$ "Different intentions" refers to cases where the attack intention differs from the driving intention.
\label{tab:intentions}
\end{table*}

\begin{table}[hbt!]
\centering
\caption{\\\textsc{Safe Driving Rate using Different Planning Algorithms}}
\begin{tabular}{c|c|c|c}
\hline
\multirow{2}{*}{}                       & Planning algorithm & GBFS            & A*              \\ \cline{2-4} 
                                        & Scenario           & \multicolumn{2}{c}{Ground Truth} \\ \hline
\multirow{3}{*}{Safe driving rate (\%)} & Left               & 87.9            & 89.7            \\
                                        & Straight           & 98.0            & 98.0            \\
                                        & Right              & 82.3            & 85.2            \\ \hline
\end{tabular}
\label{tab:plan_alg}
\end{table}

\begin{table*}[hbt!]
\centering
\caption{\\\textsc{Safe Driving Rate with Different Inputs}}
\begin{tabular}{c|c|c|ccc|ccc}
\hline
\multirow{2}{*}{}                       & Model    & —    & \multicolumn{3}{c|}{DSGN} & \multicolumn{3}{c}{Stereo R-CNN} \\ \cline{2-9} 
 &
  Scenario &
  \begin{tabular}[c]{@{}c@{}}Ground \\ Truth\end{tabular} &
  Unattacked &
  \begin{tabular}[c]{@{}c@{}}Perturbation\\ Attack\end{tabular} &
  \begin{tabular}[c]{@{}c@{}}Patch\\ Attack\end{tabular} &
  Unattacked &
  \begin{tabular}[c]{@{}c@{}}Perturbation\\ Attack\end{tabular} &
  \begin{tabular}[c]{@{}c@{}}Patch\\ Attack\end{tabular} \\ \hline
\multirow{3}{*}{Safe driving rate (\%)} & Left     & 89.7 & 87.7    & 86.4   & 87.7   & 87.0      & 79.2      & 68.9      \\
                                        & Straight & 98.0 & 95.8    & 95.6   & 95.8   & 95.3      & 87.4      & 80.1      \\
                                        & Right    & 85.2 & 83.0    & 79.7   & 82.6   & 82.1      & 73.0      & 65.9      \\ \hline
\end{tabular}
\label{tab:atk_impact}
\end{table*}

\textbf{Evaluation.} The effect of the perturbation attack in some driving scenarios is shown in Figure~\ref{fig:plot_perturbation} where we can see that the attack causes inaccurate detection of real objects and false detection of ghost objects. We present the impact of the perturbation attack with different settings on average precision of 3D object detection and on driving safety metrics in Figure~\ref{fig:ap_perturbation} and Figure~\ref{fig:ds_perturbation}, respectively. The numerical results of driving safety scores can be found in Table~\ref{tab:ds_perturbation_0.4} and Table~\ref{tab:ds_perturbation_1}. When the number of iterations $n$ is 0, it indicates that there is no attack applied. From Figure~\ref{fig:ap_perturbation}, we can observe that with the enhanced attack intensity by increasing $\alpha$ and $n$, the average precision of both object detection models drops significantly, while the driving safety metrics only show very small changes. Take DSGN as an example. When $\alpha$ is 0.4 and $n$ is increased from $0$ (no attack) to $1$, the AP declines by more than half for all three levels of the benchmark standard, i.e., from 70.94\% to 21.99\% for the category of AP easy, from 52.98\% to 14.45\% for the category of AP moderate, and from 47.29\% to 13.96\% for the category of AP hard. When $n$ is 3, the AP of DSGN almost reaches 0. However, in the meantime, the driving safety performance metrics in Figure~\ref{fig:ds_perturbation} barely change for all three intention cases (e.g., collision rate is only changed from $1.7\%$ to $3.6\%$ for the case of changing to right lane). When $\alpha = 1$ and $n$ is increased from $0$ (no attack) to $4$, the AP of DSGN drops even more significantly, but the driving safety metrics only demonstrate slightly larger changes than that when $\alpha = 0.4$ (e.g., the safe driving rate drops by $0.8\%$ when $\alpha = 0.4$ and by $1.3\%$ when $\alpha = 1$ for the case of changing to left lane). The experiment results clearly indicate that the perturbation attack can dramatically affect the performance of 3D object detection methods, but does not have much influence on the driving safety. In other words, a larger precision decline of the vision-based 3D object detectors under the perturbation attack does not indicate higher risk of driving safety.

Moreover, by comparing DSGN and Stereo R-CNN in terms of driving safety under perturbation attacks, we can observe that the changes in driving safety metrics for Stereo R-CNN tend to be larger than the changes for DSGN when both of them are tested in the same driving intention scenarios and at the same intensity. Therefore, Stereo R-CNN is more prone to  perturbation attacks than DSGN in regard to driving safety.

\subsection{Patch Attack}

Different from  perturbation attacks, the size of a patch in a patch attack is much smaller than the size of an input image. In our patch attack experiments, the radius of the patch is limited to $38$ pixels. Here, the patch attack is launched as a white-box attack, which means that the patch is trained for Stereo R-CNN and DSGN, respectively. Specifically, we train the patch according to Eqn.~\ref{eq:eq_3} by placing the patch at a random position in stereo image pair and setting $b^{*}$ in Eqn.~\ref{eq:eq_4} accordingly for each training scenario. To ensure that the patch for Stereo R-CNN and the patch for DSGN are equally optimized, we use the same learning step size and the same number of epochs when training patches. 

We design two attack approaches, namely, \emph{random attack} and \emph{specific attack}. Random attacks are to place the trained patch at a random position within the entire image no matter which driving intention case it is. In other words, the attack intention may or may not be consistent with the driving intention in random attacks. Specific attacks are also to place the patch randomly but within a certain region of the image, depending on the driving intention case, e.g., if the driving intention is  changing to right lane, then the patch is placed in the right part of the image. In other words, the attack intention is always consistent with the driving intention in specific attacks. By designing two attack approaches, we can create different attack intensities for patch attacks. Specifically, the attack intensity of random attacks is lower than that of specific attacks.

\textbf{Evaluation.} The performance of patch attacks in some driving scenarios is shown in Figure~\ref{fig:plot_patch} in which we can observe that patch attacks cause false detection of ghost objects. The  impact of  patch attacks on object detection and driving safety are shown in Table~\ref{tab:ap_patch} and Table~\ref{tab:ds_patch} respectively. From the tables, we can observe that, when different attack approaches are applied, the average precision of both object detection models declines slightly, while some of the driving safety metrics degrade significantly. For example, when random patch attacks are applied to the Stereo R-CNN model, AP declines slightly for all three levels of the benchmark standard, i.e., from $56.47\%$ to $53.17\%$ for the case of AP easy, from $38.20\%$ to $37.07\%$ for the case of AP moderate, and  $32.66\%$ from to $31.88\%$ for the case of AP hard. However, the driving safety performance metrics of Stereo R-CNN have a relatively larger drop under random patch attacks (e.g., safe driving rate drops from $95.3\%$ to $80.1\%$ for the case of keeping lane). At the same time, for specific patch attacks, Stereo R-CNN shows the similar average precision decline which is only within the range of $[0.21\%, 8.65\%]$, while significant driving safety performance degradation can be observed (e.g., the safe driving rate decreases to half for the case of keeping lane). The experiment results suggest that a slight precision decline of the 3D object detectors under  patch attacks does not indicate mild risk of driving safety. 

Since the driving safety performance of Stereo R-CNN can be significantly affected by patch attacks, we further investigate the performance under the attacks where the attack intention is the same as the driving intention, and the attacks where the attack intention is different from the driving intention. The results are listed in Table~\ref{tab:intentions}. From Table~\ref{tab:intentions}, we can see that  the driving safety performance under the attacks where the driving intention and the attack intention are different is very similar to that in unattacked scenarios, and the performance under the attacks where the attack intention is the same as the driving intention is very close to that in specific attack scenarios.

Furthermore, the DSGN model again shows its much better robustness in object detection and driving safety under patch attacks. We can observe that,  even under the well-designed specific patch attacks, DSGN's average precision decline is only less than $6\%$, and the driving safety performance metrics almost remain unchanged, while Stereo R-CNN performs worse under both  random patch attacks and  specific patch attacks.

\begin{figure*}[!t]
\subfigure[No attack.]{
    \includegraphics[width=0.32\textwidth]{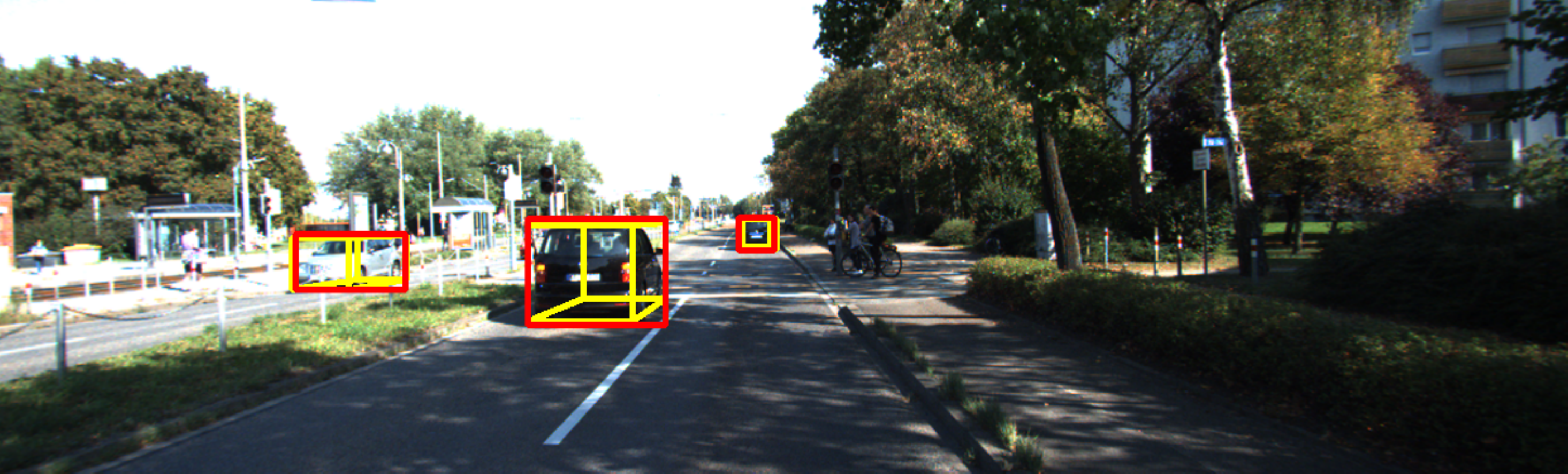}
    \label{fig:ab_1_stereo_1}
}
\subfigure[Attack applied.]{
    \includegraphics[width=0.32\textwidth]{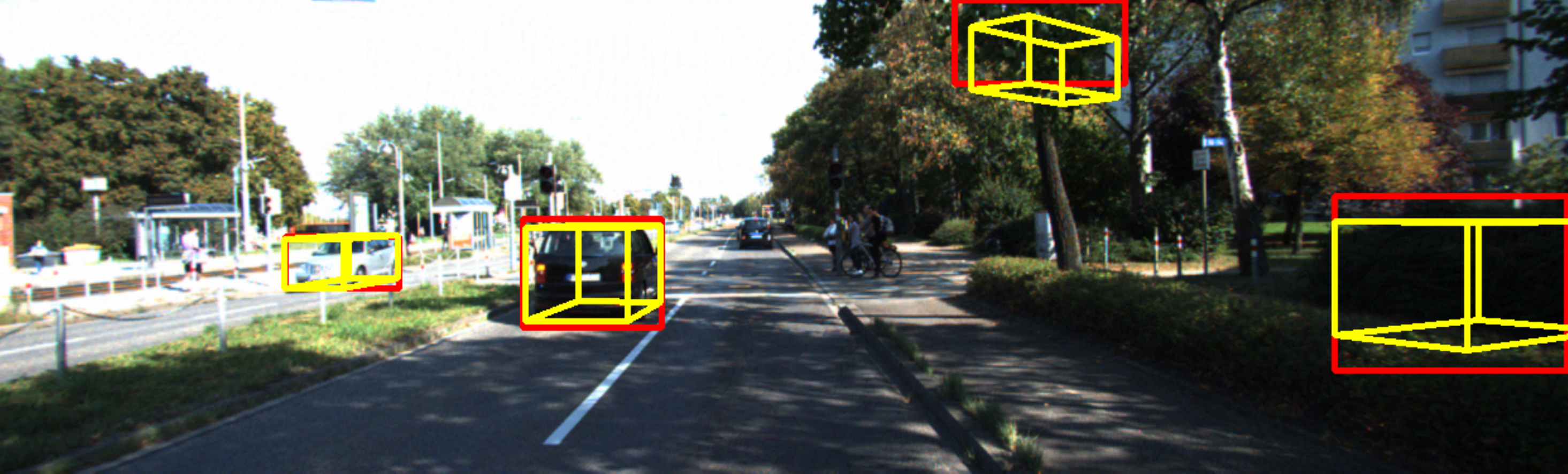}
    \label{fig:ab_1_stereo_2}
}
\subfigure[Attack applied and texture of side area replaced.]{
    \includegraphics[width=0.32\textwidth]{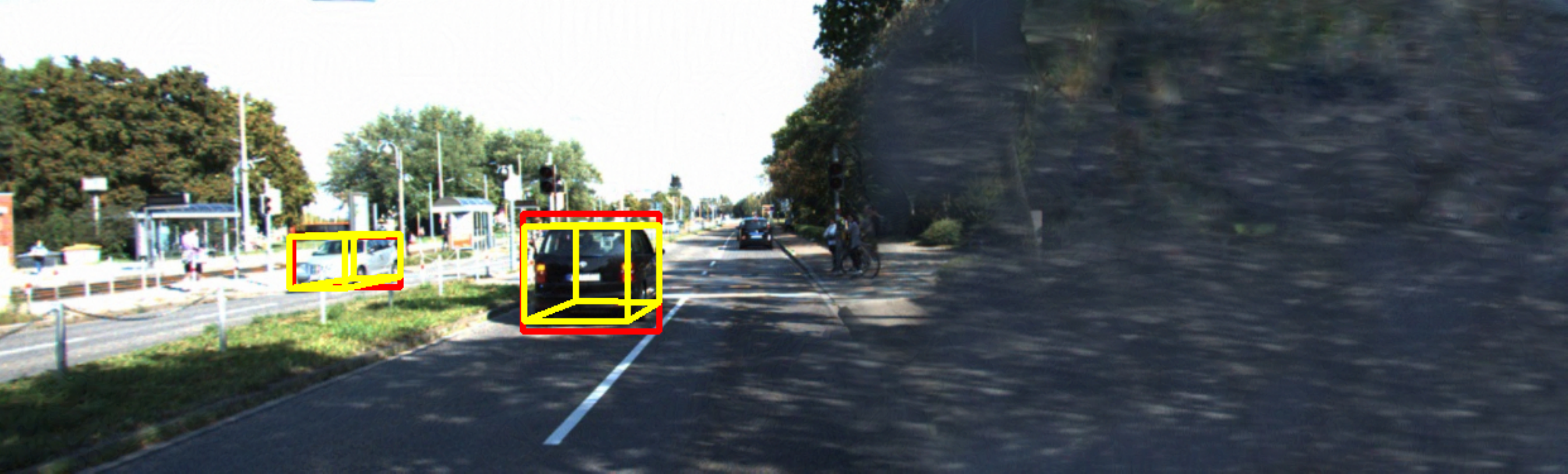}
    \label{fig:ab_1_stereo_3}
}
\caption{Results of the texture replacement experiment for Stereo R-CNN.}
\label{fig:ab_1_stereo}
\end{figure*}

\begin{figure*}[!t]
\centering
\subfigure[No attack.]{
    \includegraphics[width=0.31\textwidth]{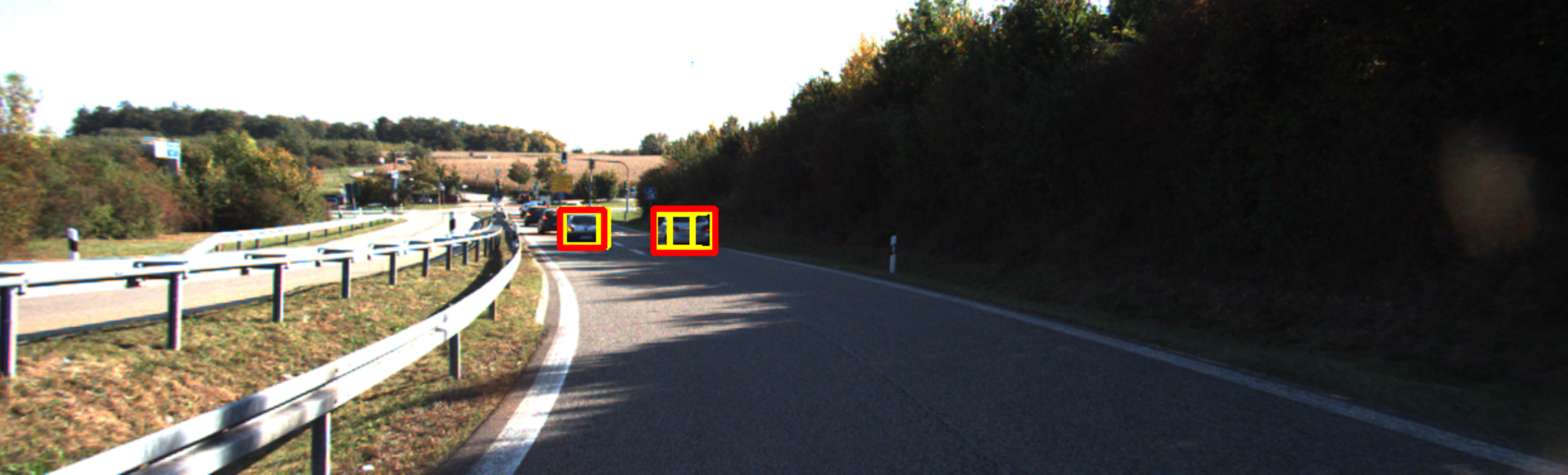}\label{fig:ab_1_dsgn_1}
}
\subfigure[Attack applied.]{
    \includegraphics[width=0.31\textwidth]{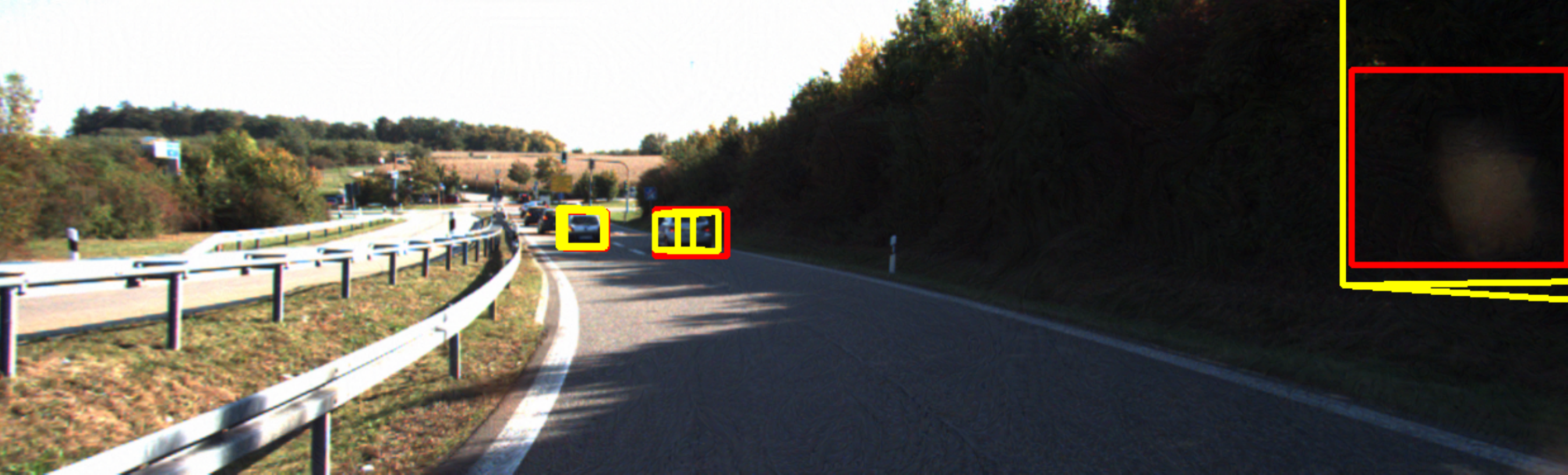}\label{fig:ab_1_dsgn_2}
}
\subfigure[Attack applied and texture of side area replaced.]{
    \includegraphics[width=0.31\textwidth]{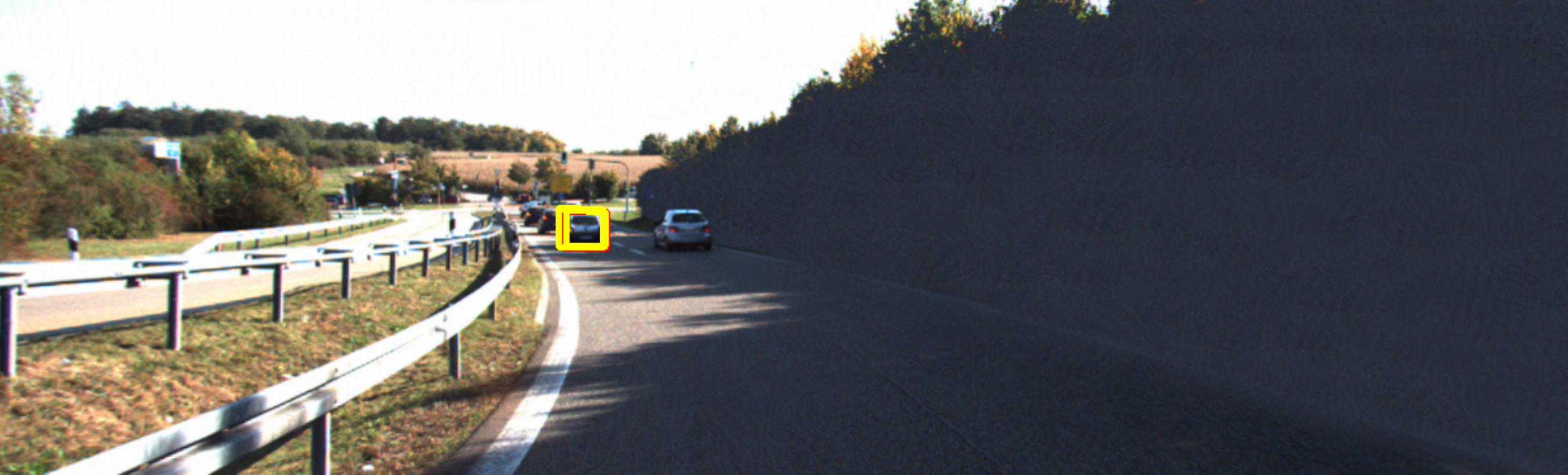}\label{fig:ab_1_dsgn_3}
}
\caption{Results of the texture replacement experiment for DSGN.}
\label{fig:ab_1_dsgn}
\end{figure*}

\subsection{Attack Impact Demonstration}

To demonstrate that the performance of different models under different attacks is mainly caused by adversarial attacks, not by the motion planning algorithms, we conduct two experiments.

We first evaluate the performance of the motion planning module using different inputs and then calculating the safe driving rates in different scenarios. Specifically, we first use the ground truth data of 3D object detection as the inputs to evaluate two popular motion planning algorithms, A* and Greedy-BFS, so as to show the impact of the motion planning module on the driving safety. The experimental results are summarized in Table~\ref{tab:plan_alg}. From Table~\ref{tab:plan_alg}, we can first observe that, when ground truth data are used, the A* planning algorithm can achieve the safe driving rates $89.7\%$, $98.0\%$, and $85.2\%$ for the three driving intention scenarios, respectively, while the Greedy-BFS algorithm can achieve the safe driving rates $87.9\%$, $98.0\%$, and $82.3\%$ for the three driving intentions scenarios, respectively. The performance of A* and the performance of the Greedy-BFS are very close. In other words, the performance variance demonstrated by DSGN and Stereo R-CNN under adversarial attacks is irrelevant to the selection of the motion planning algorithm. Since the performance of A* is slightly better than that of Greedy-BFS, we select A* as the motion planning algorithm for our driving safety evaluation framework.

We then use detection data without attacks (i.e., unattacked) and detection data under two types of attacks to demonstrate the impact of detection module and adversarial attacks on the driving safety. The results are shown in Table~\ref{tab:atk_impact}. From Table~\ref{tab:atk_impact}, we can observe that the safe driving rates produced by the detection data without attacks are slightly smaller than the safe driving rates when the ground truth data are used as inputs. This slight difference is caused by the accuracy of the two models. Finally, compared with the safe driving rates when the inputs are unattacked detection data, the safe driving rates under adversarial attacks are significantly declined in all driving intention scenarios. Since all experiments are conducted using the same motion planning algorithm, we can conclude that the declines in the driving safety performance metrics are primarily caused by adversarial attacks.

\subsection{Findings}

To briefly summarize, the findings from the experiments of perturbation attacks and patch attacks are listed as follows:

\begin{itemize}
\item
A larger precision decline of the attacked vision-based object detectors does not necessarily indicate a higher risk of driving safety. Similarly, a slight precision decline of the vision-based object detectors under attacks does not necessarily indicate a small risk of driving safety, either. Hence, the precision decline or the erroneous rate increase of the vision-based object detectors under attacks cannot represent their robustness with respect to driving safety of autonomous vehicles.

\item
Stereo R-CNN is less robust than DSGN in terms of driving safety and detection accuracy when the attacks launched on them are at the same intensity level.  Hence, DSGN is a better selection of the vision-based 3D object detection for its stronger robustness and higher detection precision.
\end{itemize}

In terms of these two findings, we further design more experiments to explain the real causes behind them in the next section.

\section{Ablation Study}
\label{sec.ablationstudy}

In this section, we first investigate the reason of the decoupling between the precision of 3D object detectors and the driving safety performance metrics under adversarial attacks. Second, we investigate the reason why the DSGN model is more robust than the Stereo R-CNN model.

\begin{figure*}[!t]
\centering
\begin{minipage}{0.070\textwidth}
    \centerline{\includegraphics[width=1\textwidth]{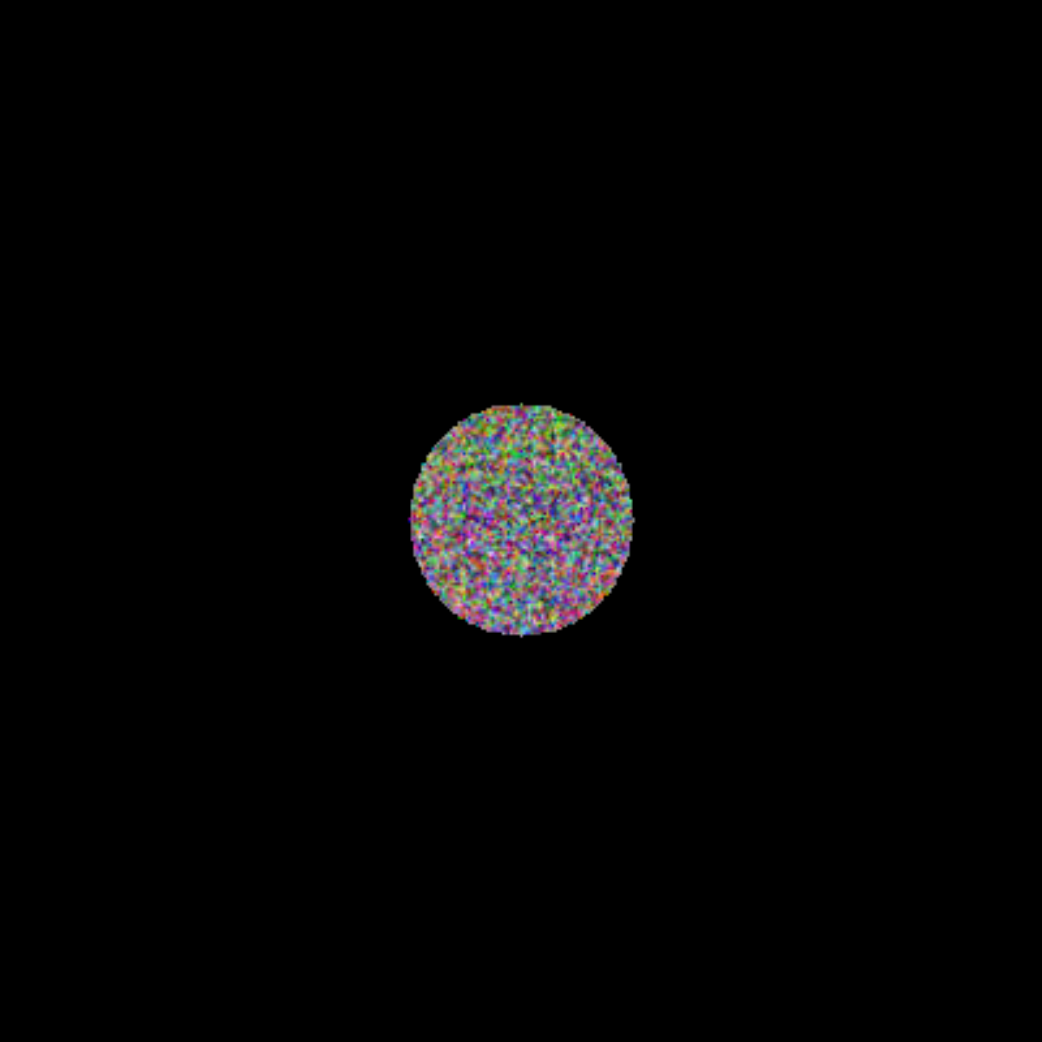}}
    \centerline{Input}
\end{minipage}
\begin{minipage}{0.070\textwidth}
    \centerline{\includegraphics[width=1\textwidth]{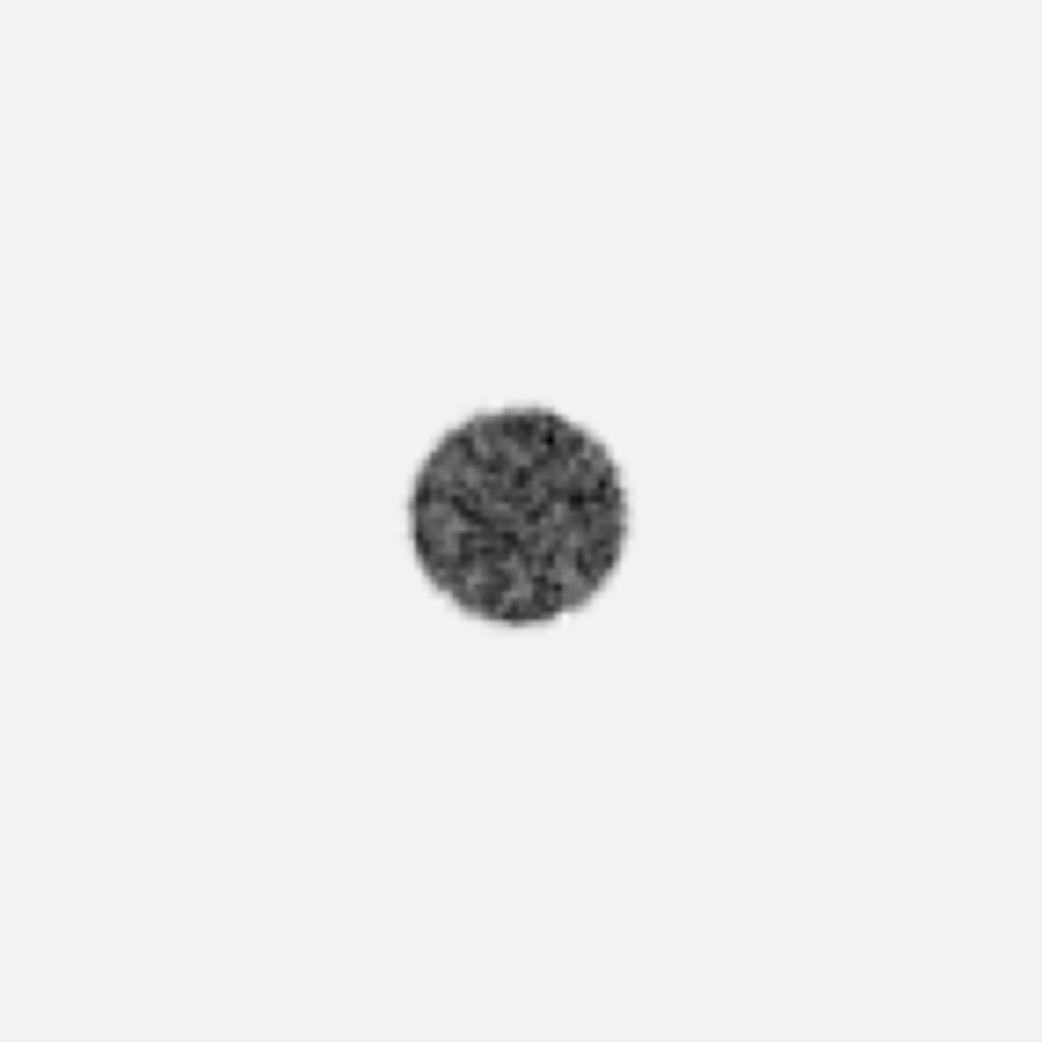}}
    \centerline{conv1}
\end{minipage}
\begin{minipage}{0.070\textwidth}
    \centerline{\includegraphics[width=1\textwidth]{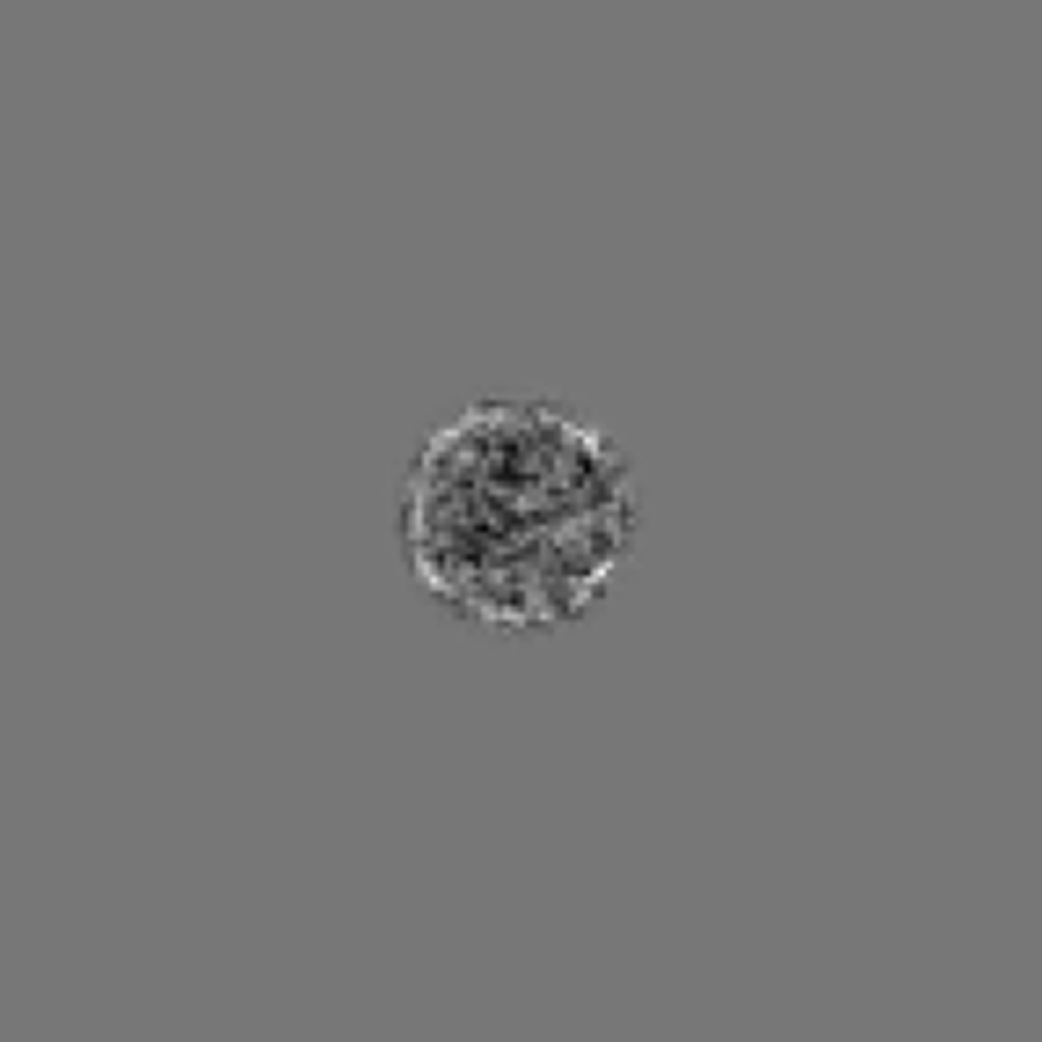}}
    \centerline{conv2}
\end{minipage}
\begin{minipage}{0.070\textwidth}
    \centerline{\includegraphics[width=1\textwidth]{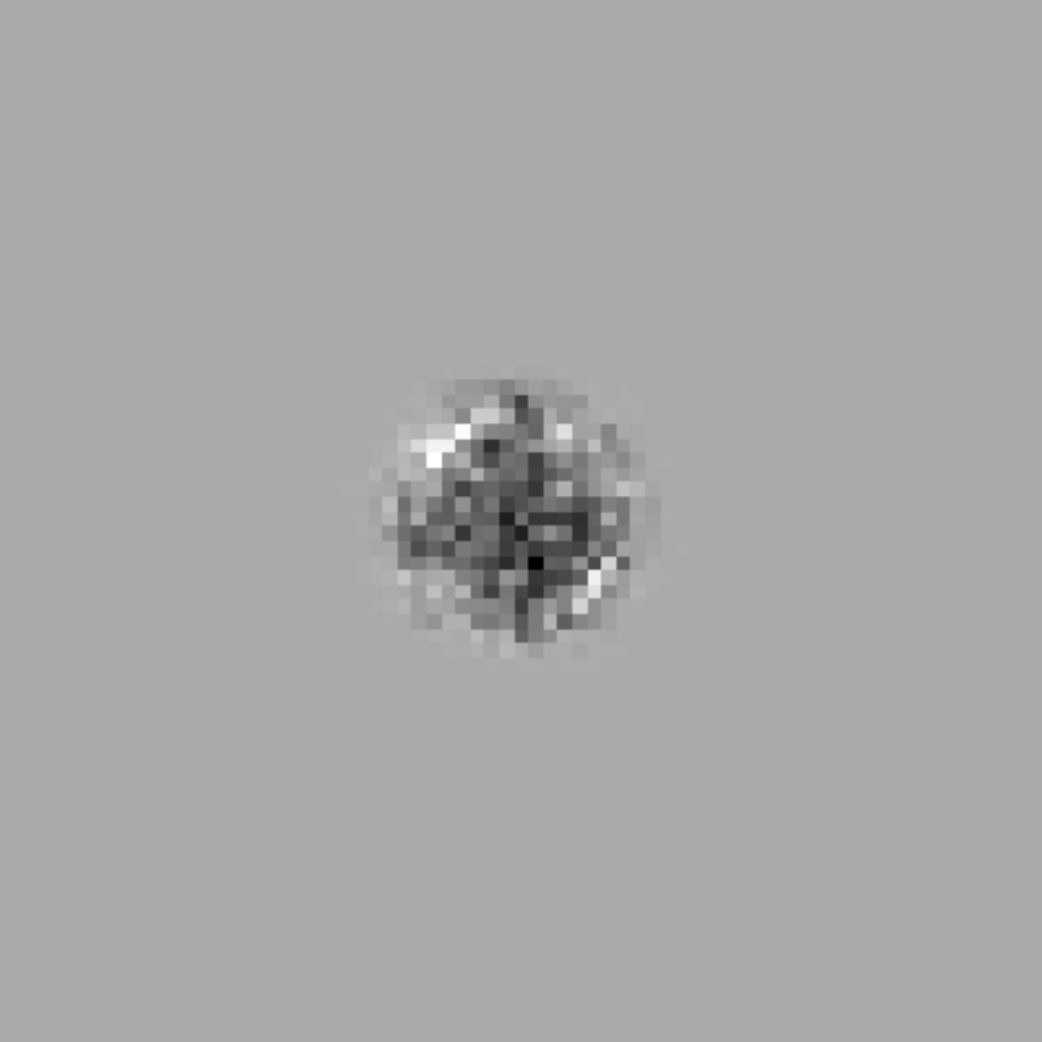}}
    \centerline{conv3}
\end{minipage}
\begin{minipage}{0.070\textwidth}
    \centerline{\includegraphics[width=1\textwidth]{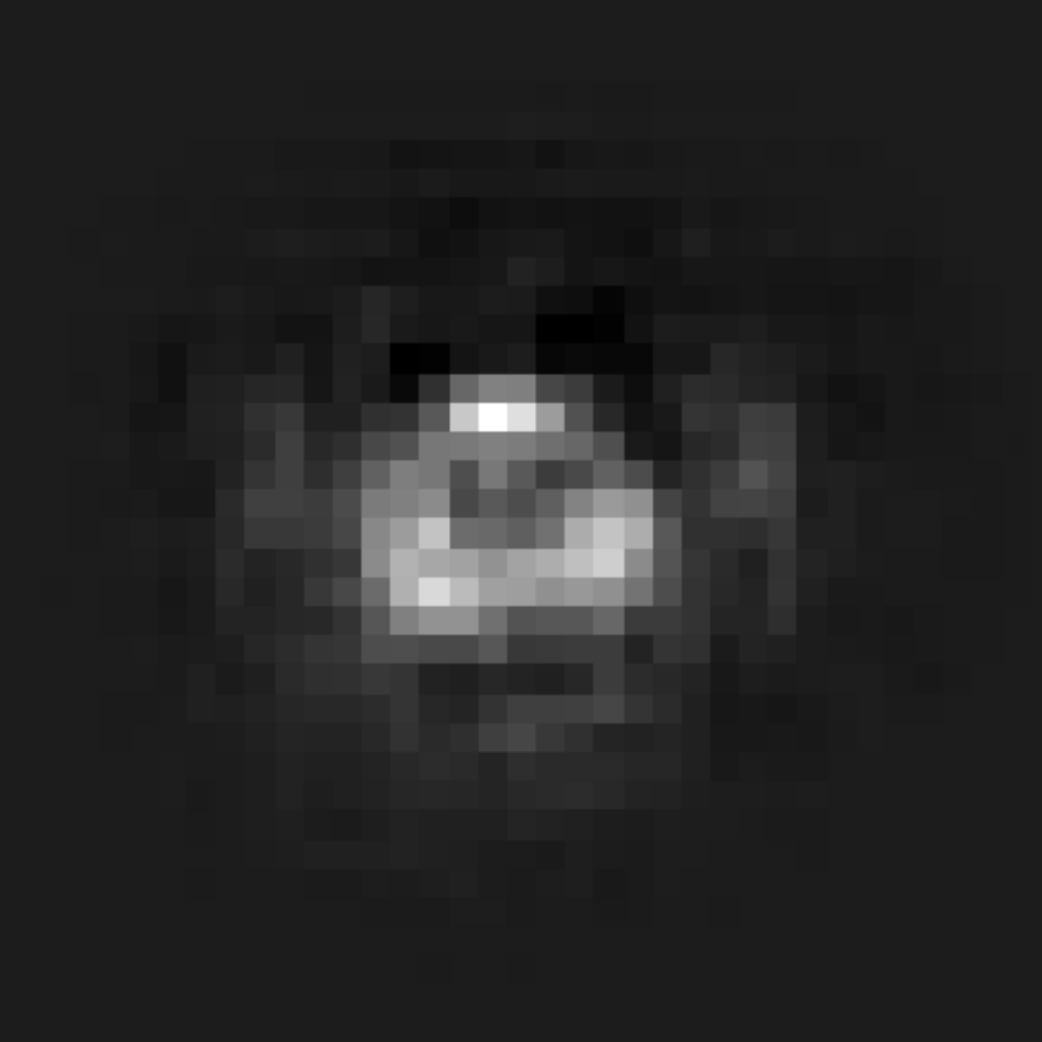}}
    \centerline{conv4}
\end{minipage}
\begin{minipage}{0.070\textwidth}
    \centerline{\includegraphics[width=1\textwidth]{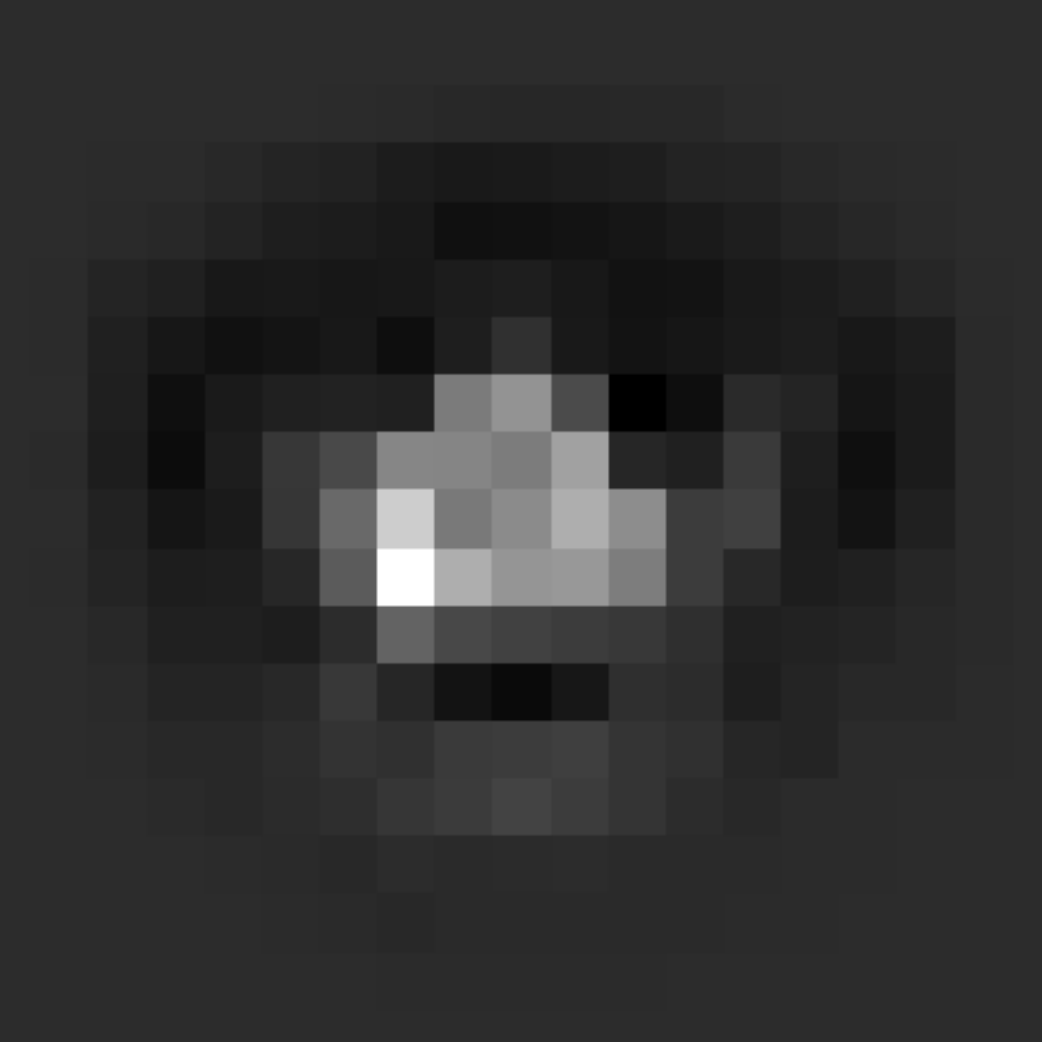}}
    \centerline{conv5}
\end{minipage}
\begin{minipage}{0.070\textwidth}
    \centerline{\includegraphics[width=1\textwidth]{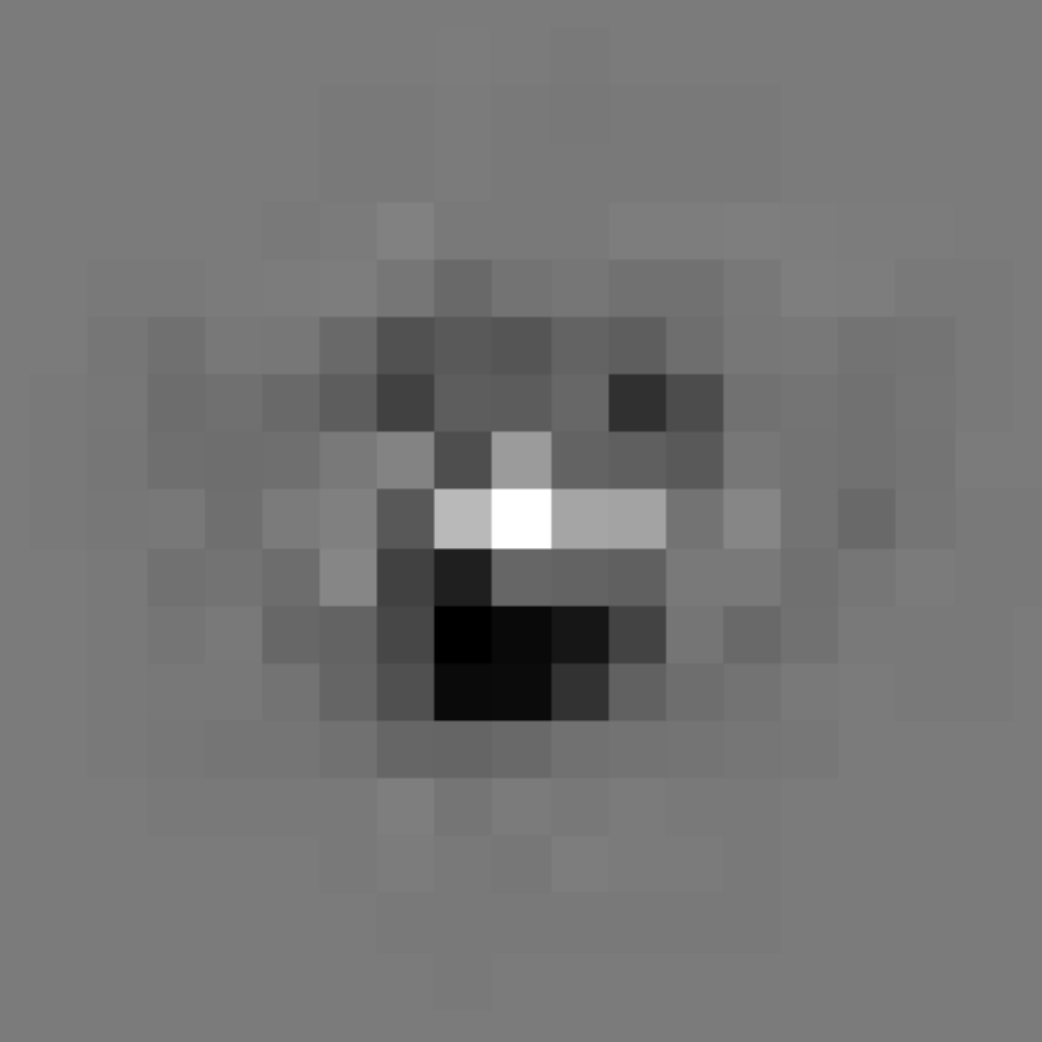}}
    \centerline{conv6}
\end{minipage}
\begin{minipage}{0.070\textwidth}
    \centerline{\includegraphics[width=1\textwidth]{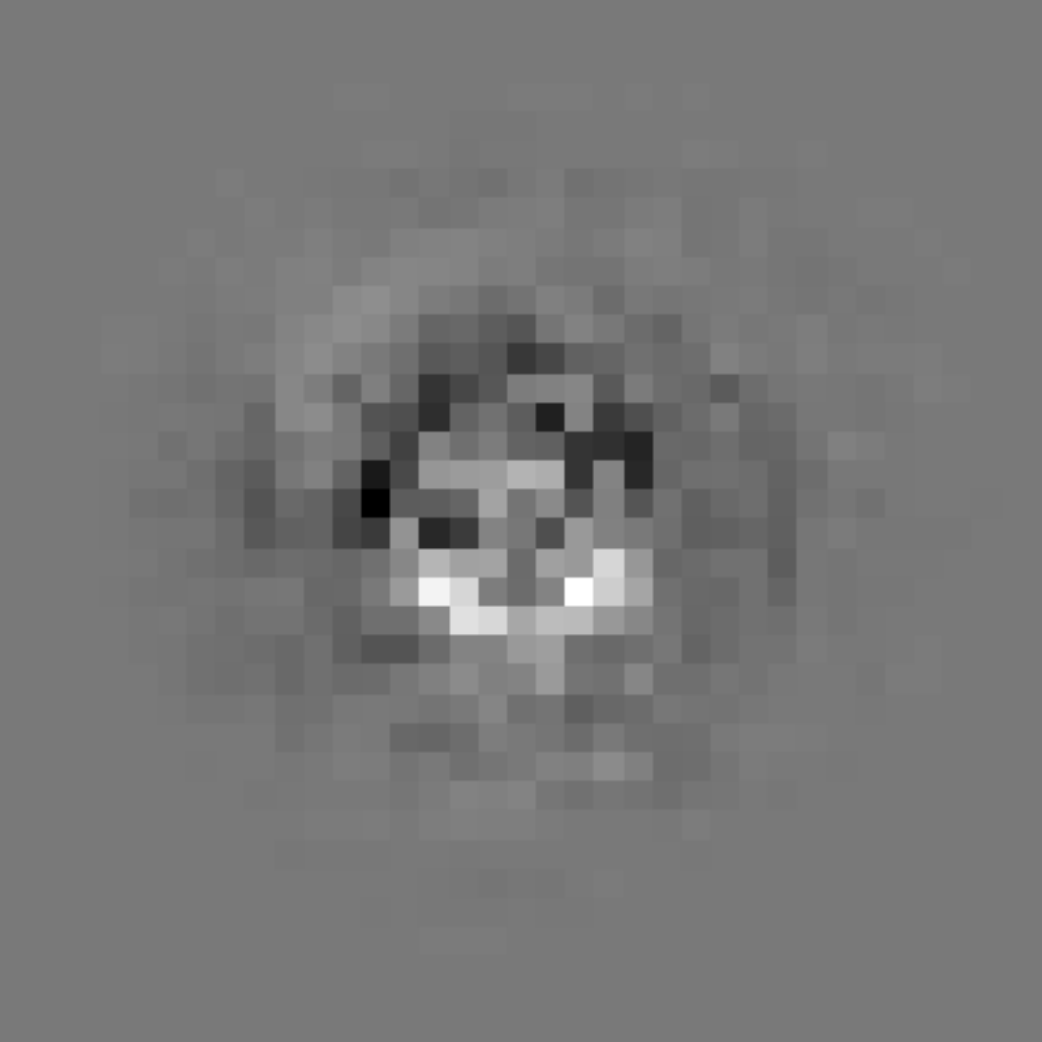}}
    \centerline{upsample1}
\end{minipage}
\begin{minipage}{0.070\textwidth}
    \centerline{\includegraphics[width=1\textwidth]{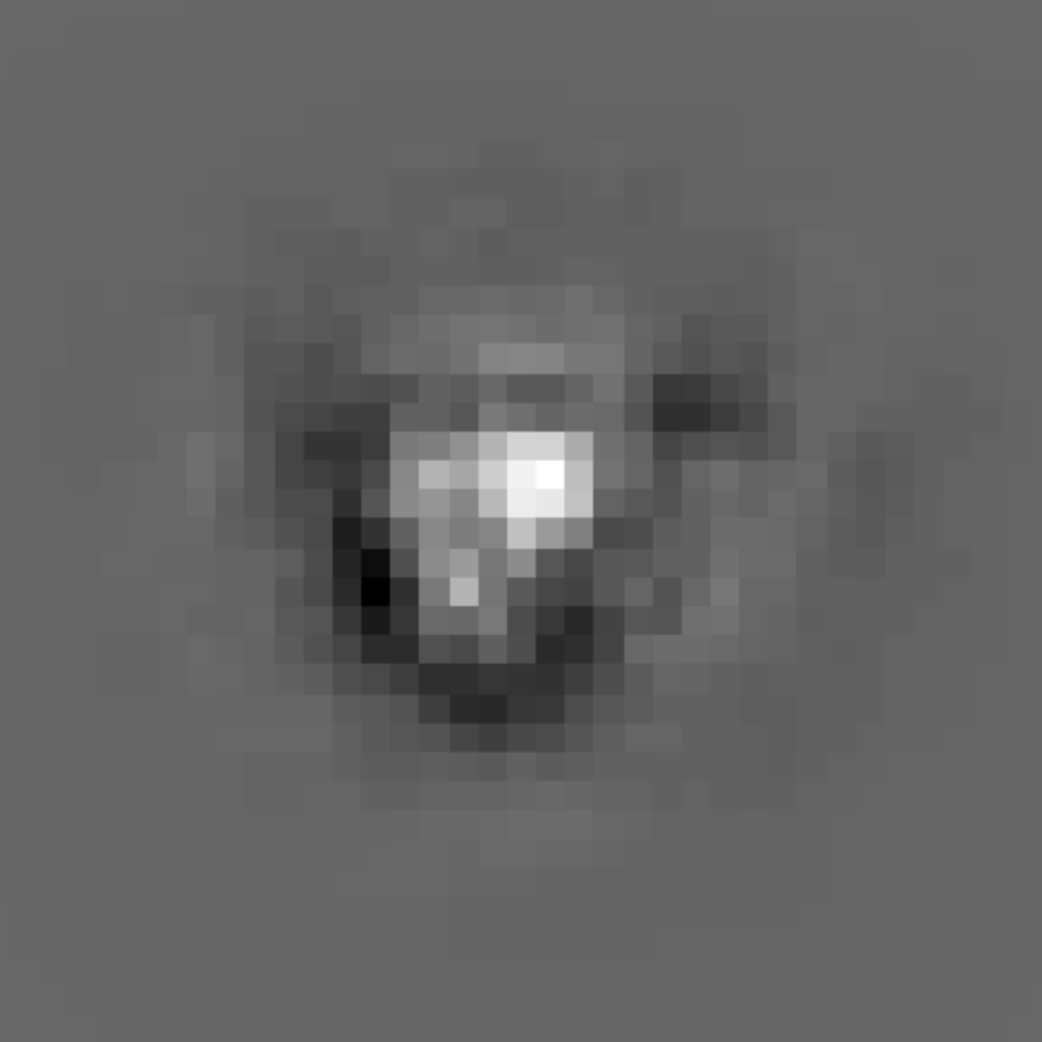}}
    \centerline{smooth1}
\end{minipage}
\begin{minipage}{0.070\textwidth}
    \centerline{\includegraphics[width=1\textwidth]{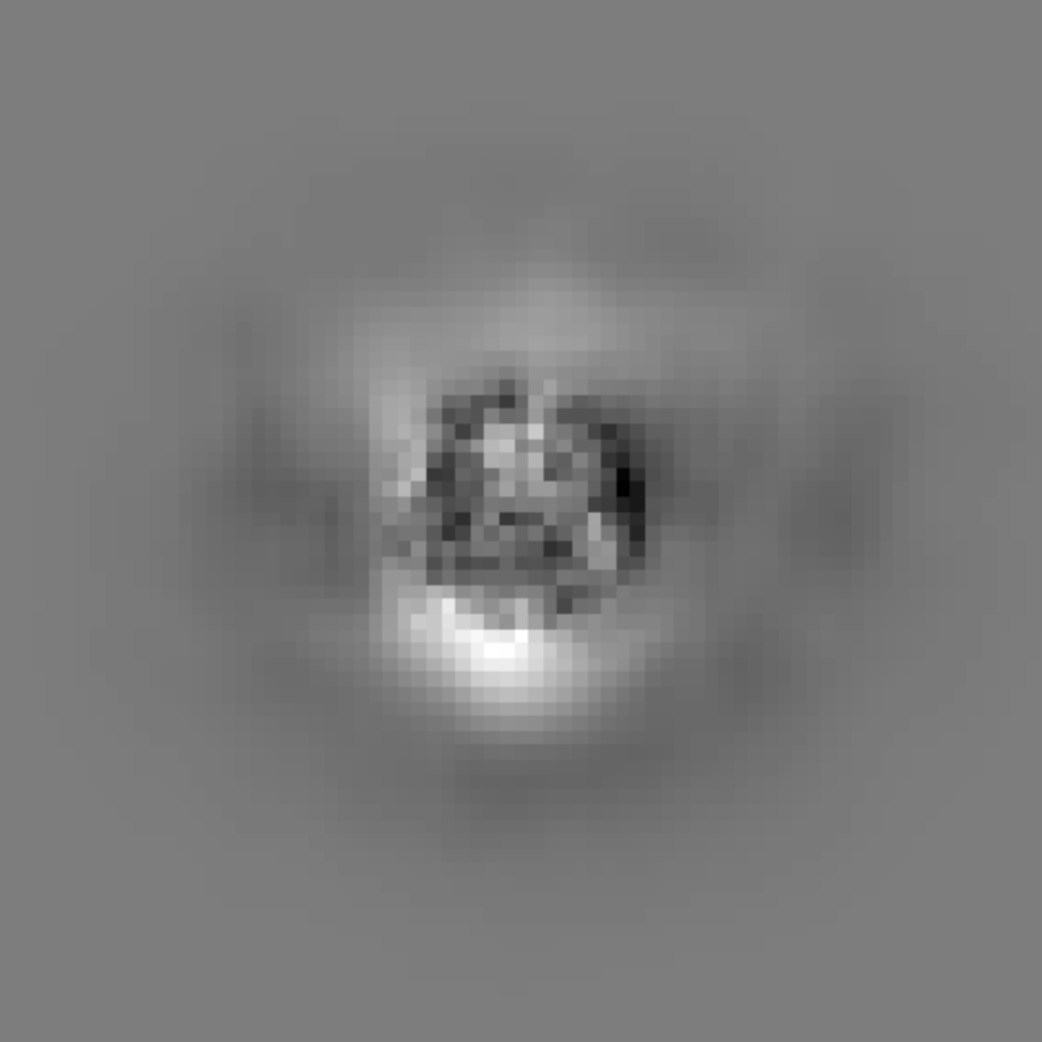}}
    \centerline{smooth2}
\end{minipage}
\begin{minipage}{0.070\textwidth}
    \centerline{\includegraphics[width=1\textwidth]{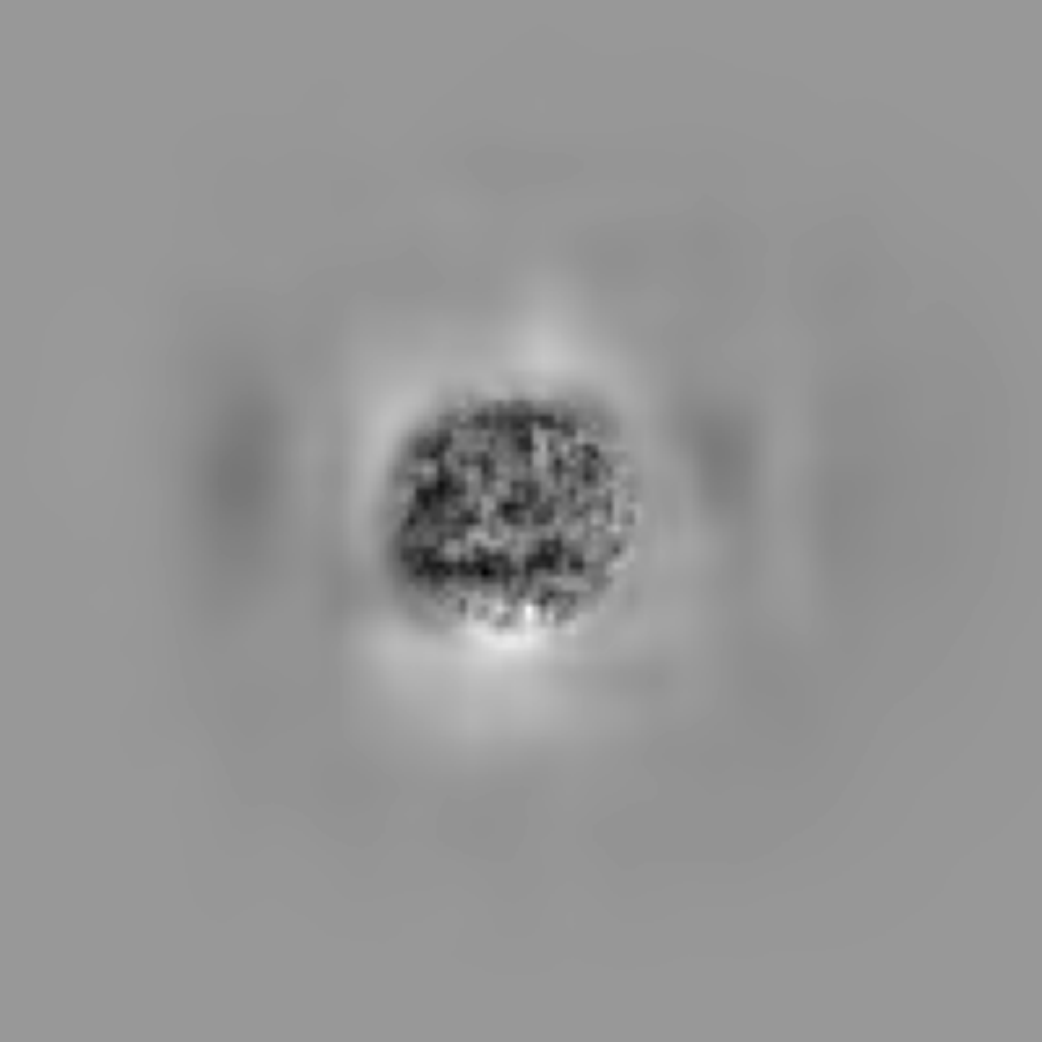}}
    \centerline{smooth3}
\end{minipage}
\begin{minipage}{0.070\textwidth}
    \centerline{\includegraphics[width=1\textwidth]{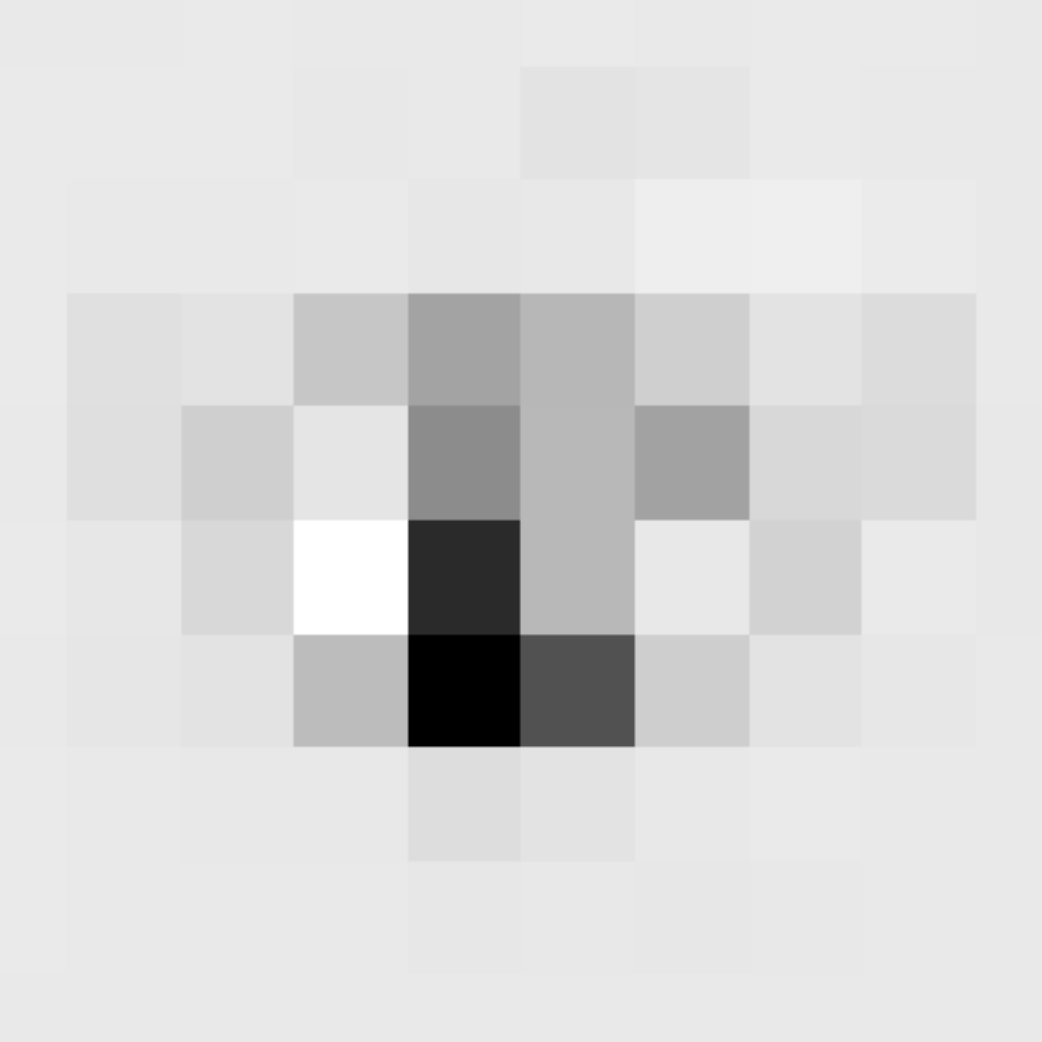}}
    \centerline{maxpool1}
\end{minipage} \\ \vspace{4.pt}
\begin{minipage}{0.070\textwidth}
    \centerline{\includegraphics[width=1\textwidth]{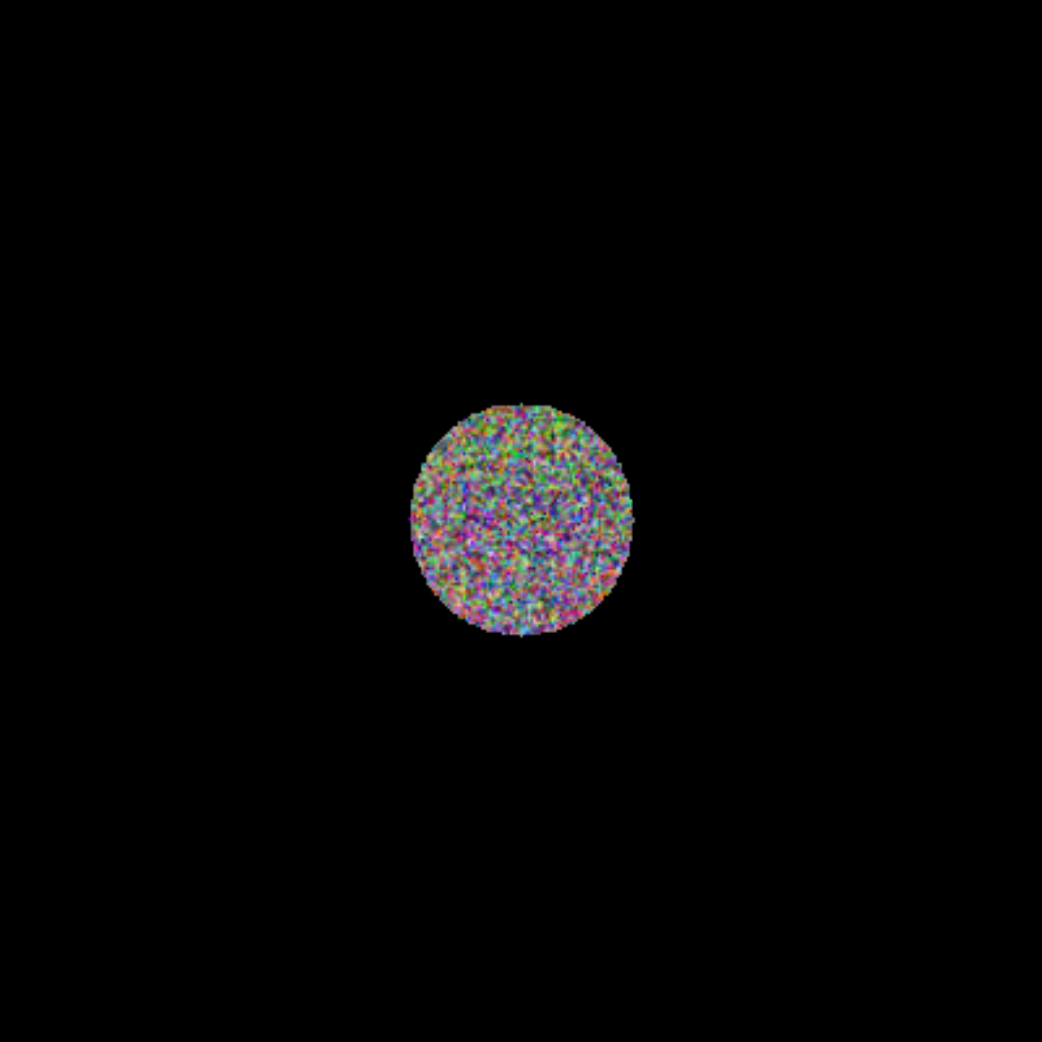}}
    \centerline{Input}
\end{minipage}
\begin{minipage}{0.070\textwidth}
    \centerline{\includegraphics[width=1\textwidth]{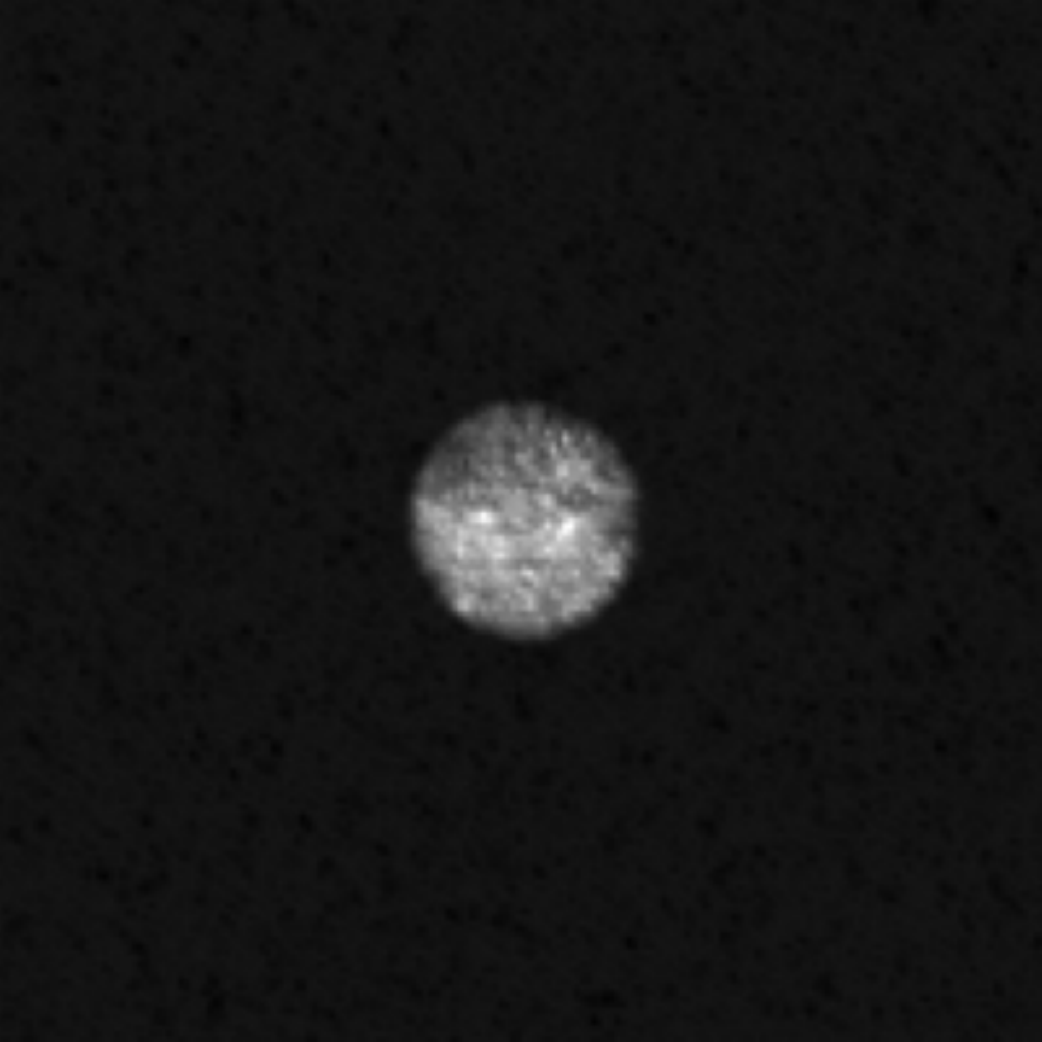}}
    \centerline{conv1}
\end{minipage}
\begin{minipage}{0.070\textwidth}
    \centerline{\includegraphics[width=1\textwidth]{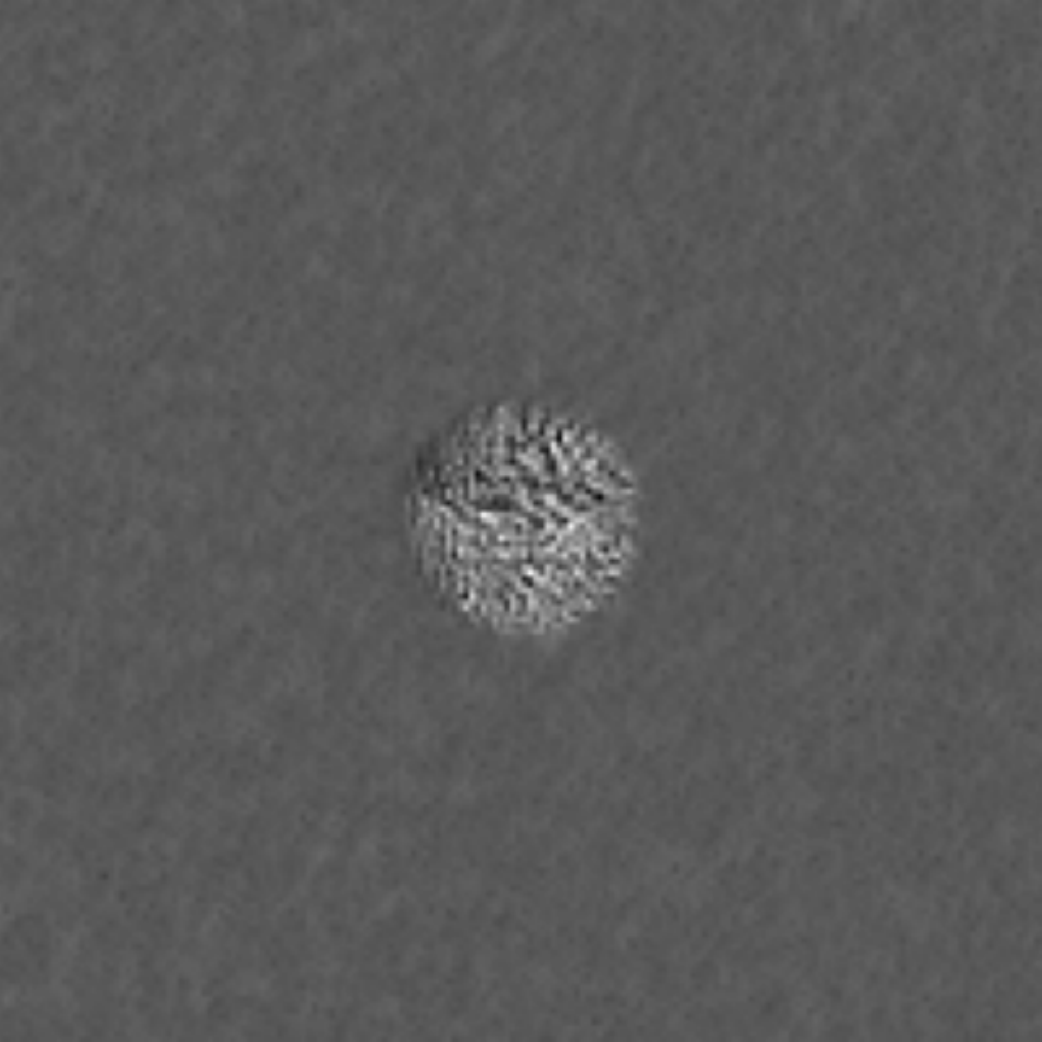}}
    \centerline{conv2}
\end{minipage}
\begin{minipage}{0.070\textwidth}
    \centerline{\includegraphics[width=1\textwidth]{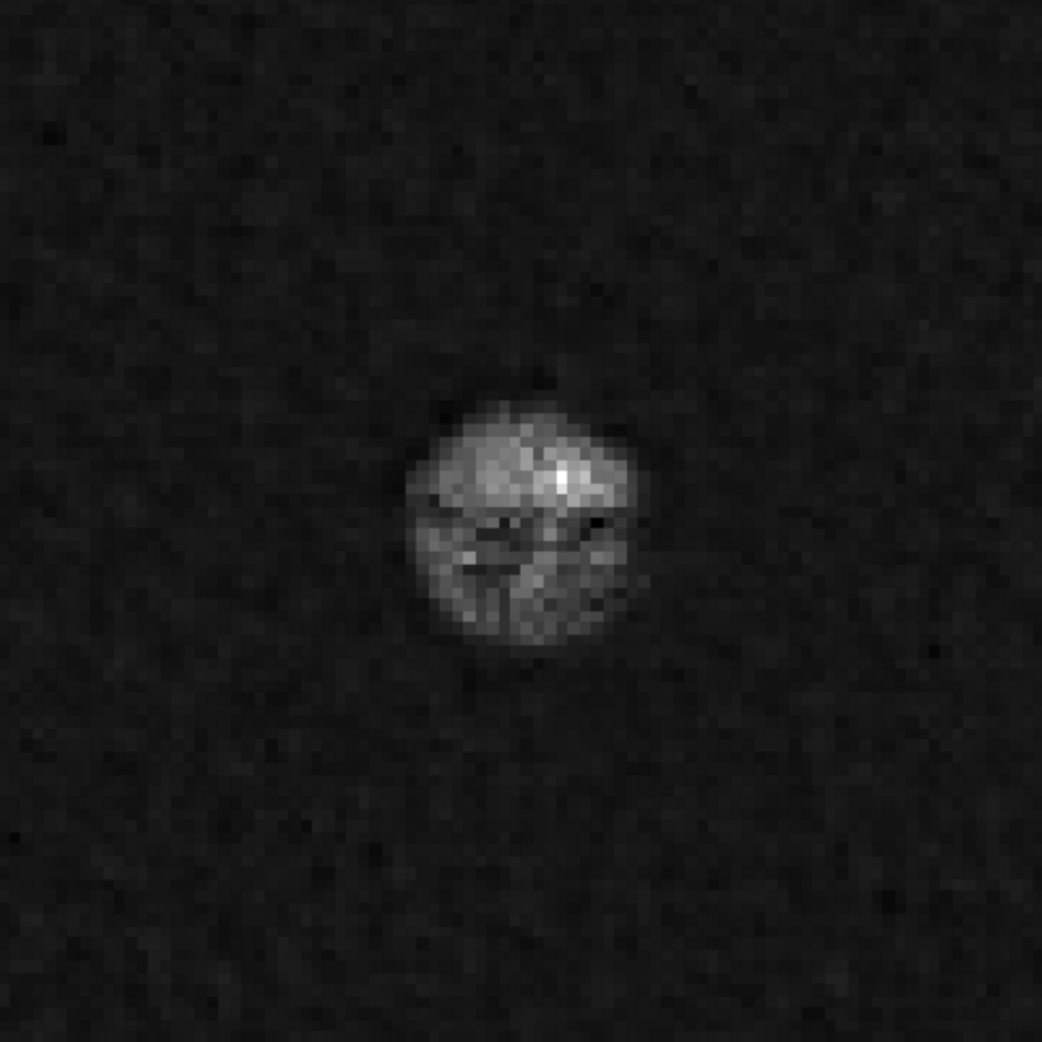}}
    \centerline{conv3}
\end{minipage}
\begin{minipage}{0.070\textwidth}
    \centerline{\includegraphics[width=1\textwidth]{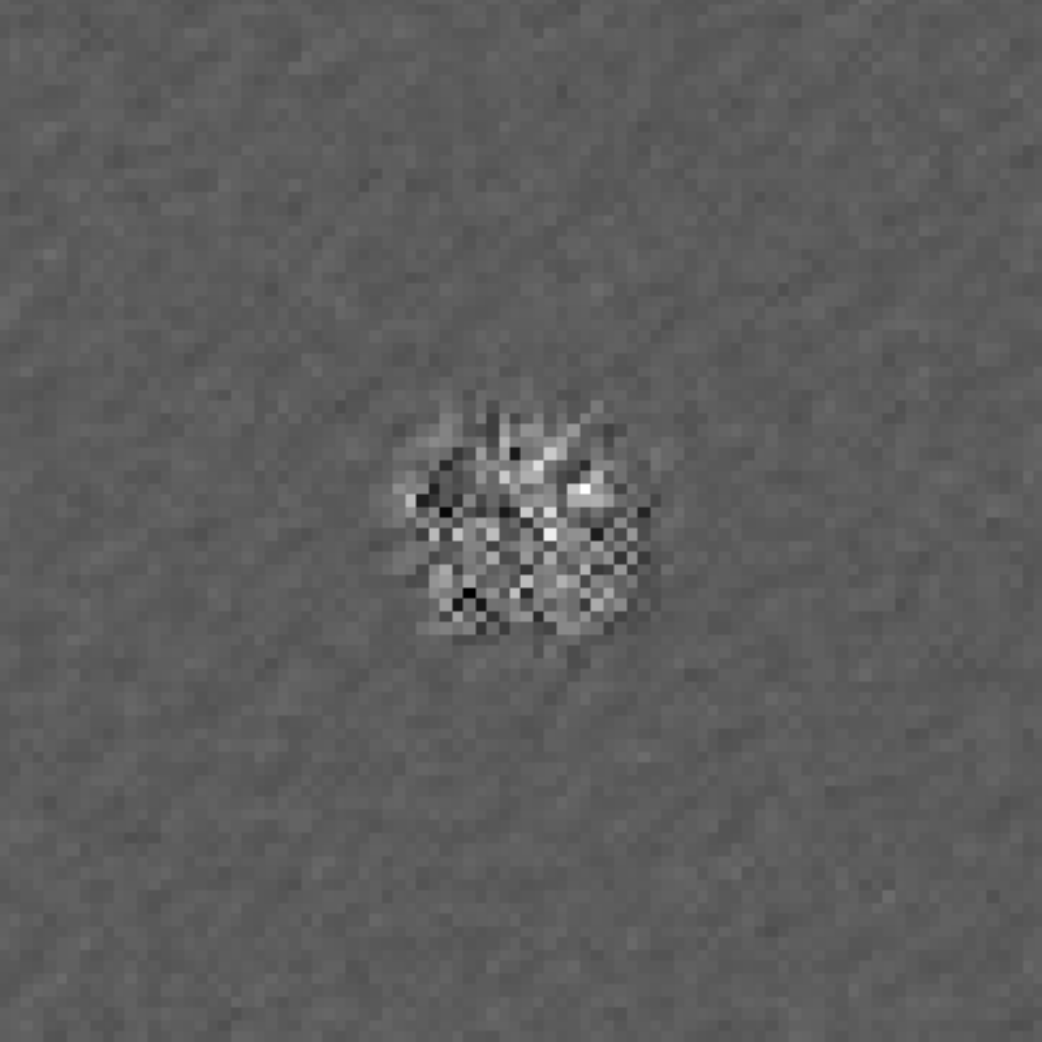}}
    \centerline{conv4}
\end{minipage}
\begin{minipage}{0.070\textwidth}
    \centerline{\includegraphics[width=1\textwidth]{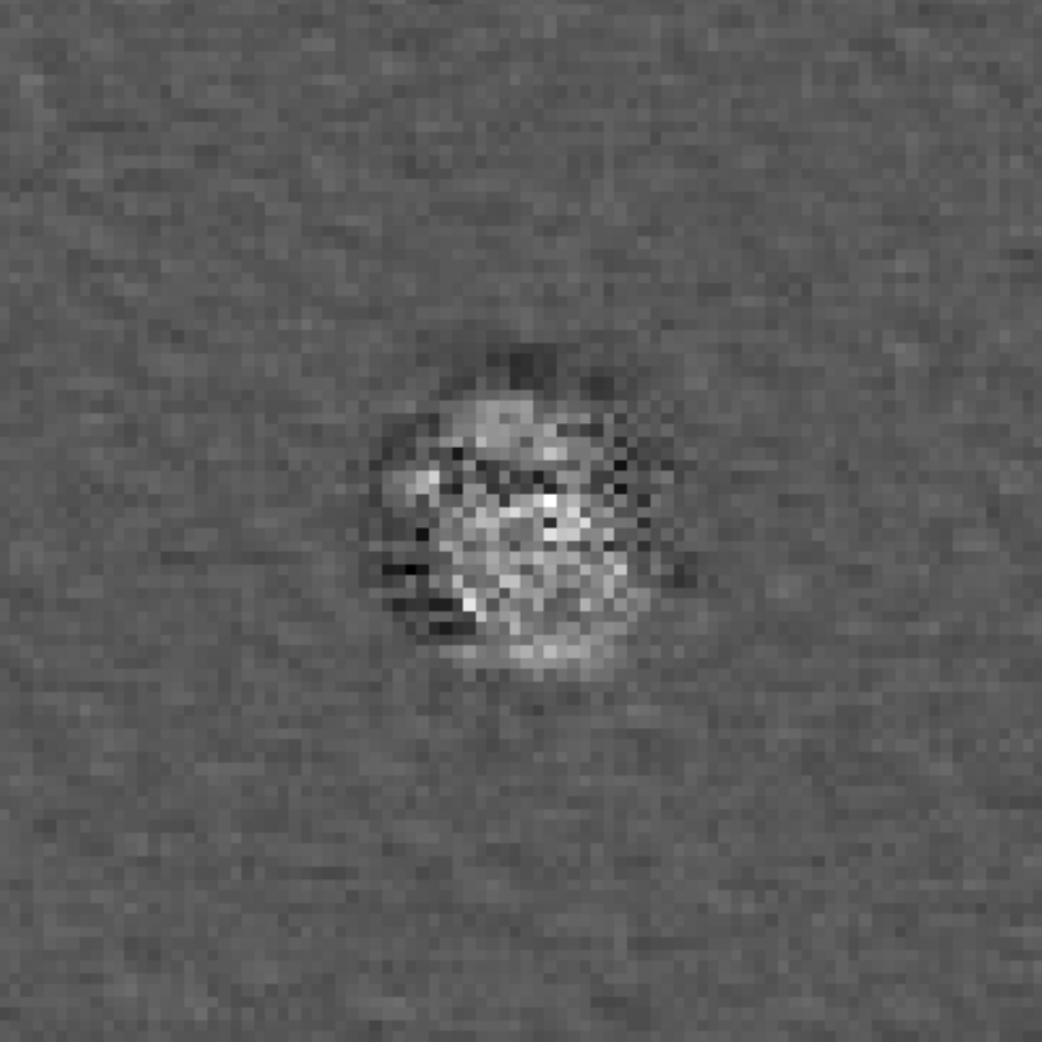}}
    \centerline{conv5}
\end{minipage}
\begin{minipage}{0.070\textwidth}
    \centerline{\includegraphics[width=1\textwidth]{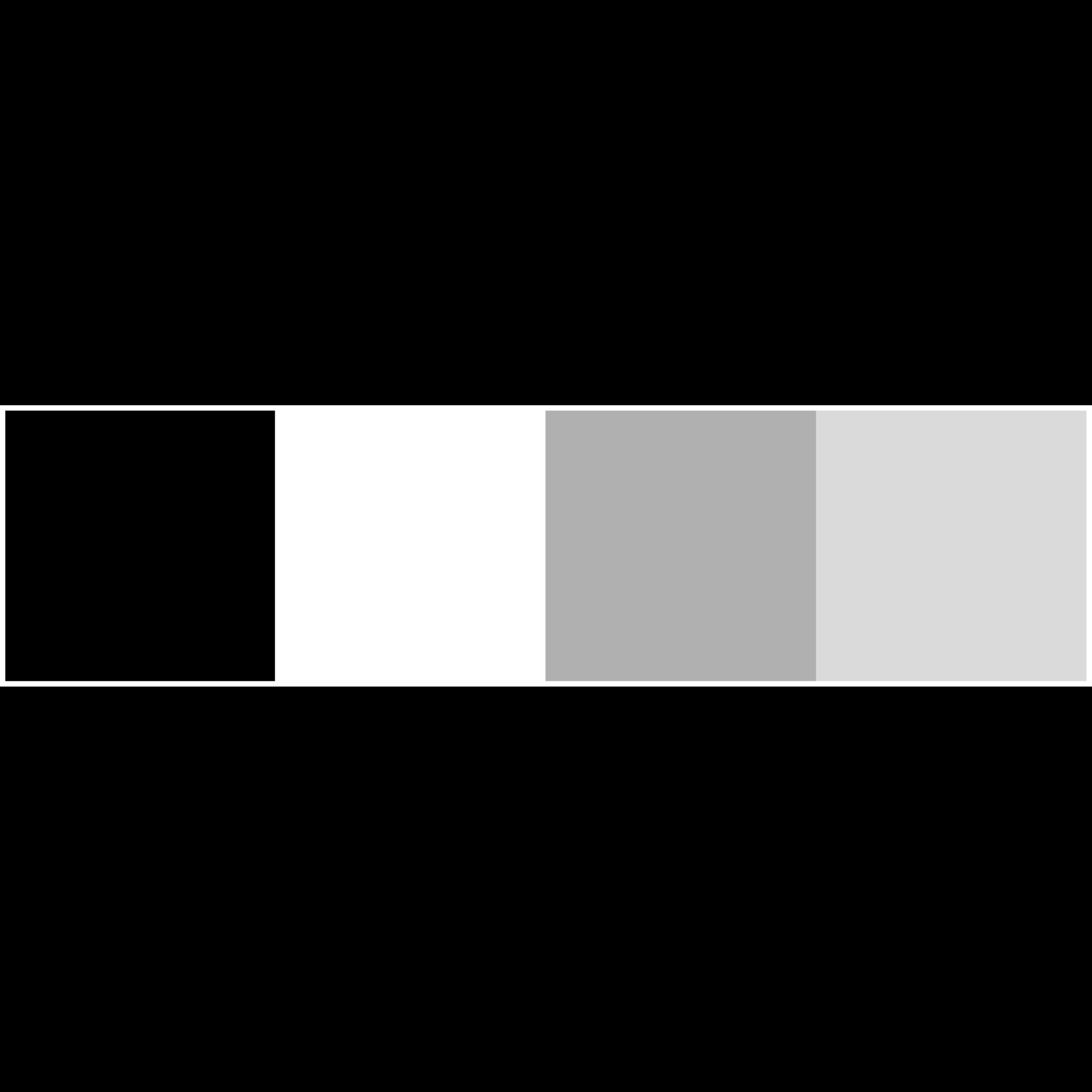}}
    \centerline{branch1}
\end{minipage}
\begin{minipage}{0.070\textwidth}
    \centerline{\includegraphics[width=1\textwidth]{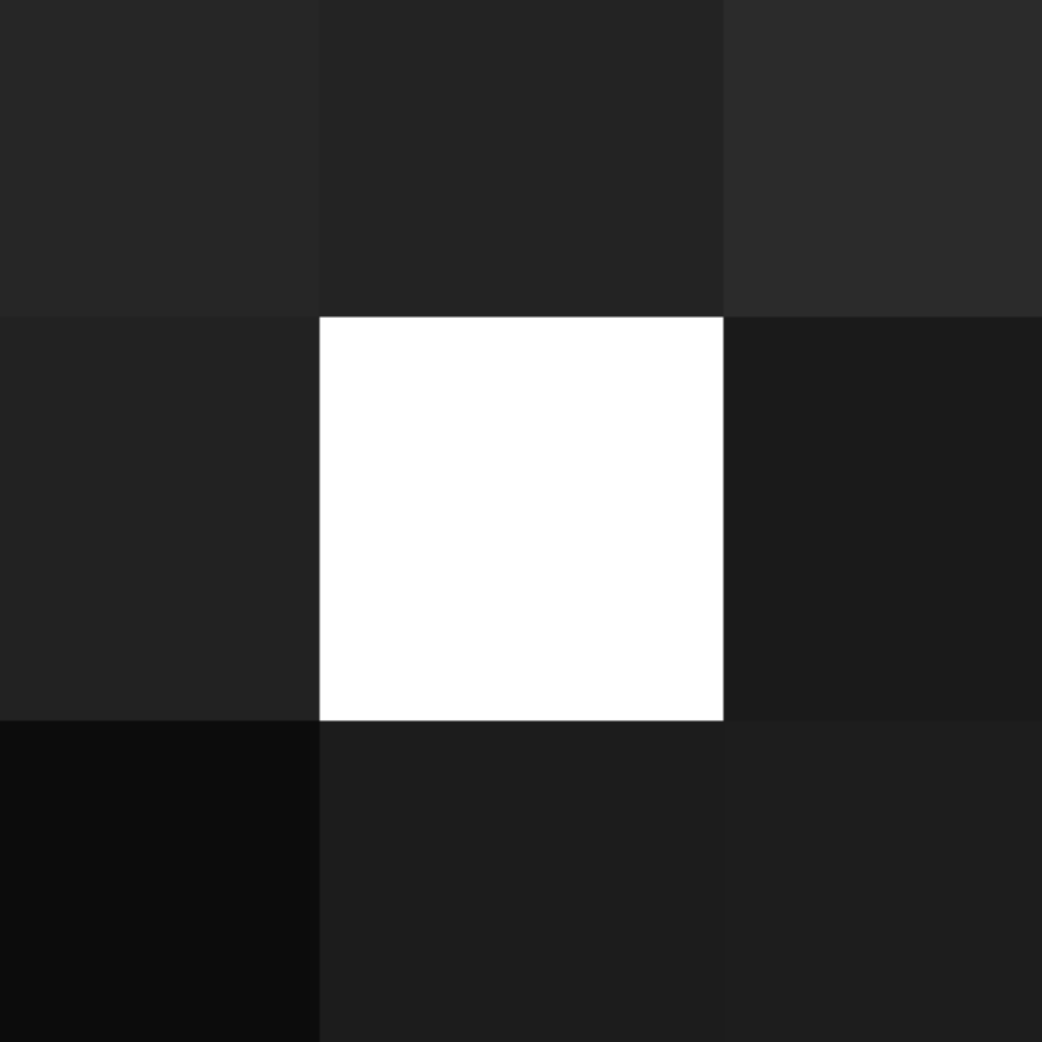}}
    \centerline{branch2}
\end{minipage}
\begin{minipage}{0.070\textwidth}
    \centerline{\includegraphics[width=1\textwidth]{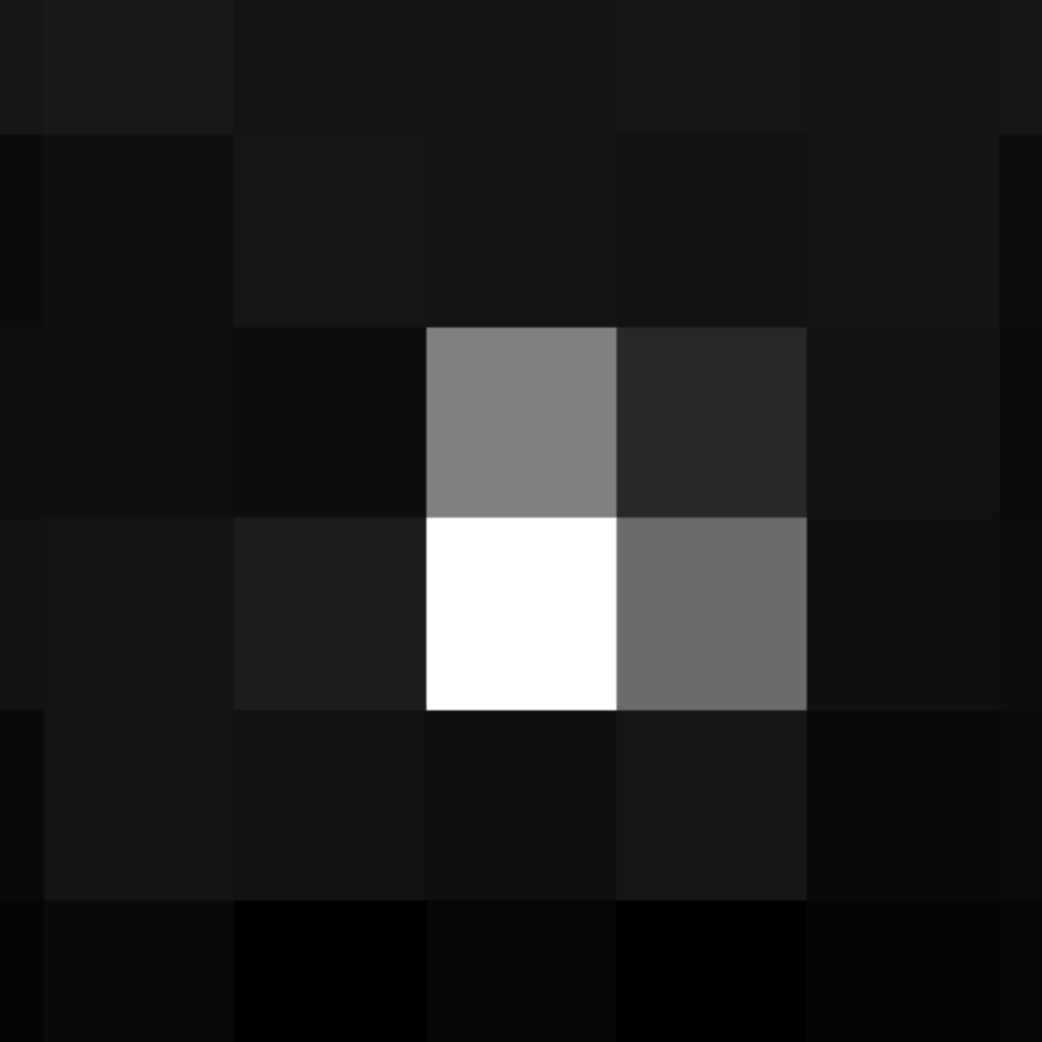}}
    \centerline{branch3}
\end{minipage}
\begin{minipage}{0.070\textwidth}
    \centerline{\includegraphics[width=1\textwidth]{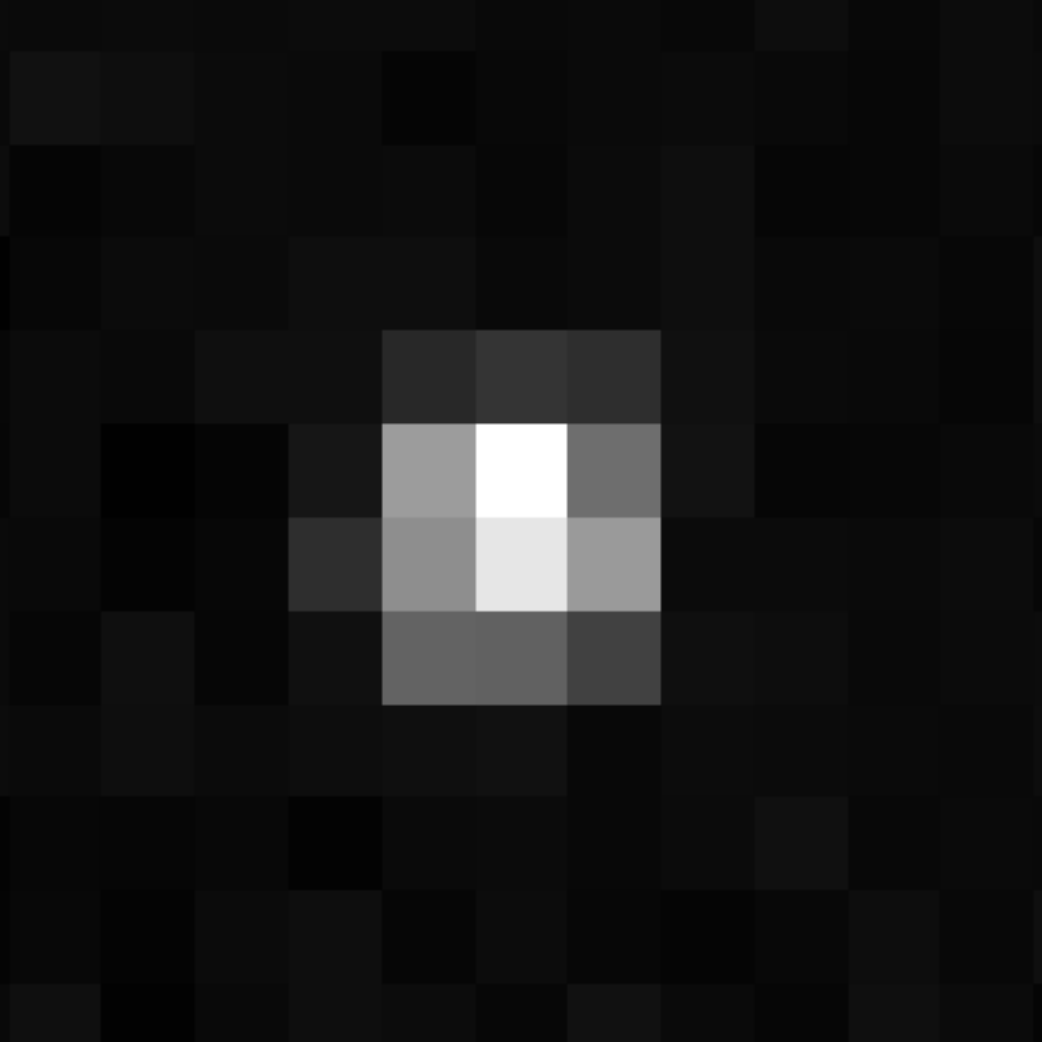}}
    \centerline{branch4}
\end{minipage}
\begin{minipage}{0.070\textwidth}
    \centerline{\includegraphics[width=1\textwidth]{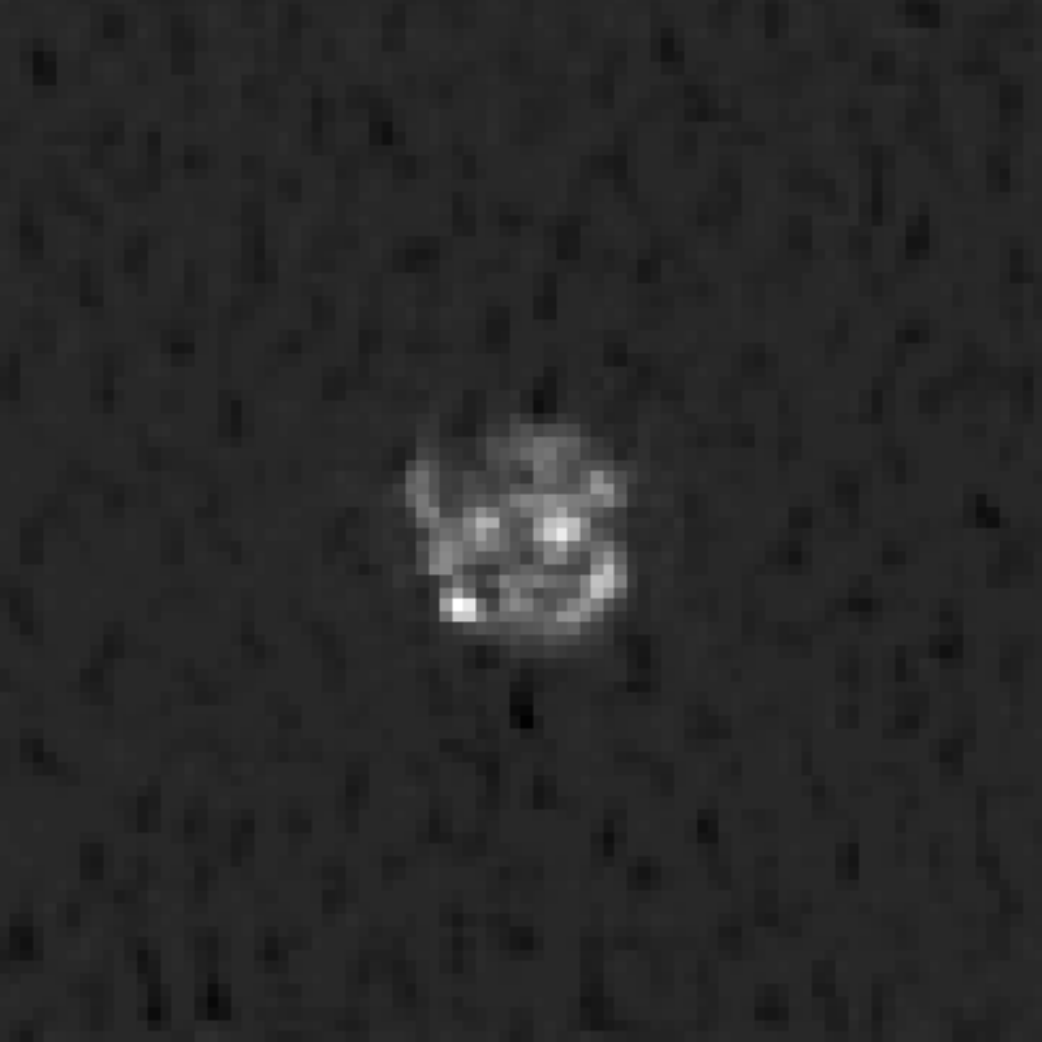}}
    \centerline{fusion1}
\end{minipage}
\begin{minipage}{0.070\textwidth}
    \centerline{\includegraphics[width=1\textwidth]{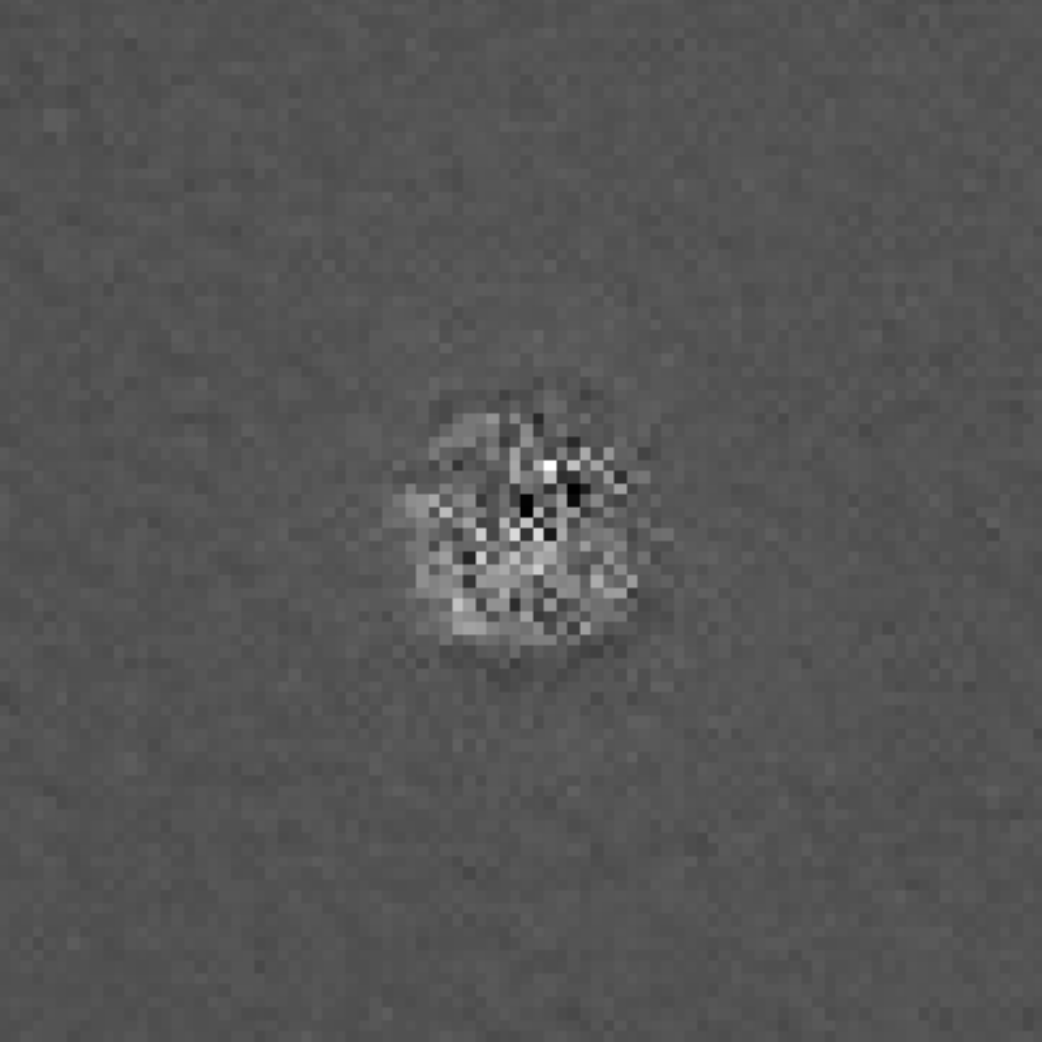}}
    \centerline{fusion2}
\end{minipage}
\caption{Visualization of the feature maps of Stereo R-CNN (the 1st row) and DSGN (the 2nd row) from shallow to deeper layers.}
\label{fig:diff_feat_map}
\end{figure*}

\subsection{The Cause of the Decoupling between the Precision of 3D Object Detectors and Driving Safety under Adversarial Attacks}

From the experiments in Section~\ref{sec.experiments}, we observe that  perturbation attacks cause a significant precision decline in 3D object detection, but  only a slight change in driving safety performance. Patch attacks cause a slight decline in 3D object detection precision, but a relatively significant performance degradation in driving safety. To figure out the reasons beneath these results, we take a closer look at the detection results of the attacked and unattacked 3D object detectors and compare them accordingly.

The reasons for the decoupling caused by patch attacks are relatively straightforward. First, we notice that the affected area of patch attacks is limited to the patch itself where the patch is usually quite small in order to make it difficult to be detected. Thus, patch attacks can only  trigger the object detectors to produce a very small number of ghost 3D bounding boxes inside the patch.  This is the reason why the object detection precision does not show a significant decline when detectors are under patch attacks. Second, since the adversarial patch is randomly placed in driving scenarios, the resulted ghost 3D bounding boxes have a fair chance to appear on the road surface and block the way of the ego-vehicle, which directly leads to noticeably driving safety performance degradation. These two reasons together explain  the decoupling between the precision of 3D object detectors and driving safety under patch attacks. In the rest of this section, we mainly focus on investigating the reasons for the decoupling caused by perturbation attacks.

Apart from the fact that the perturbation attacks cause slight drifts of 3D bounding boxes of real objects which are originally produced accurately when no attack is launched, the most significant consequence of a perturbation attack is that it triggers the object detectors to produce a lot of ghost 3D bounding boxes which do not circle any specific or meaningful object inside. In particular, almost all ghost boxes appear in the side areas of a road instead of on the surface of a road. Since the optimal trajectory generated by the motion planner most likely will not traverse the side areas of a road, the ghost objects will not affect the trajectory generated by the motion planner. In other words, the trajectories generated before and after perturbation attacks are essentially the same. Thus, driving safety is not affected dramatically by permutation attacks. 

We further investigate why the ghost 3D bounding boxes caused by perturbation attacks tend to appear in the side areas of a road. After inspecting the positions where ghost bounding boxes appear in a large number of driving scenarios, we hypothesize that the difference in the texture complexity between the side areas of a road and the road surface may be the cause of this. The reason is that the texture of the road surface is more regular than the texture of the side areas of a road. Thus,  it takes more ``efforts" for perturbation attacks to change the pixel values to generate ghost boxes on the road surface than that in the side areas of a road.

In order to validate our hypothesis, we design a texture replacement experiment. Specifically, for a driving scenario in which vision-based 3D object detectors under a perturbation attack produce ghost 3D bounding boxes in  the side areas of a road, we replace the texture of the side areas with the texture of the road surface, then feed this modified driving scenario to the attacked object detectors and check the detection results. If our hypothesis is correct,  we shall expect that the attacked object detectors do not produce any ghost boxes in the side areas of a road for the modified driving scenario.

The results of the texture replacement experiments are shown in Figure~\ref{fig:ab_1_stereo} for the Stereo R-CNN model and in Figure~\ref{fig:ab_1_dsgn} for the DSGN model. The attack setting of the perturbation attack applied here is set to be $\alpha=1$ and $n=4$ in Eqn.~(\ref{eq:eq_2}). From the figures, we can observe that both 3D object detectors can detect the object accurately when there is no perturbation attack applied (Figure~\ref{fig:ab_1_stereo_1}, \ref{fig:ab_1_dsgn_1}), and ghost 3D bounding boxes appear in the side areas of a road when the perturbation attack is launched on the same driving scenario (Figure~\ref{fig:ab_1_stereo_2}, \ref{fig:ab_1_dsgn_2}). More importantly, after we replace the texture of side area of road with the texture of road surface (Figure~\ref{fig:ab_1_stereo_3}, \ref{fig:ab_1_dsgn_3}), no more ghost boxes are produced in the side area of the road by the 3D object detectors, which matches our expectation. Hence, the texture replacement experiment results validate our hypothesis that the difference in the texture complexity between the side areas of a road and the road surface leads to the decoupling between the precision of 3D object detectors and the driving safety performance metrics when the 3D object detectors are attacked.

\vspace{-0.05in}
\subsection{The Cause of Difference in Robustness}

The experiment results in Section~\ref{sec.experiments} indicate that DSGN is more robust than Stereo R-CNN in terms of driving safety and object detection when they are under adversarial attacks. Especially, when  patch attacks are launched, DSGN is more robust than Stereo R-CNN.

To better understand the cause of such a difference in robustness, we conduct a contrast experiment by implementing the black-box patch attack where instead of training a patch for each model separately, we learn a patch $p$ that is jointly optimized for both the DSGN model and the Stereo R-CNN model using Eqn.~(\ref{eq:eq_3}). Thus, the patch is capable of attacking both models. To conduct this experiment, we also generate an image $I$ with uniformly distributed random noise and paste the patch $p$ on $I$ to form the attacked input image $\tilde{I}$. We then feed the two input images into the models to observe the intermediate results produced by their network architectures. In Figure~\ref{fig:diff_feat_map}, we visualize the difference between corresponding intermediate feature maps generated from $I$ and $\tilde{I}$ for both models respectively. In other words, the feature map $F_k$ in Figure~\ref{fig:diff_feat_map} refers to the average norm of the difference between the $k$-th layer output with an attack applied and the $k$-th layer output without any attack. For each model, we inspect the intermediate feature map of layers from shallow to deep in the feature extraction part of its network architecture. Each feature map is cropped so that only the central part is used for the propose of demonstration.

It is the spatial propagation of the patch activation area in feature maps of a model that implies the robustness of the model to patch attacks. Specifically, if the patch activation area in feature maps propagates along the data flow direction of the network architecture, then the network architecture amplifies the impact of patch attacks on the model, suggesting weak robustness of the model to patch attacks. In contrast, if the patch activation area in feature maps does not propagate or even contracts along the data flow, then the network architecture of the model is more resilient to patch attacks, indicating stronger robustness of the model to adversarial patches.

In the first row of Figure~\ref{fig:diff_feat_map}, we show how the patch activation area propagates layer by layer in the Stereo R-CNN model. In the first few convolution layers ($conv$\textless1, 2, 3\textgreater), the patch activation area  is bounded by the original region. However, starting from the last two layers of the feature extractor ($conv$\textless4, 5\textgreater), we observe that the activation area gradually propagates as we move on to deeper layers. After the $maxpool1$ layer, the patch activation area propagates to  almost the entire cropped image. Since the patch activation area keeps propagating through the network architecture, Stereo R-CNN shows poor robustness under patch attacks.

In the second row of Figure~\ref{fig:diff_feat_map}, the DSGN model shows less propagation of the patch activation in the first three convolution layers ($conv$\textless1, 2, 3\textgreater), but the activation area at the last two layers of the feature extractor ($conv$\textless4, 5\textgreater) are expanded slightly. According to the network structure of DSGN, the feature extractor is connected to the Spatial Pyramid Pooling (SPP) module, and the outputs of SPP branches ($branch$\textless1, 2, 3, 4\textgreater) are fused with features from the former layers for future prediction. Interestingly, we observe that the patch activation area shrinks to its original size after the SPP module ($fusion$\textless1, 2\textgreater). Hence, different from  Stereo R-CNN, the SPP module in the DSGN model restrains the propagation of the patch impact. This demonstrates that DSGN has strong robustness to adversarial patches due to the SPP module in its network architecture. A similar observation of the Spatial Pyramid structure can be found in another study~\cite{ranjan19attacking}. We can conclude that when the adversarial patch is used to attack the model of Stereo R-CNN whose network architecture is not equipped with the SPP module, the patch exploits the weakness of the network architecture and amplifies its impact on 3D object detection. For the DSGN model, the SPP module in its network architecture restrains the impact of the adversarial patch on 3D object detection. Therefore, DSGN and Stereo R-CNN have different robustness to patch attacks and demonstrate different performance on the average precision of 3D object detection and the driving safety metrics.

\vspace{-0.05in}
\section{Conclusion}
\label{sec.conclusion}

In this paper, we have systematically investigated the impact of adversarial attacks not only on the object detection precision, but also on the driving safety of vision-based autonomous vehicles. Specifically, we proposed an end-to-end driving safety evaluation framework with designed performance metrics for the assessment of driving safety. Through extensive evaluation experiments, we found that a significant precision decline of 3D object detectors under the perturbation attack only leads to a slight decline in the driving safety performance metrics, but a mild precision decline of 3D object detectors under the patch attack can result in a significant performance degradation in driving safety. This finding suggests that it is desirable to evaluate the robustness of deep learning models in terms of driving safety rather than model precision. The proposed work can help guide the selection of deep learning models. The code of our evaluation framework is available upon request.

In the future, based on our experiments and discovered causes, we plan to expand our study to the autonomous vehicles that fuse the information from both LiDARs and cameras, and consider other types of attacks, such as attacks against LiDARs. Furthermore, we plan to design countermeasures for deep learning models against adversarial attacks by leveraging the spatial pyramid structure. Our future studies will also be conducted in an end-to-end fashion.

\vspace{-0.1in}
\bibliographystyle{IEEEtran}
\bibliography{auto-driving_2020}

\end{document}